\documentclass{article}

\usepackage[main, final]{neurips_2026}

\usepackage[utf8]{inputenc} 
\usepackage[T1]{fontenc}    
\usepackage{hyperref}       
\usepackage{url}            
\usepackage{booktabs}       
\usepackage{amsfonts}       
\usepackage{nicefrac}       
\usepackage{microtype}      
\usepackage{xcolor}         

\title{DifFRACT: Diffusion Feature Reconstruction and Attribution for Circuit Tracing}

\author{%
  Artyom Mazur \\
  HSE University \\
  \And
  Nina Konovalova \\ MBZUAI \\
  HSE University \\ 
  \And
  Aibek Alanov \\
  HSE University \\ FusionBrain Lab \\ AXXX \\
}

\usepackage{amsmath}
\usepackage{float}
\usepackage{algorithm}
\usepackage{algorithmic}
\usepackage{multirow}
\usepackage{enumitem}
\usepackage{graphicx}
\usepackage{booktabs}
\usepackage{makecell}

\usepackage[dvipsnames,svgnames]{xcolor}
\usepackage{tikz}
\usetikzlibrary{positioning, arrows.meta, decorations.pathreplacing, shapes.geometric, fit, backgrounds, calc, matrix}

\begin{document}

\maketitle

\begin{abstract}
Mechanistic interpretability seeks to explain neural network behavior by decomposing model computations into interpretable features and circuits. While transcoder-based circuit tracing has recently enabled detailed causal analyses of large language models, it remains unclear how semantic information propagates across denoising steps and how text and image representations interact within multimodal diffusion transformers (MM-DiT) for image generation. Existing methods provide only partial insight: attention maps expose a limited view of token interactions, while sparse autoencoders can discover interpretable features but do not directly reveal how these features are transformed and composed through nonlinear MLP layers.
In this work, we extend transcoder-based circuit tracing to multimodal diffusion transformers. We train timestep-conditioned transcoders that faithfully approximate the input-output behavior of the MLP sublayers in the double-stream blocks of FLUX.1[schnell]. By replacing MLPs with transcoders and linearizing the remaining computation, we obtain exact feature-to-feature attribution and build compact attribution graphs over interpretable features. Empirically, our transcoders match or slightly outperform sparse autoencoders on the sparsity-faithfulness tradeoff. The graphs recover the known shift from prompt-driven to image-driven computation across denoising steps, point to candidate mechanisms of attribute binding and cross-stream semantic propagation, and yield intervention-testable explanations for systematic generation errors. On prompt grids labeled blind, graph-guided interventions reach effects that single-feature steering does not; at a matched budget, features chosen from the graph fix a color prior more often than the target feature alone and destroy the object less often than features chosen by co-activation. FLUX.1[dev] reproduces the temporal and color-prior results, and SD3.5-medium reproduces them in direction.
Our results demonstrate that transcoder-based circuit analysis is feasible for state-of-the-art multimodal diffusion transformers. The code is available at \href{https://github.com/Artalmaz31/DifFRACT}{https://github.com/Artalmaz31/DifFRACT}.

\end{abstract}

\section{Introduction}

Diffusion models~\cite{ho2020denoising, dhariwal2021diffusion} are the leading approach to high-fidelity text-to-image generation~\cite{rombach2022high,esser2024scaling}, yet the mechanisms that transform noise into semantically rich images remain largely opaque. Understanding \emph{how} they perform this transformation is an open challenge for reliability, controllability, and safety.

Sparse autoencoders (SAEs) have become a widely adopted tool for mechanistic interpretability in LLMs~\cite{bricken2023monosemanticity,cunningham2023sparse} and have recently been extended to diffusion models, where they identify semantically meaningful visual features and support steering of generated outputs~\cite{cywinski2025saeuron,huang2026tide}. However, an SAE feature is typically dense in the neuron basis -- a combination of many neurons, each with its own nonlinearity~\cite{nanda2023open,dunefsky2406transcoders} -- which makes it difficult to trace how a feature in one layer influences a later feature through the intervening MLP sublayers.

To overcome these limitations, circuit tracing methods developed for large language models~\cite{ameisen2025circuit} construct attribution graphs over interpretable features, recovering the sequence of intermediate computations that a model uses to produce a given output. Their key enabler is the transcoder~\cite{dunefsky2406transcoders}, a model that approximates the input-output behavior of an MLP sublayer with a wide, sparsely activating MLP. Unlike a sparse autoencoder, which only reconstructs activations at a single point, a transcoder models the nonlinear transformation itself and so makes the connections between features through the MLP explicit; attribution graphs are built from them.

Diffusion models pose additional challenges for circuit-level analysis: they operate over multiple denoising timesteps and, in multimodal diffusion transformers (MM-DiT)~\cite{esser2024scaling}, maintain separate image and text streams that interact through joint attention. We extend transcoder-based circuit tracing to diffusion transformers.

Our main contributions are as follows:
\begin{itemize}[leftmargin=*, nosep]
    \item We propose the first application of transcoders to MM-DiT architectures, targeting the MLP sublayers of the double-stream blocks. The transcoders are conditioned on the denoising timestep, a design choice shared with timestep-aware SAEs~\cite{huang2026tide} that matters for the text stream (Appendix~\ref{app:tc:film}).
    \item We show that transcoders reach a comparable or modestly better sparsity-faithfulness tradeoff than sparse autoencoders (SAEs), while, unlike SAEs, they support feature-to-feature attribution through the MLP sublayers.
    \item We adapt circuit tracing to diffusion and find diffusion circuits -- candidate pathways of interpretable features, tested by intervention -- for object placement, style, semantic composition, and cross-stream interactions. On prompt grids labeled blind, these circuits support interventions that single-feature steering does not reach, and the graphs recover, stream by stream and layer by layer, the known shift from prompt-driven to image-driven computation across denoising steps.
\end{itemize}

\section{Related work}
\label{sec:related}
Early interpretability work on diffusion models used the U-Net bottleneck to find directions that edit attributes~\cite{kwon2022diffusion,park2023understanding} and attributed image regions to prompt words through cross-attention maps~\cite{tang2023daam}; sparse autoencoders find interpretable, steerable features in U-Nets~\cite{surkov2024one,ijishakin2024h,cywinski2025saeuron} and in FLUX~\cite{shabalin2025interpreting}, and TIDE~\cite{huang2026tide} conditions the SAE on the timestep. Other methods localize or edit a concept through spectral directions of cross-attention~\cite{bae2026mechanistic}, per-component scores~\cite{nguyen2025unveiling} or closed-form edits of the text projections~\cite{gandikota2024unified}. These works explain or control a concept through one kind of component or a score per component.

In language models, transcoders approximate the input--output map of an MLP sublayer and perform at least on par with SAEs~\cite{dunefsky2406transcoders}, and circuit tracing builds attribution graphs over their features to recover the intermediate computations behind an output~\cite{ameisen2025circuit}. We carry this to MM-DiTs: we decompose the preactivation of one feature into contributions of earlier features across layers and both streams, and test these paths by intervention. We compare transcoders with SAEs of the same architecture and timestep conditioning (§\ref{sec:exp:sae}), and graph-picked features with activation-based selection that an SAE could apply (§\ref{sec:exp:color_prior}). Appendix~\ref{app:related} covers it in more detail.

\section{Method}
\label{sec:method}

\subsection{Preliminaries}\label{sec:prelim}
We focus on FLUX.1~\cite{bfl2024flux,labs2025flux}, a multimodal diffusion transformer~\cite{peebles2023scalable,esser2024scaling} consisting of 19 double-stream blocks followed by 38 single-stream blocks. Double-stream blocks process image and text tokens with separate weights, allowing interaction only through a joint attention mechanism; single-stream blocks concatenate both streams and process them jointly. At the $512 \times 512$ resolution we use, the image stream has $1{,}024$ positions, one per latent patch, and the text stream $512$ T5 token positions. We restrict our analysis to double-stream blocks (§\ref{sec:limit}). The construction below needs only stream-specific MLP sublayers added to the residual stream through AdaLN-Zero gates, a modulation that is constant within a graph, and joint attention over the concatenated streams; Appendix~\ref{app:backbones} applies it to FLUX.1[dev], whose modulation also depends on the guidance scale, and to SD3.5-medium~\cite{stabilityai2024sd35medium}, whose $24$ blocks are all double-stream, $13$ with an extra image-only attention.

Each double-stream block applies a joint attention sublayer followed by two stream-specific MLP sublayers (FF in Figure~\ref{fig:flux_double_block}), both modulated by AdaLN-Zero conditioning derived from the denoising timestep $t$ and the pooled CLIP embedding. The attention sublayer computes and concatenates queries, keys, and values across streams, splits the output back per stream, and adds it to the residual, yielding $x_{\mathrm{mid}}^{(\ell,s)}$. The MLP sublayer then operates independently on each stream:

\begin{equation}
x_{\mathrm{post}}^{(\ell, s)} = x_{\mathrm{mid}}^{(\ell, s)} + \mathrm{gate}^{\ell, s}_{\mathrm{mlp}} \odot \mathrm{MLP}^{(\ell, s)}\!\left(\mathrm{AdaLN}_{\mathrm{mlp}}\!\left(x_{\mathrm{mid}}^{(\ell, s)}\right)\right), \label{eqn:flux_mlp_update}
\end{equation}
where $\mathrm{AdaLN}_{\mathrm{mlp}}$ stands for the LayerNorm-then-affine-modulate operation parameterized by $(\mathrm{scale}^{\ell, s}_{\mathrm{mlp}}, \mathrm{shift}^{\ell, s}_{\mathrm{mlp}})$. The result is passed to the next block, $x_{\mathrm{pre}}^{(\ell + 1, s)} = x_{\mathrm{post}}^{(\ell, s)}$. The double-block scheme is illustrated in Figures~\ref{fig:flux_double_block} and \ref{fig:flux_double_attention} in Appendix~\ref{app:flux_arch}.

Every sublayer adds its output to the residual stream, and at a fixed $t$ its AdaLN gate is constant, so the hidden state sums the contributions of all preceding sublayers. Our transcoders approximate the MLP updates, decomposing each MLP's contribution into sparse interpretable features.

\subsection{Architecture and training}
\label{sec:method:architecture}

\providecolor{ttime}{RGB}{217, 119, 6}        
\providecolor{ttimebg}{RGB}{254, 243, 199}    
\providecolor{tfilm}{RGB}{5, 150, 105}        
\providecolor{tfilmbg}{RGB}{209, 250, 229}    
\providecolor{tenc}{RGB}{37, 99, 235}         
\providecolor{tencbg}{RGB}{219, 234, 254}     
\providecolor{trelu}{RGB}{124, 58, 237}       
\providecolor{trelubg}{RGB}{237, 233, 254}    
\providecolor{tdec}{RGB}{8, 145, 178}         
\providecolor{tdecbg}{RGB}{207, 250, 254}     
\providecolor{tzdata}{RGB}{202, 138, 4}       
\providecolor{tzdatabg}{RGB}{254, 249, 195}   

\begin{figure}[htb]
\centering
\begin{tikzpicture}[
    >=Stealth,
    semithick,
    node distance=1.5em and 1em,
    op/.style={
        draw, thick,
        rounded corners=3pt,
        minimum height=2.2em,
        minimum width=2.8em,
        font=\small,
        align=center,
        inner xsep=4pt,
    },
    data/.style={
        draw=black!55,
        rounded corners=2pt,
        minimum height=1.8em,
        minimum width=1.8em,
        font=\small,
    },
    arrow/.style={->, thick},
    dim/.style={font=\footnotesize\itshape, text=black!55},
]

\node[data] (t) {$t$};
\node[op, fill=ttimebg, draw=ttime, text=ttime!45!black, right=2.5em of t]      (sinemb)  {SinEmb};
\node[op, fill=ttimebg, draw=ttime, text=ttime!45!black, right=of sinemb] (timemlp) {Time\,MLP};
\node[op, fill=ttimebg, draw=ttime, text=ttime!45!black, right=of timemlp](modlin)  {Linear};

\draw[arrow, draw=ttime!85!black] (t)       -- (sinemb);
\draw[arrow, draw=ttime!85!black] (sinemb)  -- (timemlp);
\draw[arrow, draw=ttime!85!black] (timemlp) -- (modlin);

\node[data, below=3em of t] (x) {$x$};

\node[op, fill=tfilmbg, draw=tfilm, text=tfilm!40!black, right=2.5em of x]
    (film) {FiLM};
\node[op, fill=tencbg,  draw=tenc,  text=tenc!40!black,  right=of film]
    (enc)  {$W_{\!\mathrm{enc}}$};
\node[op, fill=trelubg, draw=trelu, text=trelu!40!black, right=of enc]
    (relu) {ReLU};
\node[data, fill=tzdatabg, draw=tzdata, line width=0.7pt,
      right=of relu, minimum width=2.8em, minimum height=2.2em,
      font=\small\bfseries]
    (z)    {$z$};
\node[op, fill=tdecbg,  draw=tdec,  text=tdec!40!black,  right=of z]
    (dec)  {$W_{\!\mathrm{dec}}$};
\node[data, right=2.5em of dec] (yhat) {$\hat y$};

\draw[arrow]                       (x)    -- (film);
\draw[arrow, draw=tfilm!85!black]  (film) -- (enc);
\draw[arrow, draw=tenc!85!black]   (enc)  -- (relu);
\draw[arrow, draw=trelu!85!black]  (relu) -- (z);
\draw[arrow, draw=tdec!85!black]   (z)    -- (dec);
\draw[arrow]                       (dec)  -- (yhat);

\path (modlin.south) ++(0,-1.3em) coordinate (corner1);
\path (corner1 -| film.north) coordinate (corner2);
\draw[arrow, draw=ttime!85!black]
    (modlin.south) -- (corner1)
    -- node[pos=0.5, fill=white, inner sep=2pt,
            font=\footnotesize\itshape, text=ttime!45!black]
         {scale,\,shift}
       (corner2)
    -- (film.north);

\node[dim, above=0pt of z.north, anchor=south, text=tzdata!45!black]
    (zcap) {sparse};

\node[
    draw=black!40, dashed, line width=0.5pt,
    rounded corners=5pt,
    inner sep=7pt,
    fit=(sinemb)(timemlp)(modlin)(film)(enc)(relu)(z)(dec)(zcap),
    label={[font=\small\itshape, text=black!60]above:%
        Temporal-Aware Transcoder}
] (tcbox) {};

\end{tikzpicture}

\caption{Architecture of the Temporal-Aware Transcoder for one \mbox{(layer, stream)} pair. The timestep $t$ produces per-channel scale and shift that modulate the encoder input via FiLM; the modulated input is encoded into a wide, sparse code $z$ and decoded into the reconstruction $\hat y$ of the MLP output.}
\label{fig:tc_architecture}
\end{figure}

Where a sparse autoencoder reconstructs an activation from itself, a transcoder predicts the output of an MLP from its input. We train one transcoder per block and stream of FLUX.1[schnell], denoted $TC^s_\ell$ for $s \in \{\mathrm{img}, \mathrm{txt}\}$. The MLP has no timestep input of its own; $t$ reaches it only through its input, the AdaLN-modulated residual stream, whose distribution shifts across denoising steps. We therefore condition each transcoder on the denoising timestep $t$, as timestep-aware SAEs do~\cite{huang2026tide}, by FiLM~\cite{perez2018film} modulation of its encoder input $x \in \mathbb{R}^{d_{\mathrm{model}}}$ with an embedding $e_t \in \mathbb{R}^{d_t}$ of the timestep, $x_{\mathrm{mod}} = x \odot \bigl(1 + \mathrm{scale}(e_t)\bigr) + \mathrm{shift}(e_t)$; without it the text-stream transcoders reach twice the error or need several times as many active features (Appendix~\ref{app:tc:film}). A sparse encoder and a linear decoder then approximate the MLP output (Figure~\ref{fig:tc_architecture}):
\begin{equation}
z(x, t) = \mathrm{ReLU}\!\bigl(W_{\mathrm{enc}}\, x_{\mathrm{mod}} + b_{\mathrm{enc}}\bigr), \qquad TC^{s}_{\ell}(x, t) = W_{\mathrm{dec}}\, z(x, t) + b_{\mathrm{dec}},
\end{equation}
with $W_{\mathrm{enc}} \in \mathbb{R}^{d_{\mathrm{feat}} \times d_{\mathrm{model}}}$, $W_{\mathrm{dec}} \in \mathbb{R}^{d_{\mathrm{model}} \times d_{\mathrm{feat}}}$ and $d_{\mathrm{feat}} \gg d_{\mathrm{model}}$ (Appendix~\ref{app:tc:arch}). Feature $i$ has an encoder vector $f_{\mathrm{enc}}^{(\ell,s,i)}$, the $i$-th row of $W_{\mathrm{enc}}$, which sets its activation $z_i(x, t)$, and a decoder vector $f_{\mathrm{dec}}^{(\ell,s,i)}$, the $i$-th column of $W_{\mathrm{dec}}$; the transcoder output is the sum of the decoder vectors weighted by the activations, only a sparse subset of which is nonzero on any input.

Each transcoder minimizes the variance-normalized error $\mathrm{nMSE}^{s}_{\ell}$ plus a sparsity penalty weighted by $\lambda^s$:
\begin{equation}
\mathcal{L}^{s}_{\ell} = \frac{\mathbb{E}_{x, t}\bigl\|\mathrm{MLP}^{s}_{\ell}(x) - TC^{s}_{\ell}(x, t)\bigr\|_2^{2}}{\textstyle\sum_{j} \mathrm{Var}_{x, t}(\mathrm{MLP}^{s}_{\ell}(x)_{j}) + \varepsilon} + \lambda^s \mathbb{E}_{x, t}\bigl\|z(x, t)\bigr\|_{1}.
\label{eq:tc_objective}
\end{equation}
The normalization absorbs the order-of-magnitude spread of MLP outputs across blocks; decoder columns are renormalized to unit norm after each optimizer step (Appendix~\ref{app:transcoders}).

\subsection{Circuit tracing}
Following circuit tracing in language models~\cite{dunefsky2406transcoders,ameisen2025circuit}, we linearize feature interactions in a local replacement model (LRM), decompose the preactivation of a target feature into an attribution graph over earlier features, reconstruction errors and input embeddings, and iteratively expand and prune this graph into a compact circuit. A graph explains one transcoder feature at one layer, stream, position and denoising step, not a whole image (§\ref{sec:limit}).

\textbf{Local replacement model.} We fix a prompt, a denoising timestep $t$, and a target feature $f^*$ specified by its layer $\ell^*$, stream $s^* \in \{\mathrm{img}, \mathrm{txt}\}$, position $p^*$, and transcoder feature index $i^*$. The Local Replacement Model (LRM) is the base model with three substitutions, none of which changes its output on this input: (i) every LayerNorm recomputes its mean, which is linear in the input, but keeps its variance-based denominator at the value it takes on this input, which removes the only nonlinearity of the normalization; (ii) every joint attention block keeps its value and output projections but uses the attention probabilities $P^\ell$ of this input, so it becomes an affine map of its input, and a cached correction $\varepsilon^{\ell, s}_{\mathrm{attn}}$ absorbs the floating-point difference from the original attention output; (iii) every MLP sublayer is replaced by its transcoder $TC^s_\ell(x, t)$ plus the cached reconstruction error $\varepsilon_{\mathrm{mlp}}^{\ell,s}$.

With the set of active features $\{(\ell, s, i, p) : z_i^{(\ell, s)}(p) > 0\}$ held fixed, the LRM is an affine function of the input embeddings $r_0^s$ and of the active feature activations, so the target's preactivation $h^*$ decomposes exactly into per-source contributions plus a constant. In practice, one forward pass of the frozen base model caches everything the substitutions need (the embeddings $r_0^s$, the AdaLN modulation, which is constant for fixed $t$, the LayerNorm denominators, $P^\ell$, the transcoder activations $z^{\ell,s}$, and both errors). In the backward pass the MLP outputs are held constant, so gradients flow only through the residual stream and the frozen attention and normalization.

\begin{figure}[htb]
\centering
    \includegraphics[width=0.9\linewidth, trim=15 16 13 10, clip]{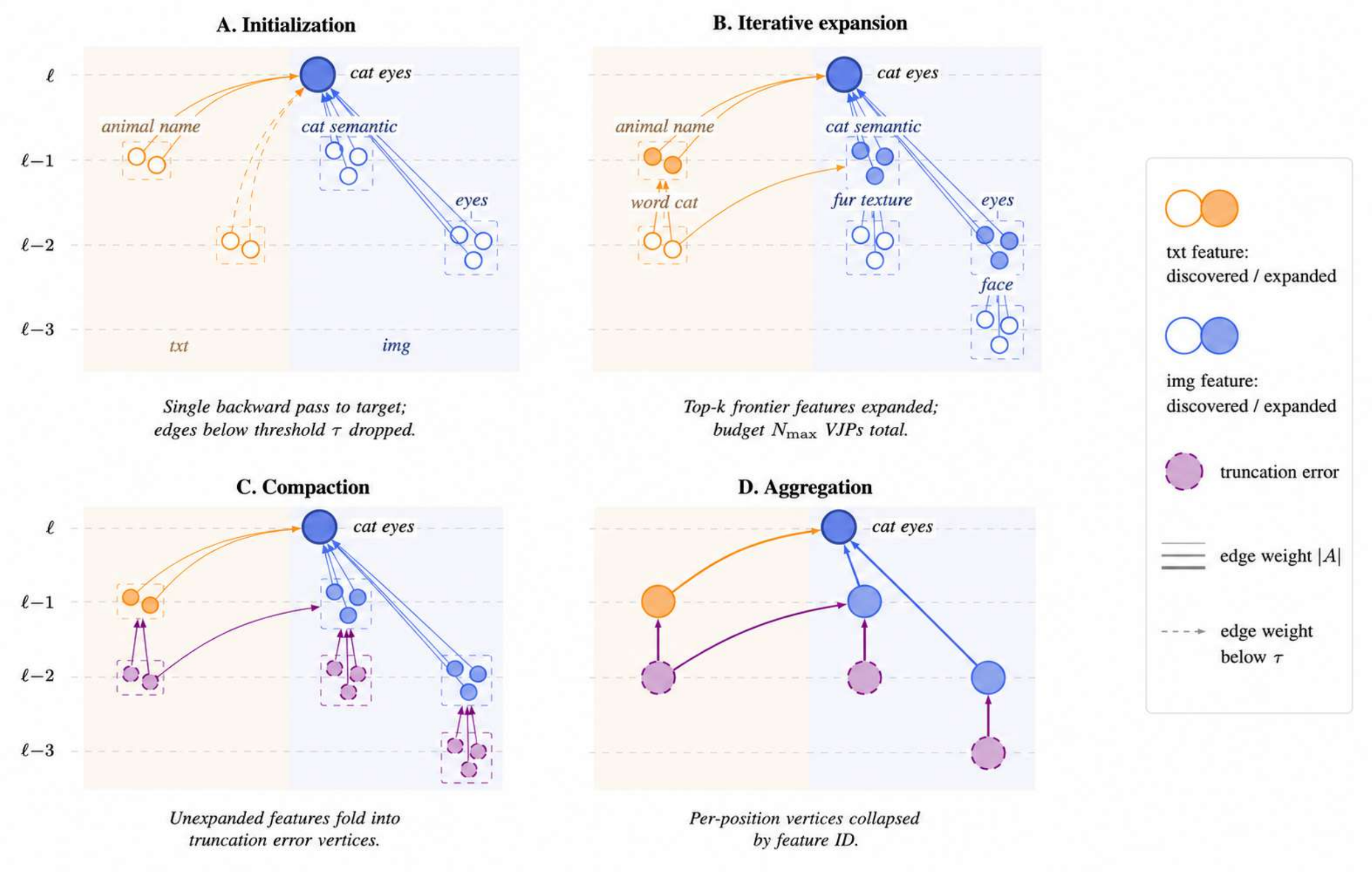}
    \caption{\textbf{Iterative graph construction and position aggregation} for a target image-stream feature \emph{cat eyes} at layer $\ell$. A cluster of small circles is one feature at several positions; outlined circles are \emph{discovered} sources, whose incoming edges are not yet computed, filled circles are \emph{expanded} ones. MLP-error and input vertices are omitted.}
    \label{fig:attribution-graph}
\end{figure}

\textbf{Attribution graph.} Given the LRM, we decompose the preactivation of the target feature $h^*$ (full derivation in Appendix~\ref{app:attribution}) rather than its activation $z^* = \mathrm{ReLU}(h^*)$, since $h^*$ is additive in its sources by linearity of the LRM, which makes the decomposition exact, and remains informative even when the feature is inactive ($h^* < 0$, $z^* = 0$). All input-independent contributions except the MLP reconstruction errors (encoder/decoder biases, AdaLN/FiLM shifts, the constant part of the frozen attention pathway, and the cached attention reconstruction terms) are collected into a target-specific scalar $b^*_{\mathrm{eff}}$ and excluded from the decomposition (Appendix~\ref{app:attribution:beff}).

The attribution graph contains a \emph{target} vertex $f^*$ and three types of source vertices: a \emph{feature} vertex for each active earlier-layer feature $(\ell, s, p, i)$ ($\ell < \ell^*$, $z^{(\ell, s, i)}(p) > 0$), an \emph{error} vertex carrying the MLP reconstruction residual $\varepsilon^{\ell,s}_{\mathrm{mlp}}(p)$, and an \emph{input} vertex carrying the model's input embeddings $r^s_0(p)$ of noisy latent patches ($s = \mathrm{img}$) or prompt tokens ($s = \mathrm{txt}$).

To compute the contribution of all vertices to the target, we run a single backward pass of $h^*$ through the LRM, denoting by $g^{\ell,s}(p)$ the gradient at the input to block $\ell$. A feature, an MLP reconstruction error and an input embedding then contribute (derivations in Appendix~\ref{app:attribution:edges})
\begin{gather}
A_{(\ell, s, p, i) \to f^*} = \underbrace{z^{(\ell, s, i)}(p)}_{\text{input-dependent}} \cdot \underbrace{\bigl(g^{\ell + 1, s}(p) \odot \mathrm{gate}^{\ell, s}_{\mathrm{mlp}}\bigr)^{\!\top} f^{(\ell, s, i)}_{\mathrm{dec}}}_{\text{virtual weight}}, \\
A_{(\ell, s, p)_{\mathrm{err}} \to f^*} = \bigl(\varepsilon^{\ell,s}_{\mathrm{mlp}}(p) \odot \mathrm{gate}^{\ell,s}_{\mathrm{mlp}}\bigr)^{\!\top} g^{\ell+1,s}(p), \qquad A_{(s,p)_{\mathrm{in}} \to f^*} = r^s_0(p)^\top g^{0,s}(p),
\end{gather}
where $\mathrm{gate}^{\ell, s}_{\mathrm{mlp}} \in \mathbb{R}^{d_{\mathrm{model}}}$ is the AdaLN-Zero MLP gate of~\eqref{eqn:flux_mlp_update}, constant across positions; the error enters through the same gate, the input embedding without one.

By construction, before any edge is dropped the attributions sum exactly to $h^* - b^*_{\mathrm{eff}} = \sum_{\mathrm{src}} A_{\mathrm{src} \to f^*}$; the residual after thresholding and pruning measures graph completeness (Appendices~\ref{app:attribution:invariant} and~\ref{app:validation:invariant}). Because frozen joint attention concatenates both streams before applying $P^\ell$, the gradient $g^{\ell+1,s}(p)$ flows across streams, producing $\mathrm{txt} \to \mathrm{img}$ and $\mathrm{img} \to \mathrm{txt}$ edges (Appendix~\ref{app:attribution:cross}).

\textbf{Iterative graph construction.} A backward pass per feature vertex is infeasible: the recursion reaches every per-position feature below the target. We therefore use a budgeted greedy procedure (Figure~\ref{fig:attribution-graph}; Appendix~\ref{app:iterative}) with at most $N_{\max} = 1{,}000$ backward passes per graph, about $40$ minutes on one H100:

\begin{enumerate}[nosep, leftmargin=*]
\item \emph{Initialize.} Compute all incoming edges to the target $f^*$ in a single backward pass; add every source with $|A| \geq \tau$ (a feature source must also have activation $z > \tau$) to the discovered set $\mathcal{D}$.

\item \emph{Score.} Give each discovered but unexpanded feature $v$ a score $\sigma(v)$: the sum of the magnitudes of its edges into already-expanded vertices, each weighted by how strongly that vertex reaches the target through the expanded part of the graph (Appendix~\ref{app:iterative:scoring}).

\item \emph{Expand.} Pick the top $k$ unexpanded features by $\sigma$, compute their incoming edges via a backward pass with the same threshold $\tau$, add their new sources to $\mathcal{D}$, and update the scores. Repeat until the budget $N_{\max}$ is exhausted (or no unexpanded feature vertex remains).

\item \emph{Compaction.} Fold edges from unexpanded features into truncation-error vertices (Appendix~\ref{app:iterative:compaction}), distinct from the MLP reconstruction errors above. This preserves the attribution sum, except for position buckets whose total falls below $\tau$, which are dropped.
\end{enumerate}

\textbf{Position aggregation.} The attribution graph is inherently per-position: features, errors and inputs appear as separate vertices at each position, producing graphs with about $10^4$ vertices for an image-stream target (median $10{,}356$, Table~\ref{tab:graph_size}). Since the question is usually \emph{which} features participate, not \emph{where}, we aggregate all position-specific vertices for the same feature $(\ell, s, i)$ into a single vertex by summing their attributions, $\bar{A}_{(\ell, s, i) \to f^*} = \sum_p A_{(\ell, s, p, i) \to f^*}$; error vertices are aggregated per $(\ell, s)$ pair, input vertices per stream. Each aggregated vertex keeps the feature's activation at every position where it fires as a sparse map, so the graph still records where each feature acts. Aggregation preserves the total attribution sum exactly but can hide cancellations between positions; it reduces the vertex count by about an order of magnitude for image-stream targets (Appendix~\ref{app:position}).

\textbf{Pruning.} After aggregation a graph typically has about $900$--$1{,}000$ vertices and $2.7$--$4.1 \times 10^{5}$ edges (Table~\ref{tab:graph_size}). First, we prune vertices (Appendix~\ref{app:pruning:vertices}) by their indirect influence on the target, the summed strength of all paths from $v$ to $f^*$: within each stream, feature vertices and the stream's input vertex are ranked together and the smallest prefix accounting for 80\% of their total influence is kept; error vertices are kept unpruned. Second, we prune edges (Appendix~\ref{app:pruning:edges}) by their normalized contribution score, retaining those that cover 98\% of the remaining influence. By default this reduces the number of vertices by about $2.4\times$ and the number of edges by about $10\times$, while increasing the mean conservation-invariant relative error by about $4$ percentage points (Tables~\ref{tab:graph_size} and~\ref{tab:invariant}).

\section{Experiments}
\label{sec:experiments}

All case studies use FLUX.1[schnell] with four denoising steps and the $32$ transcoders of layers $\ell \in \{0, \dots, 15\}$ for both streams (Appendix~\ref{app:transcoders}); the temporal and color-prior analyses are repeated on FLUX.1[dev] and SD3.5-medium (Appendix~\ref{app:backbones}). By an \textit{intervention}, we mean scaling the activation of a specific feature $z_i^{(\ell,s)}(p)$ by a scalar $\alpha$ at every position $p$ and, unless stated otherwise, every denoising timestep ($\alpha < 1$ suppresses the feature, $\alpha > 1$ amplifies it, $\alpha < 0$ inverts its contribution). The MLP sublayer that hosts the feature is then computed by its transcoder; every other sublayer runs unchanged, including the double-stream blocks $16$--$18$ and the $38$ single-stream blocks that the graphs do not model, and the effect is judged on the final image. A feature named in a case study often stands for a small supernode of two to six features that fire on the same tokens.

Each case study (i) finds a candidate feature for the phenomenon by browsing the dictionary or by contrastive prompting, (ii) computes its attribution graph on a representative prompt, (iii) groups active source features into supernodes and forms a hypothesis about the mechanism, and (iv) tests the hypothesis by interventions during generation. The search behind the case studies was exploratory: we inspected roughly $2{,}000$ features (Appendix~\ref{app:interpretation}), built about $150$ attribution graphs and inspected $50$--$100$ vertices of each, and kept a graph when it suggested a testable hypothesis, not when the intervention later succeeded. Further case studies decompose an artistic style into perceptual-linguistic primitives (Appendix~\ref{sec:exp:style}), find more localized steering targets via circuit tracing (Appendix~\ref{sec:exp:reflections}), control spatial composition (Appendix~\ref{sec:exp:spatial}), and diagnose common failure modes: color leakage (Appendix~\ref{sec:exp:color_leakage}), counting errors (Appendix~\ref{sec:exp:counting}), and negation (Appendix~\ref{sec:exp:failures}).

\paragraph{Blind evaluation.} Ten of the intervention experiments on FLUX.1[schnell] were run on prompt grids and labeled blind (Appendix~\ref{app:blind}, Table~\ref{tab:blind}): for each pair of a baseline and an intervened image of the same seed, annotators answered one question with a closed label set fixed in advance, without seeing the condition; five experiments have three annotators (majority vote, pairwise Cohen's $\kappa$ of $0.71$--$0.80$), five have one. A vision-language model relabeled all ten as a further rater and is the only rater on FLUX.1[dev] and SD3.5-medium (Appendix~\ref{app:blind:vlm}).

\subsection{Comparison with sparse autoencoders}
\label{sec:exp:sae}

Transcoders provide a capability that SAEs do not: feature-to-feature attribution through MLP sublayers, which underlies the circuit tracing of §\ref{sec:method}. We check that it does not cost sparsity or faithfulness. At three double-stream blocks of FLUX.1[schnell], early, middle and late ($\ell \in \{6, 12, 18\}$; Appendix~\ref{app:interpretation} compares their features), we train for each stream three transcoders and three SAEs with identical architectures (timestep conditioning included) and training setup (Appendix~\ref{app:transcoders}). The only difference is the objective: an SAE reconstructs the MLP output from that output, a transcoder predicts it from the MLP input, so both produce reconstructions in the same space and their errors are directly comparable. We measure sparsity by the mean $L_0$ of $z(x,t)$ and faithfulness by the nMSE of~\eqref{eq:tc_objective}, with $SAE^{s}_{\ell}(\mathrm{MLP}^{s}_{\ell}(x), t)$ in place of $TC^{s}_{\ell}(x, t)$ for an SAE. In all six configurations transcoders reach a comparable or modestly better tradeoff than SAEs at matched $L_0$ (Figure~\ref{fig:sae_comparison}), which, together with their support for circuit tracing, makes them the better basis for the analyses below.

\begin{figure}[htb]
    \centering
    \includegraphics[width=\linewidth, trim=22 24 15 16, clip]{figures/tc_vs_sae.pdf}
    \caption{Sparsity--faithfulness frontiers of transcoders and SAEs on FLUX.1[schnell], stream $\in \{\text{img}, \text{txt}\}$ $\times$ $\ell \in \{6, 12, 18\}$. Each curve: $3$ models at increasing $\lambda$; lower left is better.}
    \label{fig:sae_comparison}
\end{figure}

\subsection{Temporal evolution of attribution graphs}
\label{sec:exp:temporal}
Unlike language models, diffusion transformers apply the same network over several denoising steps, during which activation statistics change qualitatively. Does the structure of circuits change along the trajectory, and at which step should interventions be applied?

To investigate, we compute attribution graphs for 20 (prompt, target image-stream feature) pairs at each of the four denoising steps of FLUX.1[schnell], yielding 80 graphs. For each pruned graph we quantify (i) the share of absolute attribution carried by edges whose source lies in the image stream versus the text stream, summed over all edges of the graph and over all source types (features, errors and inputs), and (ii) the fraction of cross-modal edges, whose endpoints lie in different streams.

On average over the 20 targets, the text-stream share decreases monotonically from 89.9\% at step 0 to 5.4\% at step 3, while the image-stream share rises from 10.1\% to 94.6\%, and the fraction of cross-modal edges drops from 14.9\% to 2.0\% (Figure~\ref{fig:temporal}). The direction holds for every target: all 20 end with a higher image share and a lower cross-modal fraction than they start with.

\begin{figure}[htb]
    \centering
    \includegraphics[width=\linewidth, trim=16 18 16 16, clip]{figures/temporal_stats.pdf}
    \caption{\textbf{Left:} graph structure along the denoising trajectory: \textbf{(1)} share of absolute attribution by source stream; \textbf{(2)} share of cross-modal edges; error bars: $\pm 1$ s.d.\ across graphs. \textbf{Right:} median edges per pruned graph by source layer and step, from \textbf{(3)} image-stream and \textbf{(4)} text-stream sources.}
    \label{fig:temporal}
\end{figure}

\paragraph{Per-layer refinement.} The shift is not uniform across depth. Counting pruned-graph edges by source layer for image-stream targets at $\ell^* = 12$ (Figure~\ref{fig:temporal}, right), image-stream growth is concentrated at specific source layers: by step $3$, most image-stream edges come from $\ell \in \{0, 1\}$ and the mid-depth layers $\ell \in \{4, \ldots, 7\}$, while $\ell \in \{2, 3\}$ remain nearly inactive at every step. In the text stream, shallow layers contract sharply and deeper layers more slowly.

This matches a two-phase view of denoising that earlier work established without circuits: text conditioning is relied upon early and largely ignored late~\cite{balaji2022ediff}, guidance helps only in a middle range of noise levels~\cite{kynkaanniemi2024applying}, and SAE probes of a diffusion model show the composition set in the first steps and the style in the middle ones~\cite{tinaz2025emergence}; a low image-stream share at the first step is expected, as generation starts from noise. The attribution graphs recover this dynamic from the inside, stream by stream and source layer by source layer. The dynamic also holds causally (Appendix~\ref{app:exp:causal_temporal}, Table~\ref{tab:blind}): suppressing the text-stream supernode for the cat at steps $0$--$1$ removes the cat in $72.2\%$ of generations, at steps $2$--$3$ in none. Graphs at different timesteps thus capture different mechanisms, and semantic interventions work when applied early.

\paragraph{Other backbones.} We repeated the analysis on FLUX.1[dev] and SD3.5-medium: the same $20$ prompts, seeds and target layers, the target feature re-selected in each dictionary, and steps matched to the four above by noise level (Appendix~\ref{app:backbones:temporal}). FLUX.1[dev] reproduces the shift (Table~\ref{tab:backbones_temporal}): its image-stream share rises from $22.1\%$ to $93.9\%$ and its cross-modal fraction falls from $10.7\%$ to $2.2\%$. SD3.5-medium moves in the same direction but only halfway, from $8.5\%$ to $63.9\%$ image share and from $21.5\%$ to $12.4\%$ cross-modal edges: its text-driven phase lasts longer.

\subsection{Circuit-guided steering}
\label{sec:exp:steering}
Single-feature steering, the standard baseline for SAE-based control, picks features by their activation on the concept. Graphs add what this ranking cannot show: context features that shape a concept without firing on its tokens, and suppressors whose negative edges hold a competing concept down.

\paragraph{Concept vs.\ context steering.}
\label{sec:exp:concept_vs_context}
Attribution graphs expose two classes of features for intervention (Appendix~\ref{sec:exp:polysemy}). \textit{Concept} features fire directly on the tokens of the concept itself; \textit{context} features fire on semantically related but syntactically distinct tokens. For a prompt \textit{a baseball bat on a table}, the concept set is $f^{(\text{txt}, 11)}_{\text{baseball bat}}$, active on the tokens \textit{baseball bat}, together with four features of layers $7$--$9$ from its graph that fire on the same tokens; the context set is five text-stream features of layers $5$--$9$ that fire on \textit{baseball}, \textit{batter}, \textit{hand}, and \textit{glove}. Context features are the most influential sources in the attribution graph of the concept feature that do not activate on the concept tokens; both sets have five features, are read off this one graph and apply unchanged to every prompt of the grid below.

\begin{figure}[htb]
    \centering
    \includegraphics[width=0.8\linewidth]{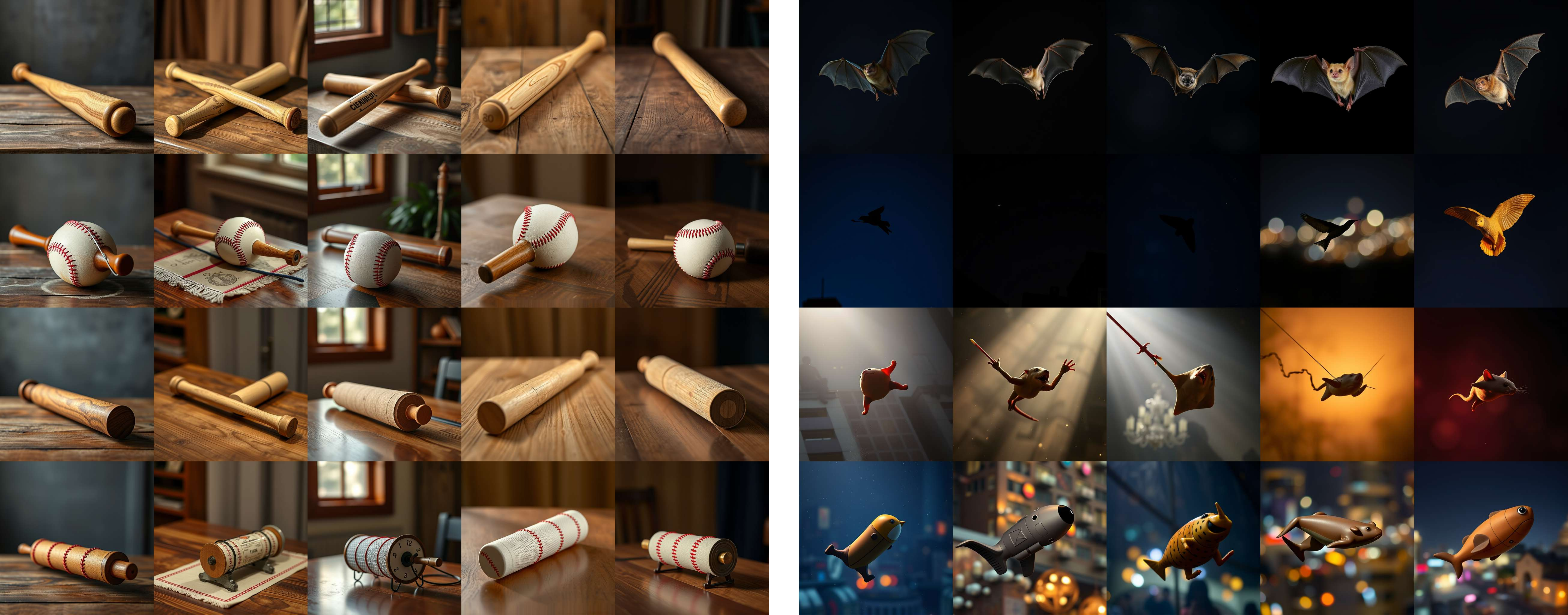}
    \caption{\textbf{Concept and context are separate channels.} Rows: baseline, then the concept features, the context features, and both suppressed; columns: seeds. \textbf{Left:} \textit{a baseball bat on a table}, $\alpha = -15$. \textbf{Right:} \textit{a bat flying at night}, $\alpha = -30$. Suppressing the concept swaps the object for a neighbor, suppressing the context keeps its shape but strips its identity, and suppressing both removes it.}
    \label{fig:concept_vs_context}
\end{figure}

Suppressing each class produces a distinct effect (Figure~\ref{fig:concept_vs_context}). Concept steering replaces the concept with a semantically nearby substitute (the bat becomes a ball, the flying animal a bird); context steering preserves the morphology but dismantles associations (the bat becomes a featureless wooden stick, the flying animal loses its wings); the combined intervention leaves nothing reminiscent of a bat or baseball. On a grid of $16$ prompts $\times$ $3$ seeds at $\alpha = -15$, labeled blind (Table~\ref{tab:blind}), the channels separate as predicted: context steering keeps the morphology while stripping the identity in $87.5\%$ of generations against $27.1\%$ for concept steering, while concept steering replaces or removes the object in $45.8\%$ against $6.3\%$ for context steering. Both together leave a recognizable object in $2.1\%$ of generations. Ranking features by activation on the concept tokens, as single-feature SAE steering does, finds the concept set; the context features do not fire on those tokens and come from the graph.

\paragraph{Suppressor features.}
\label{sec:exp:suppressor}
Attribution graphs capture not only positive but also negative connections. In the graph of $f^{(\text{img}, 12)}_{\text{cat}}
$, we identify $f^{(\text{txt}, 7)}_{\text{dog-suppressor}}$: a text-stream feature with many outgoing negative edges whose top activations occur on the token \textit{dog}. We hypothesize it actively suppresses cat features on dog prompts, keeping irrelevant cat semantics out of the generation.

\begin{figure}[htb]
    \centering
    \includegraphics[width=0.9\linewidth]{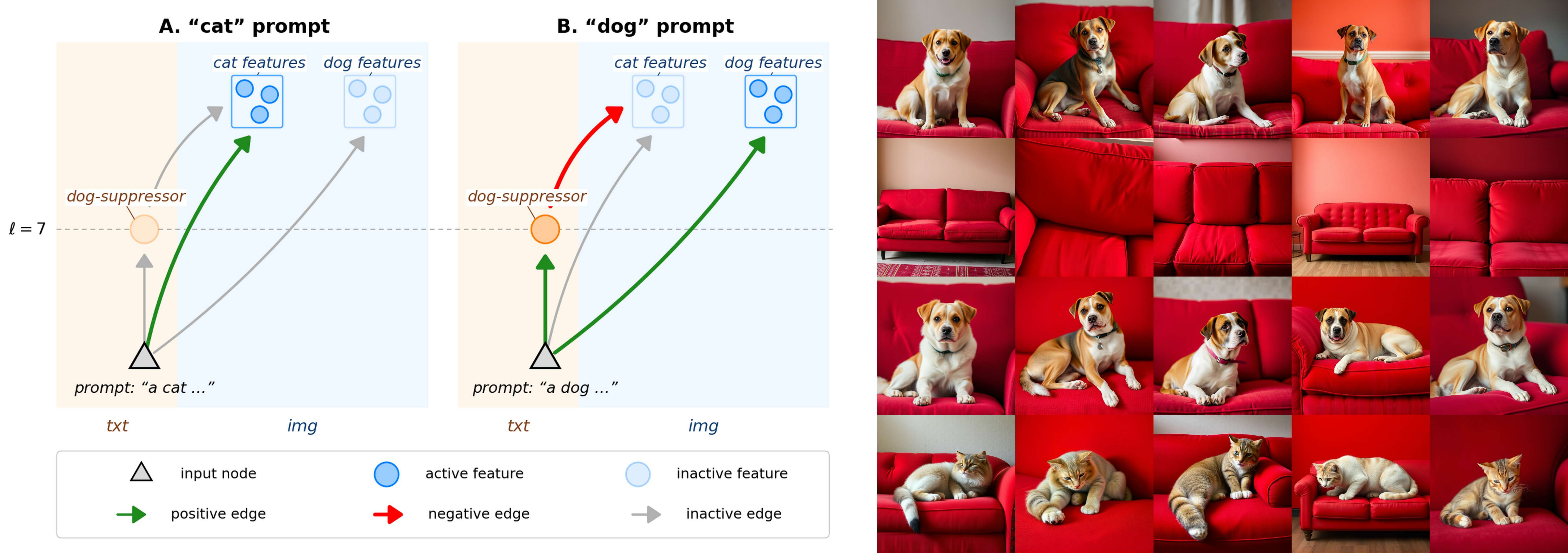}
    \caption{\textbf{A feature that suppresses a competing concept.} \textbf{Left:} on dog prompts $f^{(\text{txt}, 7)}_{\text{dog-suppressor}}$ is active and its negative edges hold cat features down; on cat prompts it is inactive. \textbf{Right:} a dog prompt; rows: baseline; dog features at $\alpha = -60$ (the dog leaves, no cat appears); the suppressor inverted alone at $\alpha = -60$ (nothing changes); dog features at $\alpha = -20$ with the suppressor at $\alpha = -30$ (a cat appears); columns: seeds.}
    \label{fig:suppressor_exp}
\end{figure}

We test this with four interventions on a grid of $12$ scenes, each prompted once with a dog and once with a cat, $\times$ $3$ seeds, labeled blind (Table~\ref{tab:blind}) and reported as the share of generations with the stated outcome (Figure~\ref{fig:suppressor_exp}): (i) suppressing cat features at $\alpha = -60$ on a cat prompt removes the cat in $88.9\%$ of generations, confirming the graph; (ii) inverting $f^{(\text{txt}, 7)}_{\text{dog-suppressor}}$ alone at $\alpha = -60$ on a dog prompt never produces a cat ($0$ of $36$), a null the graph predicts, since dog semantics is carried by other features; (iii) suppressing dog features alone at $\alpha = -60$ removes the dog in $29$ of $36$ generations but produces a cat in only one ($2.8\%$) -- switching concepts requires more than removing one pole; (iv) the joint intervention, dog features at $\alpha = -20$ together with the suppressor at $\alpha = -30$, turns the dog into a cat in $36.1\%$ of generations, an order of magnitude more often than any single channel and at a weaker suppression of the dog features. Graphs thus capture active suppression, distinct from absence of activation, and concept switching requires intervening on both what is present and what suppresses the alternative, which single-feature steering does not do. The joint condition differs from the single-channel one in both the suppressor and the strength, so it is evidence that the suppressor channel contributes rather than a fully isolated single-variable comparison.

\subsection{Color circuits}
\label{sec:exp:color}
Color is easy to verify visually, yet diffusion models fail on it systematically, through color leakage and prior bias. The image stream of FLUX contains per-color features (e.g., $f^{(\mathrm{img},10)}_{\mathrm{red}}$, active on red regions of any object) whose attribution graphs draw on three classes of text-stream sources: direct lexical color features, linguistically proximal colors, and associative features of objects with strong color priors (Appendix~\ref{app:exp:color_structure}). Here the same graph diagnoses a failure mode and lets us correct it.

\paragraph{Mitigating semantic priors via circuit intervention.}
\label{sec:exp:color_prior}
Color priors from the training data turn atypical colors (e.g., \textit{a white stop sign}, \textit{a black ladybug}, \textit{a blue pomegranate}) into the usual red (Figure~\ref{fig:atypical_colors}). The graph of $f^{(\mathrm{img},10)}_{\mathrm{red}}$ shows two competing text-stream signals: associative \texttt{red} features on the object tokens (e.g., ``stop sign'') and features of the requested color (e.g., \texttt{white}). Often the associative prior wins: the red feature's positive preactivation overrides the prompt's color.

\begin{figure}[htb]
    \centering
    \includegraphics[width=0.9\linewidth, trim=14 12 11 6, clip]{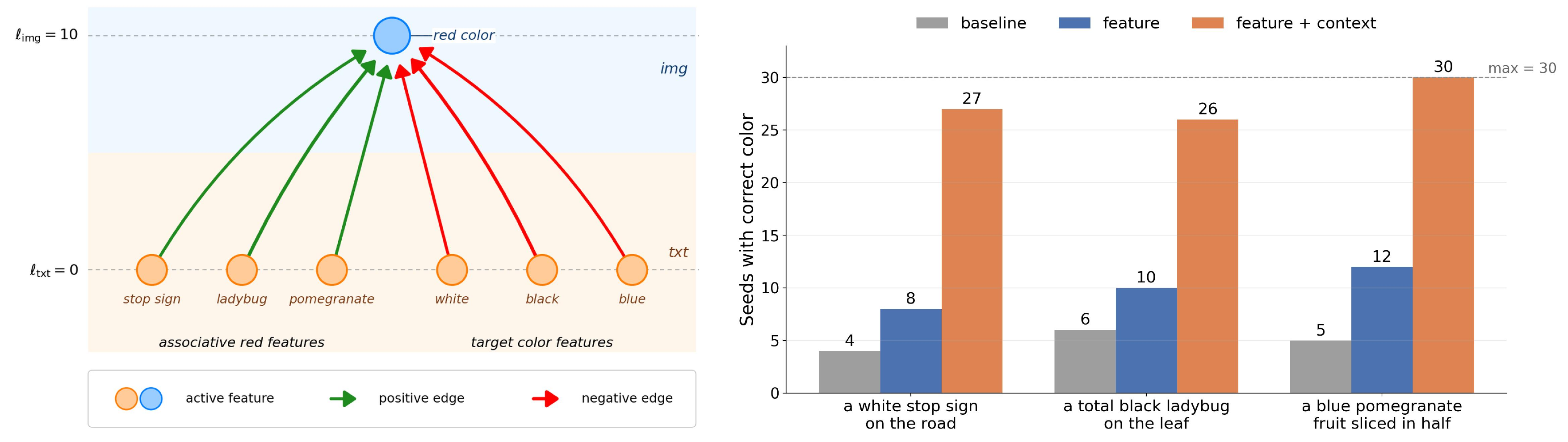}
    \caption{\textbf{Prior bias and how the graph fixes it.} \textbf{Left:} on \textit{a white stop sign}, red features on the object tokens contribute \emph{positively} to $f^{(\mathrm{img},10)}_{\mathrm{red}}$, those of the requested color \emph{negatively}; the positive side wins. \textbf{Right:} seeds out of $30$ showing the requested color (blind annotator, $\alpha = -80$): the target alone (\textit{feature}) fixes a minority, with three associative graph sources (\textit{feature + context}) $26$--$30$.}
    \label{fig:prior_bias_exp}
\end{figure}

We intervene on the graph and compare three modes on the three prompts above, $30$ seeds each, every intervention at $\alpha = -80$: \textit{baseline}, the unmodified model; \textit{feature}, suppressing $f^{(\text{img}, 10)}_{\text{red}}$ only; \textit{feature + context}, suppressing it together with three associative text-stream sources picked from its attribution graph. The circuit-wide mode dominates on all three prompts (Figure~\ref{fig:prior_bias_exp}).

\paragraph{A matched budget.} A four-feature intervention could win simply by touching more features, so we repeat the comparison at a fixed budget of four features, varying only how the three companions are chosen. The target is the same in every condition and fixed in advance; the companions come from one candidate pool, the text-stream features of layers $0$--$7$. Three rules use activations only, so an SAE could apply them: the top three by activation on the prompt tokens, by contrast against a neutral prompt, and by co-activation with the target over a $48$-prompt corpus. The fourth set is ours: the three graph sources of Figure~\ref{fig:prior_bias_exp}, picked from the graph and fixed before this comparison was run. Every pair of images in the grid of $8$ prompts $\times$ $30$ seeds $\times$ $6$ conditions was labeled blind (Appendix~\ref{app:blind}).

\begin{table}[htb]
\centering
\small
\caption{Color prior at a matched budget of four features: $8$ prompts $\times$ $30$ seeds, one blind annotator, $\alpha = -80$. Three prompts are those of Figure~\ref{fig:prior_bias_exp} ($16.7\%$ at baseline), five are new, two with weak priors ($66.7\%$ and $83.3\%$). The last four rows differ only in how the companions are chosen.}
\label{tab:matched_budget}
\begin{tabular}{lccc}
\toprule
Selection of the three companions & Requested color & Still red & Object destroyed \\
\midrule
none, baseline model & 36.2 & 62.9 & 0.8 \\
none, target feature only & 56.2 & 43.8 & 0.0 \\
top three by activation on prompt tokens & 64.2 & 35.4 & 0.4 \\
top three by contrast against a neutral prompt & 23.3 & 0.8 & 75.8 \\
top three by co-activation over a $48$-prompt corpus & 76.7 & 1.7 & 21.7 \\
\textbf{three picked from the attribution graph (ours)} & \textbf{82.1} & \textbf{4.2} & \textbf{13.8} \\
\bottomrule
\end{tabular}
\end{table}

The graph is not better than co-activation at producing the requested color (Table~\ref{tab:matched_budget}): $82.1\%$ against $76.7\%$, a difference of $5.4$ points, not statistically significant. The graph contributes selectivity: co-activation destroys the object in $21.7\%$ of generations against $13.8\%$ for the graph; a vision-language labeler sees the same direction but a smaller gap (Appendix~\ref{app:blind:vlm}). Contrast-based selection removes the red almost entirely and the object with it ($75.8\%$ destroyed); activation-based selection barely touches the object ($0.4\%$) but leaves $35.4\%$ red. Graph-guided selection thus buys precision rather than raw success, within this dictionary and at this budget. The graph set fails worst on \textit{a blue strawberry on a white plate}, destroying the object on all $30$ seeds: when the color is part of what makes the object recognizable, suppressing its associative channel removes the object too.

\paragraph{Other backbones.} We repeated the comparison on FLUX.1[dev] and SD3.5-medium on the same grid, judged by one vision-language model (Table~\ref{tab:backbones_color}). Where the baseline shows the prior, the graph set fixes it in $81.9\%$ on FLUX.1[dev] and $65.7\%$ on SD3.5-medium, against $62.7\%$ and $41.4\%$ for co-activation, and destroys the object less often ($7.2\%$ and $33.3\%$ against $31.3\%$ and $57.6\%$).

\section{Limitations}
\label{sec:limit}

\paragraph{What a graph explains.} A graph explains how the preactivation of one transcoder feature (one layer, stream, position and step; on FLUX.1[dev] also one guidance scale) is assembled from earlier features, errors and input embeddings, not an image; under classifier-free guidance, as on SD3.5-medium, it explains only the conditional half of the batch. On image-stream targets the feature vertices of a pruned graph carry $46\%$ of $h^* - b^*_{\mathrm{eff}}$, error vertices $30\%$ (reconstruction and truncation errors), input vertices $12\%$, and thresholds and pruning the rest; on text-stream targets features carry $84\%$ (Table~\ref{tab:shares}). Attention probabilities are frozen, so a graph shows where information flows through attention but not why the model attends there.

\paragraph{Why double-stream MLPs only.} MLP sublayers are the only components internal to one stream, which makes the per-stream decomposition well defined and gives text-to-image edges an unambiguous meaning; attention is kept as a frozen output-value pathway, as in transcoder circuit tracing for language models~\cite{ameisen2025circuit}; and the $38$ single-stream blocks of FLUX would need $38$ more transcoders, beyond our compute budget. Interventions are judged on the final image, so their effects also pass through blocks $16$--$18$ and the single-stream blocks, whose computation the graphs do not explain. Single-stream blocks and non-MM-DiT models are future work.

\paragraph{Misuse.} Feature-level steering is dual use: the control that fixes a color prior could as well insert or remove content against a user's or a provider's intent. The interventions need white-box access to an open-weight model, which already permits fine-tuning; we release code and transcoders for analysis.

Appendix~\ref{sec:exp:failures} shows failure cases; Appendix~\ref{app:limit} lists the limitations of our experiments: which analyses were repeated on other backbones, and how the case studies were selected and evaluated.

\section{Conclusion}
\label{sec:conclusion}
We introduced transcoders for diffusion transformers, extending circuit tracing from LLMs to MM-DiT models. On FLUX.1[schnell] their sparsity-faithfulness tradeoff is comparable to that of SAEs, and they support feature-to-feature attribution. The graphs expose structure behind color, polysemy, style and suppression, and point to interventions that single-feature steering does not reach: context features, a suppressor, and companions that fix a color prior more often than the target alone and destroy the object less often than co-activation at a matched budget. FLUX.1[dev] reproduces the temporal shift and the color-prior advantage; SD3.5-medium reproduces both in direction only.

\newpage
\bibliographystyle{plainnat}
\bibliography{bib}


\appendix

\raggedbottom

\newpage

\section{Limitations of the experiments}
\label{app:limit}

§\ref{sec:limit} discusses the limitations of the method; this section collects those of our experiments.

\paragraph{Scope.} Our case studies use the double-stream blocks of FLUX.1[schnell]. Two of the analyses, the temporal evolution of the graphs and the matched-budget color-prior comparison, were repeated on FLUX.1[dev] and SD3.5-medium (Appendix~\ref{app:backbones}); the other case studies were not. On SD3.5-medium both hold only in direction, its graphs are the least faithful of the three backbones, and a graph explains the conditional half of the classifier-free-guidance batch only. On FLUX.1[dev] the modulation also depends on the guidance embedding, so a graph holds for one timestep and one guidance scale. The color-prior comparison on the new backbones rests on labels from a vision-language model rather than from annotators; Appendix~\ref{app:blind:vlm} compares the two on FLUX.1[schnell].

\paragraph{Evaluation.} The case studies came out of an exploratory search (§\ref{sec:experiments}); their targets are not an independent sample, and the $86$ validation graphs of Appendix~\ref{app:validation} are those targets. Each intervention was run at one or two strengths, with no per-method sweep except for the spatial pair. At the matched budget we compared the graph with three activation-based selection rules within the same dictionary; we did not run SAE steering methods outside this dictionary. Five of the ten blind evaluations report one annotator's labels.

\section{Extended related work}
\label{app:related}

\subsection{Diffusion interpretability and Sparse Autoencoders}
Despite substantial advances in generation quality and efficiency through the shift from UNets~\cite{podell2023sdxl,rombach2022high} to Diffusion Transformers (DiT)~\cite{peebles2023scalable,esser2024scaling,labs2025flux}, the interpretability of diffusion models still requires extensive research. Early efforts used the U-Net bottleneck to find directions that edit attributes~\cite{kwon2022diffusion, park2023understanding} and cross-attention maps to attribute image regions to prompt words~\cite{tang2023daam}.

Sparse autoencoders (SAEs) have emerged as a popular tool for mechanistic interpretability, decomposing dense model activations into sparse, human-interpretable features. First used to interpret language models~\cite{bricken2023monosemanticity,cunningham2023sparse}, building on sparse dictionary learning of transformer embeddings~\cite{yun2021transformer}, SAEs have more recently been applied to diffusion models. Early work focused on UNet-based architectures~\cite{surkov2024one,ijishakin2024h,cywinski2025saeuron}, where they successfully identified interpretable concepts and enabled causal steering. More recent efforts extend SAEs to Diffusion Transformers (DiTs): on FLUX, SAE features are interpretable and steer generation~\cite{shabalin2025interpreting,surkov2024one}, and TIDE~\cite{huang2026tide} conditions the SAE on the timestep to account for shifting activation statistics across denoising steps.
However, because SAEs operate on activations rather than modeling the full input-output behavior of MLP sublayers, a connection between two SAE features across an MLP runs through that MLP's nonlinearity and has to be estimated input by input. As a result, SAE-based methods struggle to support fine-grained circuit tracing through the nonlinear computations inside MLP sublayers.

\subsection{Interpreting and editing diffusion models beyond SAEs}
Several works localize or control concepts in diffusion models with other tools. Revelio~\cite{kim2025revelio} probes how visual semantics is represented across the layers and denoising steps of diffusion models with k-sparse autoencoders and light-weight classifiers. Component attribution~\cite{nguyen2025unveiling} scores model components by their positive or negative contribution to a concept and edits the model by removing them. Cross-attention subspaces~\cite{bae2026mechanistic} decompose the output-value circuits of cross-attention into spectral directions that carry concepts. DiffLens~\cite{shi2025dissecting} finds sparse-autoencoder features in the U-Net bottleneck that drive social bias (gender, age, race) and intervenes on them. Unified concept editing~\cite{gandikota2024unified} edits the text projections in closed form to debias, erase styles or moderate content, a control method rather than an interpretability one. These works explain or control a concept through one kind of component or a score per component; ours decomposes the preactivation of one feature into contributions of earlier features across layers and both streams, and tests the resulting paths by intervention. On the denoising dynamics, text conditioning is known to matter mostly early~\cite{balaji2022ediff}, guidance helps only at intermediate noise levels~\cite{kynkaanniemi2024applying}, and SAE probes of Stable Diffusion show the composition set in the first steps and the style in the middle ones~\cite{tinaz2025emergence}; our temporal analysis recovers this from attribution graphs (§\ref{sec:exp:temporal}).

\subsection{Transcoders for Language Models}
Transcoders are an alternative to SAEs for interpreting MLP sublayers in LLMs that performs at least on par with them~\cite{dunefsky2406transcoders}. Rather than reconstructing activations at a single point, a transcoder approximates the entire input-output mapping of a target MLP, so on a path through MLPs alone the connection between two transcoder features is a fixed, input-invariant product of an encoder and a decoder vector; in MM-DiT the attention between them is frozen per prompt, so our graphs remain prompt-specific (Appendix~\ref{app:attribution:edges}). This opens up the possibility of tracing computational circuits at the feature level: identifying which features in earlier layers cause later features to activate, understanding how information flows across layers and components, and ultimately recovering compact, interpretable subgraphs responsible for specific model behaviors~\cite{ameisen2025circuit}.
Despite this progress in LLM, circuit-level analysis of diffusion transformers remains unexplored. We bridge this gap by introducing timestep-conditioned transcoders and a circuit tracing pipeline tailored to the MM-DiT architecture of FLUX.1[schnell].

\section{FLUX.1 double-stream block architecture}
\label{app:flux_arch}

\begin{figure}[h]
    \centering
    \includegraphics[width=0.7\linewidth]{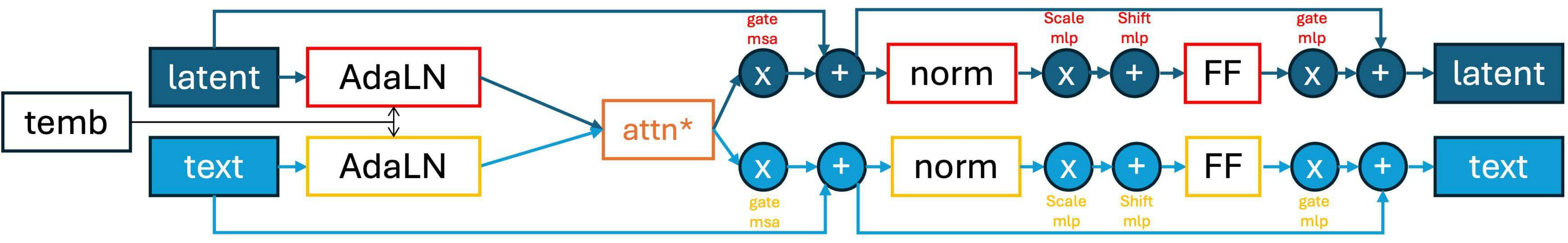}
    \caption{Schematic of a FLUX.1 double-stream block at layer $\ell$. The image and text streams are processed by stream-specific weights and interact only through the joint attention sublayer; both sublayers are wrapped by AdaLN-Zero modulation whose scale, shift, and gate parameters are produced from the denoising timestep and the pooled CLIP embedding. In the notation of Equation~\eqref{eqn:flux_mlp_update}, \emph{latent} is the image stream, the residual stream entering \emph{norm} is $x^{(\ell,s)}_{\mathrm{mid}}$, \emph{norm} followed by \emph{Scale mlp} and \emph{Shift mlp} is $\mathrm{AdaLN}_{\mathrm{mlp}}$, \emph{FF} is the MLP sublayer $\mathrm{MLP}^{(\ell,s)}$, \emph{gate mlp} is $\mathrm{gate}^{\ell,s}_{\mathrm{mlp}}$, and the stream leaving the last addition is $x^{(\ell,s)}_{\mathrm{post}}$.}
    \label{fig:flux_double_block}
\end{figure}

\begin{figure}[h]
    \centering
    \includegraphics[width=0.7\linewidth]{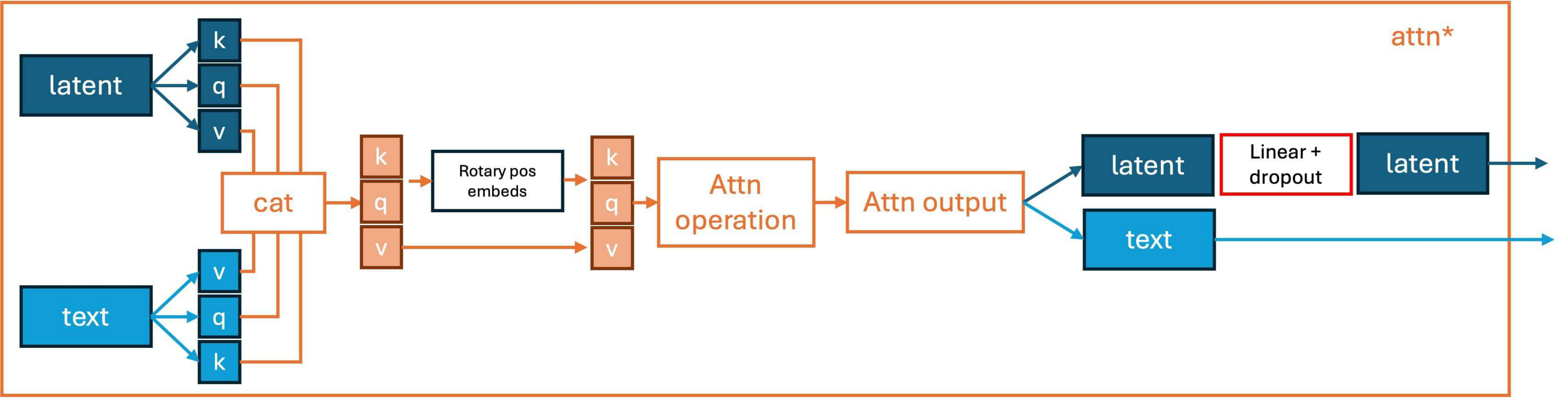}
    \caption{Joint attention sublayer of a FLUX.1 double-stream block. Queries, keys, and values are projected per stream from the AdaLN-modulated inputs, concatenated along the token axis, passed through a single scaled dot-product attention, and split back into per-stream outputs that are added to their respective residual streams via the AdaLN-Zero gate.}
    \label{fig:flux_double_attention}
\end{figure}

For visual reference accompanying the textual description in §\ref{sec:prelim}, Figure~\ref{fig:flux_double_block} shows the overall structure of a FLUX.1[schnell] double-stream block, and Figure~\ref{fig:flux_double_attention} details its joint attention sublayer. Both diagrams are adapted from~\cite{greenberg2025demystifyingfluxarchitecture}.

\section{Additional experiments}
\subsection{Additional evidence for the two-phase interpretation}
\label{app:exp:causal_temporal}

\paragraph{Qualitative graph evolution.}
Figure~\ref{fig:temporal_graph_grid} visualizes the structural shift documented quantitatively in §\ref{sec:exp:temporal} on a single (prompt, target feature) pair. The four panels show the attribution graph for the same target at each of the four denoising steps. At $t = 0$, the text-stream half of the graph is densely populated and connected to the target through numerous cross-modal edges; the image-stream half is sparse. As the trajectory proceeds the text-stream side contracts and cross-modal connectivity drops, while the image-stream side grows progressively richer. The same pattern holds visually across the prompts and target features we inspected, mirroring the aggregate trend of Figure~\ref{fig:temporal}.

\begin{figure}[h]
    \centering
    \includegraphics[width=\linewidth]{figures/temporal_graph_grid.pdf}
    \caption{Attribution graphs for a single (prompt, target feature) pair at each of the four denoising steps of FLUX.1[schnell]. Panels left to right: $t = 0, 1, 2, 3$. Feature nodes are colored by stream (image: blue, text: orange) and edges by attribution sign (positive: blue, negative: red); error nodes appear as red diamonds and input nodes as purple circles.}
    \label{fig:temporal_graph_grid}
\end{figure}

\begin{figure}[h]
    \centering
    \includegraphics[width=\linewidth]{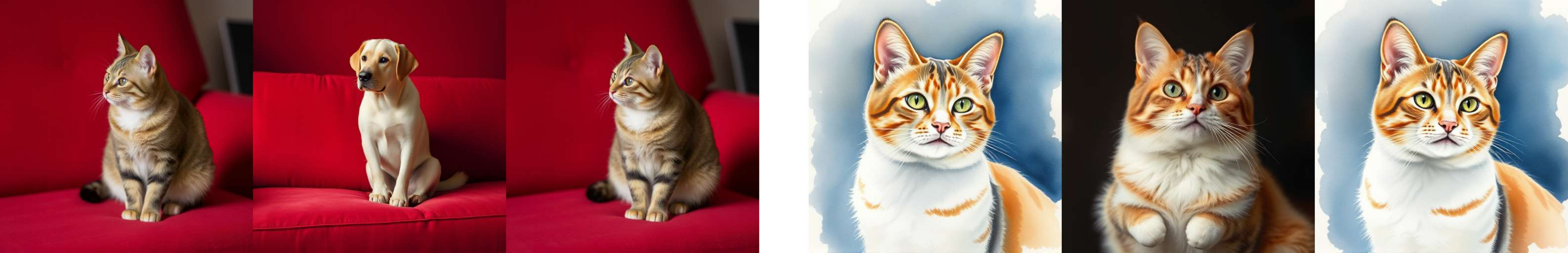}
    \caption{\textbf{Text-stream interventions work only early.} Each panel shows one prompt under three conditions, left to right: the original generation, suppression of the indicated text-stream supernode at $t \in \{0, 1\}$, and the same suppression at $t \in \{2, 3\}$. \textbf{Left:} \texttt{A cat sitting on a red couch}, the \texttt{cat} supernode at $\alpha = -30$; under early suppression the cat is removed, under late suppression it is untouched. \textbf{Right:} \texttt{A cat, watercolor painting}, the \texttt{watercolor} supernode; early suppression removes the style, late suppression does not. Over $12$ cat prompts $\times$ $3$ seeds the cat is removed in $72.2\%$ of early-suppression generations and in $0$ of $36$ late ones.}
    \label{fig:stepwise_intervention_grid}
\end{figure}

\paragraph{Causal validation.}
The structural shift documented in §\ref{sec:exp:temporal} predicts that interventions on text-stream features should be effective only at early denoising steps. To test this, we suppress the text-stream supernode that encodes \texttt{cat} at $\alpha = -30$, either at $t \in \{0, 1\}$ or at $t \in \{2, 3\}$, leaving its features untouched at the other steps, on a grid of $12$ cat prompts $\times$ $3$ seeds that was labeled blind (Appendix~\ref{app:blind}). At the same strength, early suppression removes the cat in $72.2\%$ of generations and late suppression in none ($0$ of $36$). Figure~\ref{fig:stepwise_intervention_grid} shows the effect on one prompt, together with a stylistic supernode: for \texttt{A cat, watercolor painting}, suppressing the \texttt{watercolor} supernode at early steps eliminates the style, while late suppression leaves the image visually identical to the baseline.

\subsection{Polysemy and contextual disambiguation}
\label{sec:exp:polysemy}

Transcoder features should ideally capture a single concept. We tested sense separation using the polysemous token \texttt{bat}. Attribution graphs for $f^{(\text{txt}, 11)}_{\text{baseball bat}}$ confirm that the model recruits qualitatively different source features depending on context. The animal-context graph is dominated by wings, Batman, and darkness features, while the baseball-context graph activates sport and equipment features. In ambiguous cases (e.g., \textit{``a bat''}), the graph reveals simultaneous activation of animal, baseball, and ``party'' senses, yet the model produces a flying bat in 5 out of 5 seeds.

\begin{figure}[h]
    \centering
    \includegraphics[width=\linewidth]{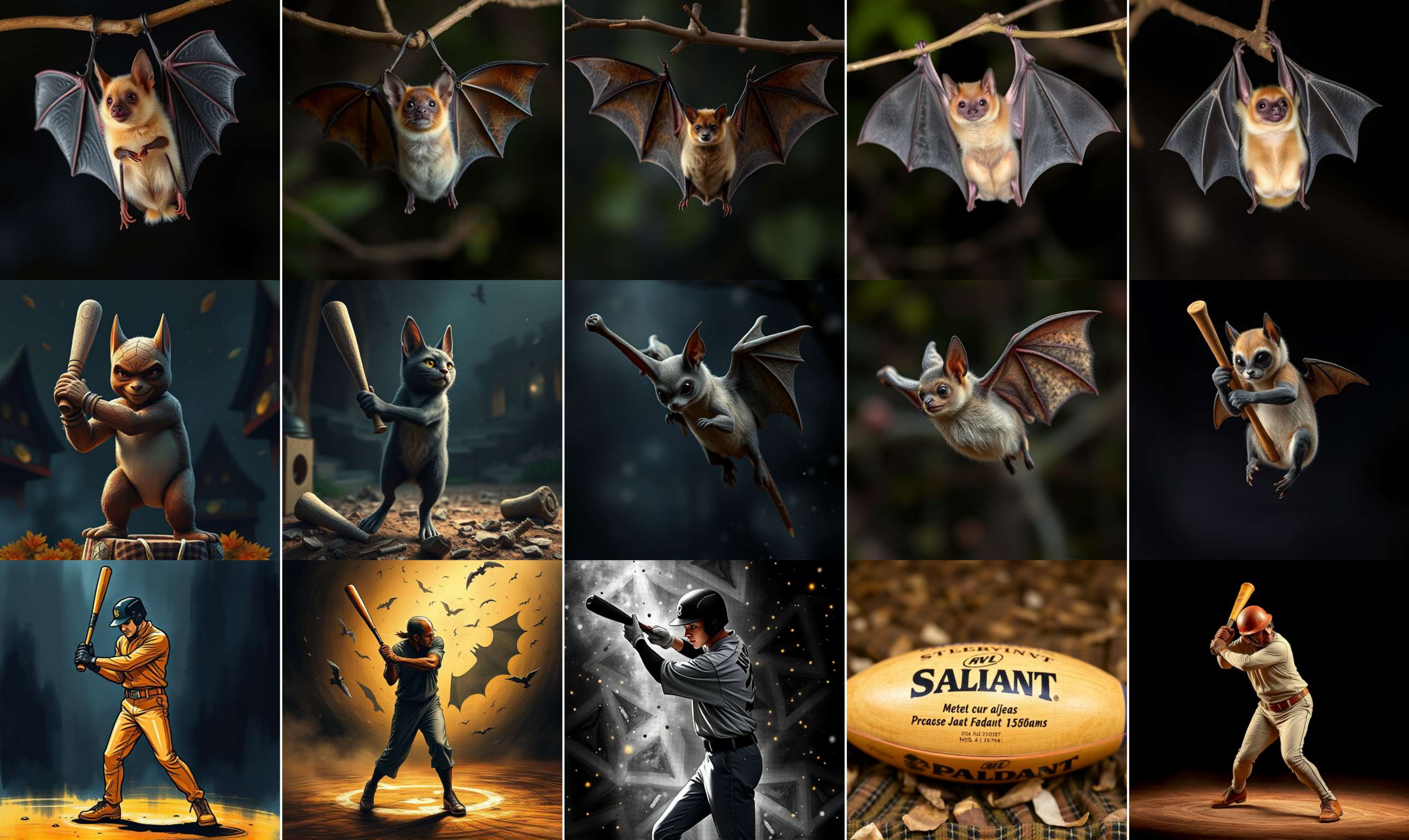}
    \caption{Steering toward the weaker sense of \textit{bat}. Prompt \textit{a bat}; the five baseball-context text-stream features of §\ref{sec:exp:concept_vs_context} are amplified together. Rows: baseline, a flying bat on every seed; $\alpha = +30$, where a baseball bat appears alongside the animal; $\alpha = +100$, where a baseball player replaces the animal on most seeds. Columns: seeds.}
    \label{fig:bat_steering}
\end{figure}

This discrepancy reveals a key insight: even when the visual output is biased toward one sense, the attribution graph for the ambiguous case contains baseball-related features with positive attribution. By selectively \textit{amplifying} contextual text-stream nodes (e.g., \textit{batter}, \textit{glove}) rather than suppressing the dominant sense, we can steer the output without explicit suppression. The effect depends on steering strength (Fig.~\ref{fig:bat_steering}): at intermediate $\alpha$, a baseball bat appears alongside the animal; at larger $\alpha$, the baseball player supersedes the animal entirely. On a grid of $10$ prompts $\times$ $3$ seeds, labeled blind (Appendix~\ref{app:blind}), the image shows the baseball reading in $33\%$ of generations at $\alpha = +30$ and in $70\%$ at $\alpha = +100$.

\subsection{Structure of a color circuit}
\label{app:exp:color_structure}

Color-related features in the image stream form a predictable and interpretable circuit. We identify a specific red-sensitive feature, $f^{(\mathrm{img},10)}_{\mathrm{red}}$, which activates in response to red regions regardless of the generated object. The attribution graph for this feature remains stable across various prompts: the majority of the attribution flows from the text stream through three distinct channels: (i) direct lexical color features (e.g., \textit{red}), (ii) linguistically proximal color features (e.g., \textit{orange}, \textit{purple}), and (iii) associative features linked to red objects (e.g., \textit{tomatoes}, \textit{Canadian flag}).

\begin{figure}[h]
    \centering
    \includegraphics[width=0.75\linewidth]{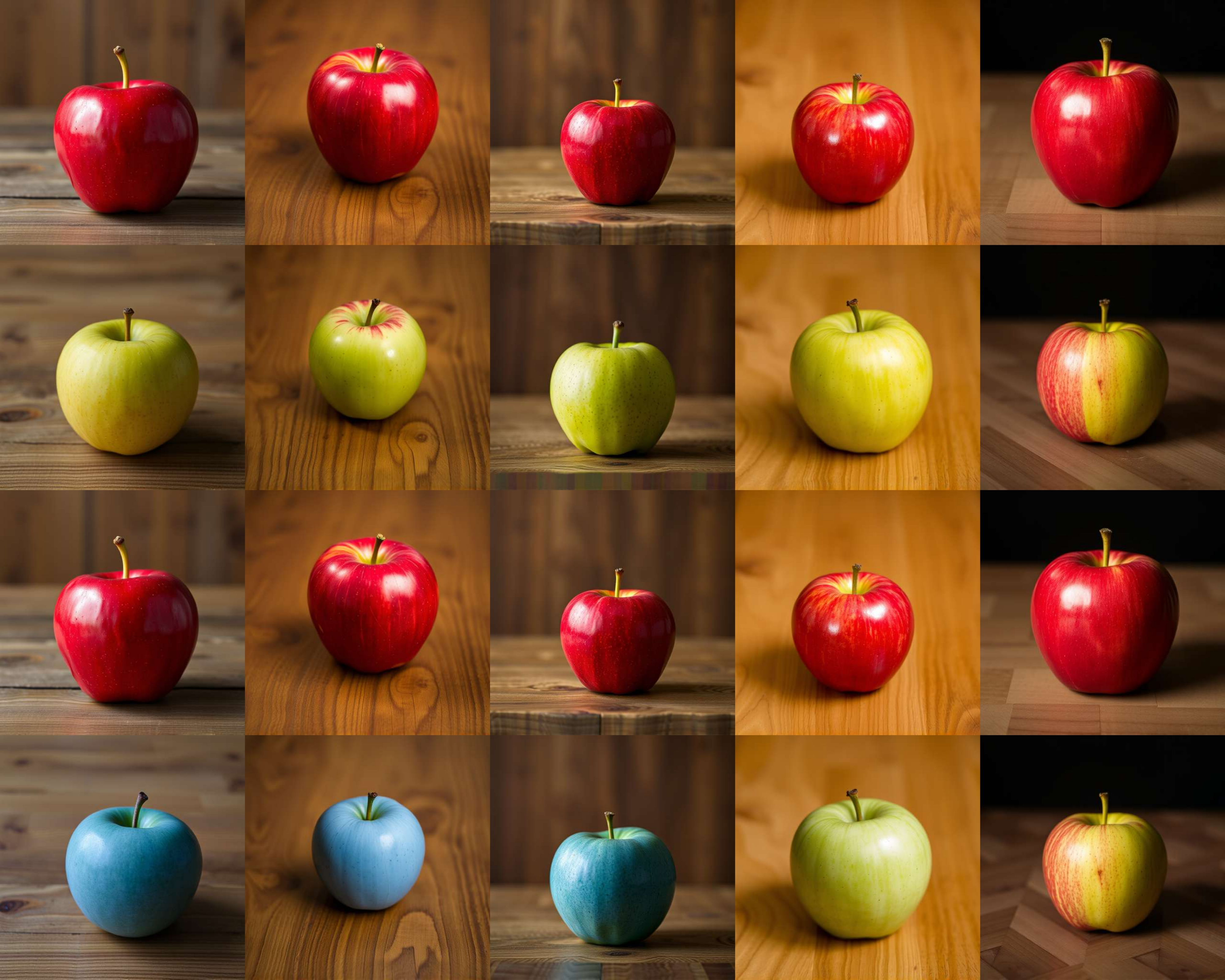}
    \caption{Overriding an explicit color token. Prompt \textit{a red apple on a wooden table}. Rows: baseline; red features suppressed, which mostly yields a green apple; blue features amplified, with no visible effect; both interventions jointly, which yields a blue apple on $3$ of $5$ seeds. Steering strength $|\alpha| = 15$ throughout. Columns: seeds.}
    \label{fig:color_apple}
\end{figure}

While the image-stream representation is largely prompt-invariant, text-stream contributions adapt to context (e.g., the prompt \textit{a red dress} additionally activates a feature for \textit{pink}, while \textit{a red sunset} activates one for \textit{orange}). Intervention experiments on the prompt \texttt{a red apple on a wooden table} (Fig.~\ref{fig:color_apple}) reveal a clear functional asymmetry: suppressing the red color features shifts the apple to the model's natural green prior, while amplifying competing blue color features in isolation has no visible effect. Only joint suppression of red and amplification of blue produces a blue apple, on $3$ of $5$ seeds; this example was not evaluated on a larger grid. This suggests that explicit color tokens in a prompt create a robust activation that must be actively suppressed to overcome the model's internal state.

\subsection{Prior bias mitigation: qualitative examples}
\label{app:exp:prior_bias_qualitative}

The bar chart of Fig.~\ref{fig:prior_bias_exp} reports aggregate success rates but conceals what the failures and successes look like. Figure~\ref{fig:atypical_colors} shows representative generations for the three prompts of §\ref{sec:exp:color_prior} under all three intervention modes. The \textit{baseline} mode overrides the explicit color token and renders the object red on all three prompts. The \textit{feature} mode succeeds on a minority of seeds: suppressing $f^{(\text{img}, 10)}_{\text{red}}$ alone partially weakens the red feature, but the associative prior carried by the object-token features keeps pushing it back up, so most generations remain incorrect. The \textit{feature + context} mode -- which additionally suppresses the associative red sources from the graph -- reliably produces the requested color across seeds.

\begin{figure}[ht]
    \centering
    \includegraphics[width=\linewidth]{figures/atypical_colors.pdf}
    \caption{Qualitative results for prior bias mitigation (§\ref{sec:exp:color_prior}). \textbf{Left:} \textit{a white stop sign on the road}. \textbf{Mid:} \textit{a total black ladybug on the leaf}. \textbf{Right:} \textit{a blue pomegranate fruit sliced in half}. Rows: \textit{baseline}, no intervention; \textit{feature}, $f^{(\text{img}, 10)}_{\text{red}}$ suppressed; \textit{feature + context}, the same feature and three associative sources from its graph suppressed ($\alpha = -80$). Only the last row shows the requested color on most seeds of all three prompts. Columns: five seeds; the rates over $30$ seeds are in Figure~\ref{fig:prior_bias_exp}.}
    \label{fig:atypical_colors}
\end{figure}

\subsection{Style decomposition: watercolor}
\label{sec:exp:style}

Style concepts are useful for interpretability because  they should be content-invariant: a feature representing style $X$ should activate on images in style $X$ regardless of subject matter. Whether such a feature exists as an atomic representation or as a composition of simpler primitives can be found out by examining its attribution graph.

Using contrastive prompts we identify a watercolor-style feature $f^{(\text{img}, 11)}_{\text{watercolor}}$. Its graph contains no nodes related to the depicted object -- confirming the feature's content-invariance. Instead, the graph decomposes into five stylistic components across the two streams: three image-stream features at layers $8$, $9$ and $10$, among them features for clouds and steam and for bright multicolored imagery (flags, colored pencils); $f^{(\text{txt}, 7)}_{\text{light-haze}}$ for tokens such as \textit{light haze} and \textit{smoke}; and $f^{(\text{txt}, 6)}_{\text{pastel}}$ for constructions of the form \textit{pastel-colored} or \textit{lavender-colored}.

\begin{figure}[h]
    \centering
    \includegraphics[width=\linewidth]{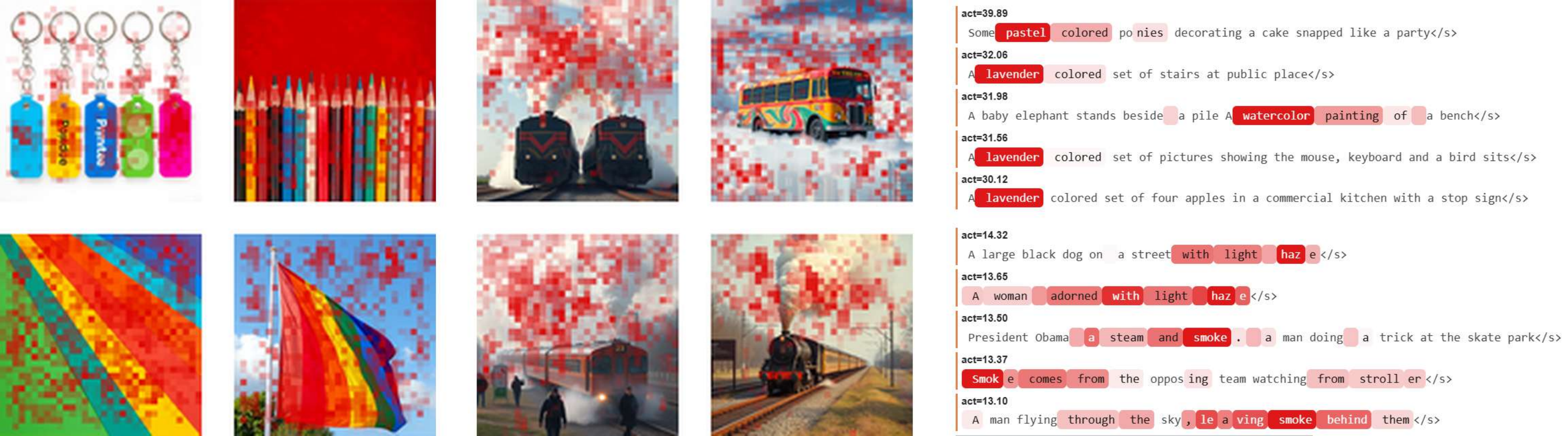}
    \caption{Top-activating examples for four of the five component features of $f^{(\text{img},\,11)}_{\text{watercolor}}$. Two of the image-stream features are shown side by side on the left: the multicolor and the steam feature; each panel is a $2\times2$ tile of top-activating images with top-activating patches highlighted. The two text-stream features are stacked on the right: $f^{(\text{txt},\,6)}_{\text{pastel}}$ (top) and $f^{(\text{txt},\,7)}_{\text{light-haze}}$ (bottom); each panel lists the top-activating prompts with token-level activations highlighted.}
    \label{fig:watercolor_gallery}
\end{figure}

\begin{figure}[h]
    \centering
    \includegraphics[width=\linewidth]{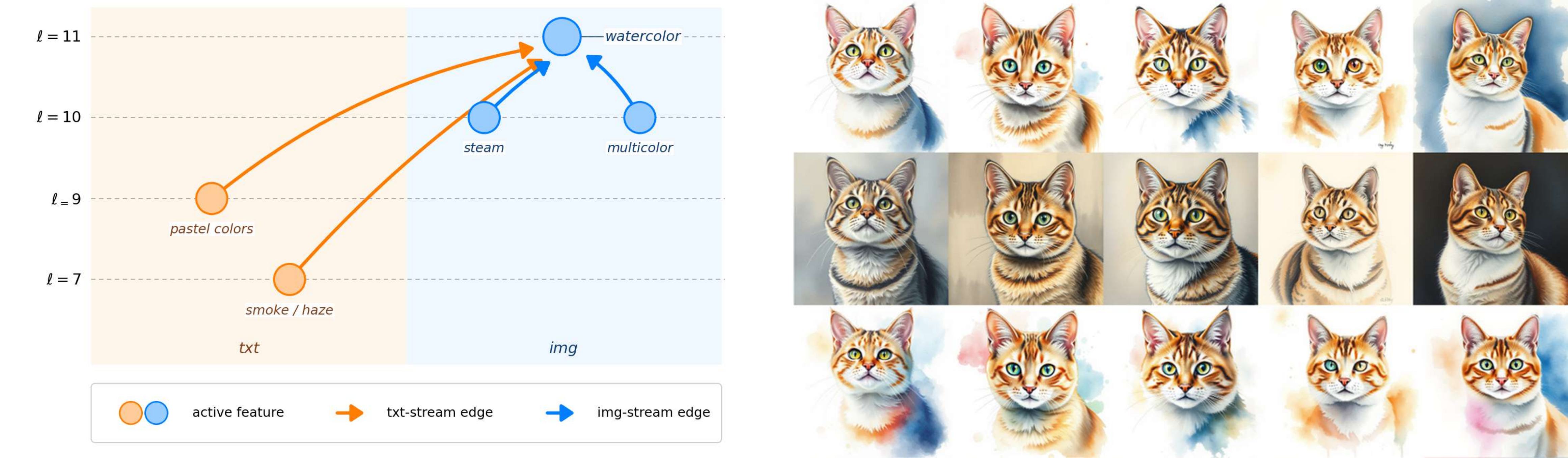}
    \caption{Style decomposition of $f^{(\text{img}, 11)}_{\text{watercolor}}$. \textbf{Left:} its attribution graph reduced to the style components named in the text. \textbf{Right:} prompt \textit{a cat, watercolor painting}, with the component features steered jointly. Rows: baseline; $\alpha = -15$, where the style weakens into a more realistic portrait; $\alpha = +15$, where it strengthens into looser washes and color spills. The subject is unchanged in both. Columns: seeds.}
    \label{fig:watercolor_steering}
\end{figure}

These components substantively cover what natural language descriptions of watercolor typically include: \textit{a pastel palette}, \textit{soft hazy edges}, and diverse color choices. Consequently, the model represents style not as a monolithic atomic feature, but as a structured composition of fundamental perceptual-linguistic primitives. Interventions on these five components monotonically strengthen or weaken the resulting style (Fig.~\ref{fig:watercolor_steering}). Their joint shift produces a clean control of style without affecting the semantic content of the scene. Style has no binary success criterion, so this case study was not part of the blind evaluation of Appendix~\ref{app:blind} and is presented as an illustration of what the decomposition buys, not as a measured effect.

\subsection{Circuit-guided feature discovery: reflections}
\label{sec:exp:reflections}

In the dictionary of the image-stream transcoder at $\ell = 12$, we identify a feature $f^{(\text{img}, 12)}_{\text{refl}}$ that activates on reflections of objects in water, mirrors, and other reflective surfaces, but not on the objects themselves. The feature is robust and an attractive candidate for steering; the question is whether it captures the model's representation of the concept of reflection as such, or whether it is merely one of several components into which the model decomposes that concept.

The attribution graph for $f^{(\text{img}, 12)}_{\text{refl}}$, computed across several reflection prompts, consistently contains the same source feature $f^{(\text{img}, 11)}_{\text{refl}}$ with attribution an order of magnitude larger than any other source. The activation map of $f^{(\text{img}, 11)}_{\text{refl}}$ qualitatively matches that of $f^{(\text{img}, 12)}_{\text{refl}}$. Together -- earlier layer plus dominant role in the graph of the later target -- these facts suggest the hypothesis that $f^{(\text{img}, 11)}_{\text{refl}}$ represents the concept of reflection closer to its origin within the model, while $f^{(\text{img}, 12)}_{\text{refl}}$ is a downstream derivative localized to a later layer.

\begin{figure}[h]
    \centering
    \includegraphics[width=\linewidth]{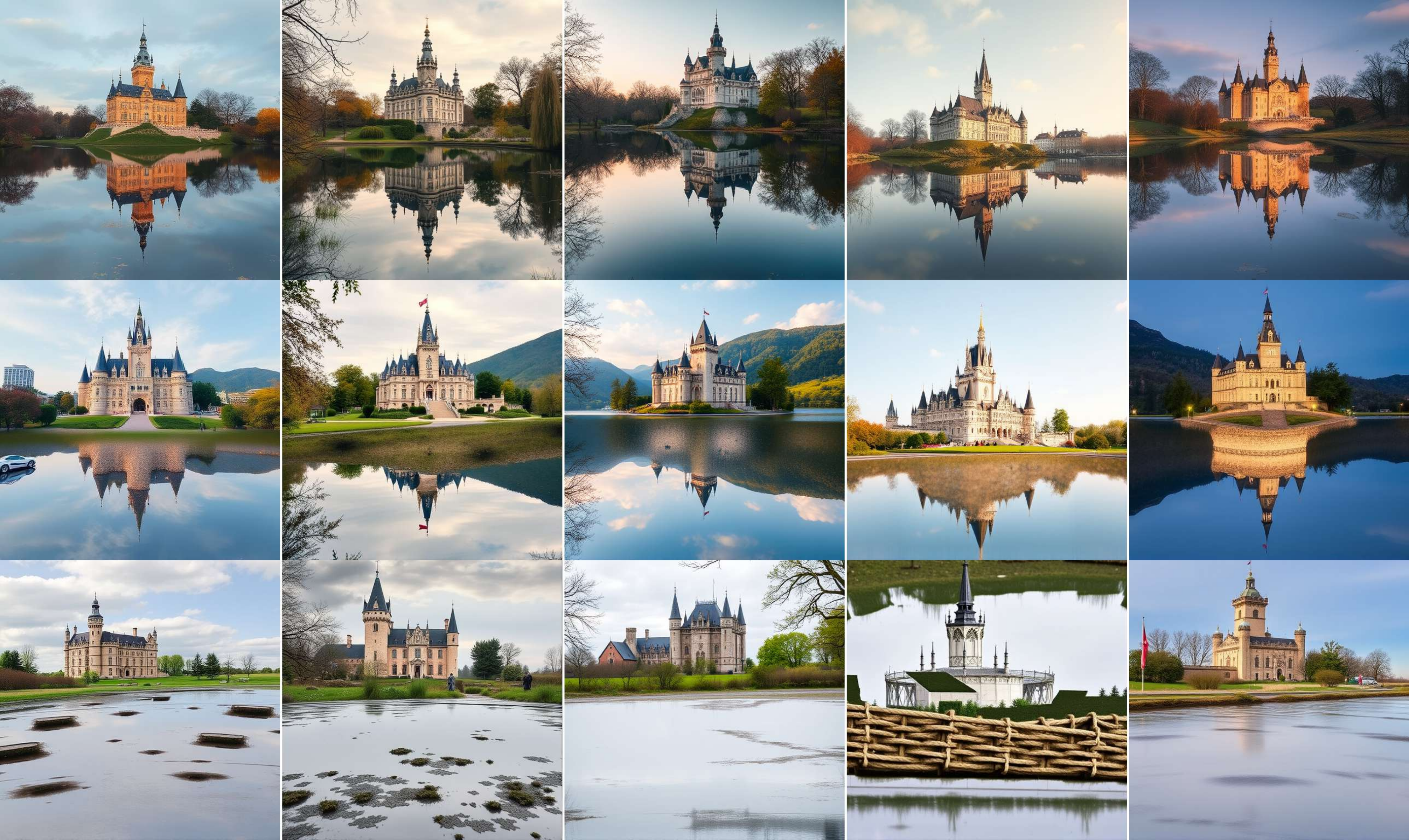}
    \caption{Suppressing the two reflection features, prompt \textit{a castle reflected in a calm pond}. Rows: baseline; $f^{(\text{img}, 11)}_{\text{refl}}$ at $\alpha = -30$, where the reflection becomes a blurred silhouette and the water surface stays intact; $f^{(\text{img}, 12)}_{\text{refl}}$ at $\alpha = -30$, where the reflection disappears and the surface is deformed. Columns: seeds.}
    \label{fig:refl_features}
\end{figure}

If the hypothesis is correct, the later feature should be entangled not only with the reflection itself but also with the surrounding perceptual context, whereas the earlier feature should be tied to a narrower concept. The intervention comparison (Fig.~\ref{fig:refl_features}) is consistent with this idea. Suppressing $f^{(\text{img}, 12)}_{\text{refl}}$ removes the reflection but deforms the reflective surface, introducing visual artifacts. Suppressing $f^{(\text{img}, 11)}_{\text{refl}}$ leaves the surface intact, while the reflection of the object turns into a blurred patch. The contrast holds on a grid of $17$ prompts $\times$ $5$ seeds $\times$ $3$ conditions at $\alpha = -30$, labeled blind (Appendix~\ref{app:blind}): among the generations in which the reflection is removed or noticeably weakened, the surface stays intact in $38$ of $44$ ($86.4\%$) for the layer-$11$ feature, against $18$ of $48$ ($37.5\%$) for the layer-$12$ feature. As a control, suppressing at the same strength two layer-$11$ features that the graph ranks far below the dominant source (direct attribution about $1\%$ of it) leaves all $85$ generations unchanged. Circuit tracing thus enables the selection of a feature for steering that satisfies a stricter locality criterion than candidates accessible via feature interpretation alone.

\subsection{Spatial composition}
\label{sec:exp:spatial}

Spatial composition of the scene is a known weakness of text-to-image models. We investigate how spatial understanding is encoded within the model and demonstrate how this internal logic can be leveraged to achieve control over object positioning.

In the text stream around $\ell = 7$ we identify location features $f^{(\text{txt}, 7)}_{\text{left}}$ and $f^{(\text{txt}, 7)}_{\text{right}}$ that fire on their respective tokens regardless of which object the location is bound to. On the prompts \textit{a red house on the left and a blue car on the right} and \textit{a blue car on the left and a red house on the right}, the attribution graph for $f^{(\text{img}, 9)}_{\text{red}}$ contains, respectively, $f^{(\text{txt}, 7)}_{\text{left}}$ and $f^{(\text{txt}, 7)}_{\text{right}}$ -- that is, which spatial token enters the graph is determined by which object it is assigned to in the prompt (Figure~\ref{fig:position_maps} shows the tokens on which these features fire). At the same time, we did not find clearly interpretable spatial features in the image stream.

\begin{figure}[h]
    \centering
    \includegraphics[width=0.8\linewidth]{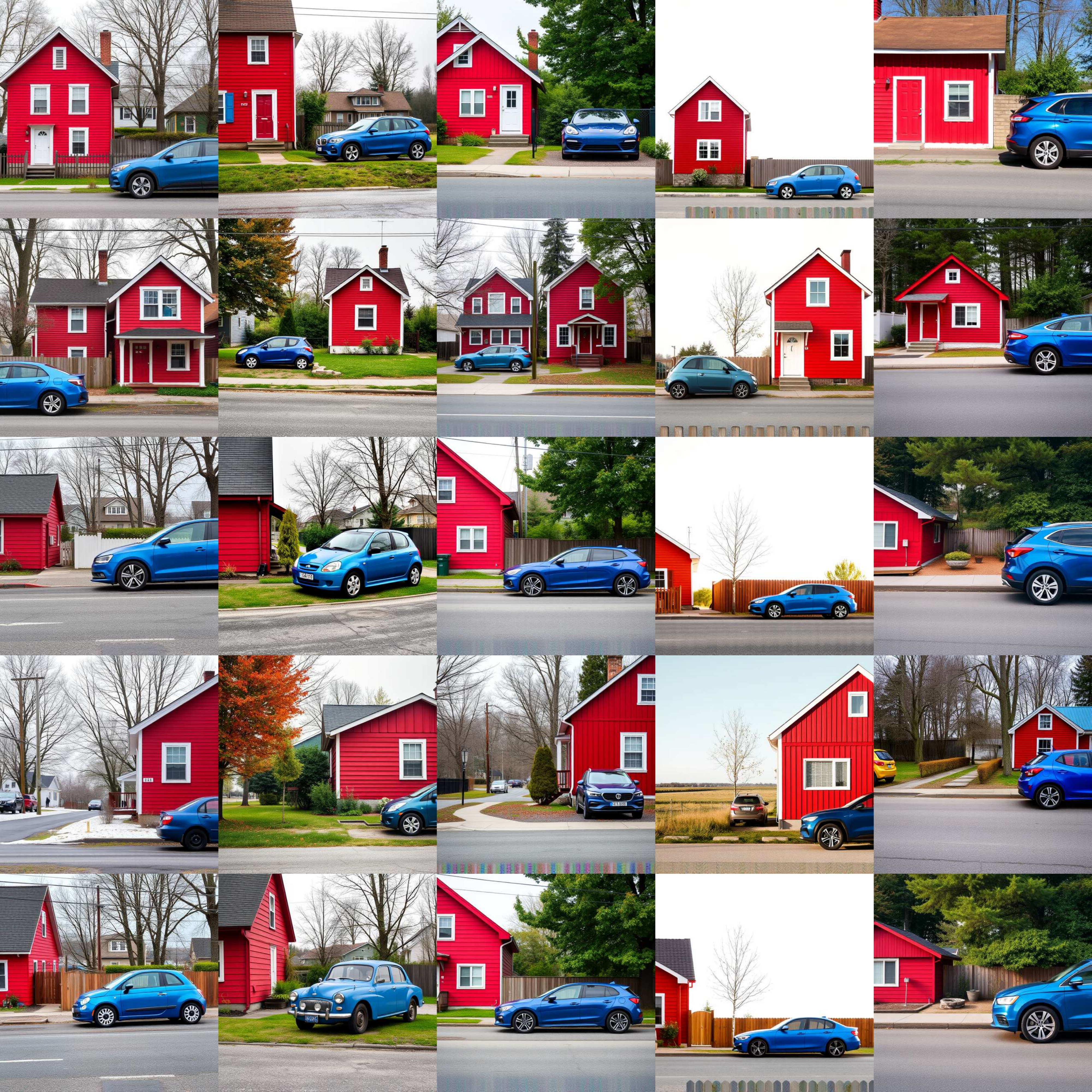}
    \caption{Prompt \textit{a red house on the left, a blue car on the right}. Steering grid, rows: baseline; $f^{(\text{txt}, 7)}_{\text{left}} \to -30$ (composition is mirrored); $f^{(\text{txt}, 7)}_{\text{left}} \to +30$ (the house slides off the left edge of the image); $f^{(\text{txt}, 7)}_{\text{right}} \to +10$, $f^{(\text{txt}, 7)}_{\text{left}} \to -10$ (both objects on the right); $f^{(\text{txt}, 7)}_{\text{right}} \to -10$, $f^{(\text{txt}, 7)}_{\text{left}} \to +10$ (no composition changes). Columns: seeds.}
    \label{fig:spatial}
\end{figure}

The interventions (Fig.~\ref{fig:spatial}) are consistent with this text-stream localization. On a grid of $16$ prompts $\times$ $5$ seeds, labeled blind (Appendix~\ref{app:blind}), the pair $f^{(\text{txt}, 7)}_{\text{right}} \to +\alpha$, $f^{(\text{txt}, 7)}_{\text{left}} \to -\alpha$ at $\alpha = 10$ puts the red object in the right half of the image in $85\%$ of generations; suppressing \textit{left} alone does so in $39\%$ at the same strength and needs $\alpha = 30$ to reach $78\%$. The symmetric pair $f^{(\text{txt}, 7)}_{\text{right}} \to -\alpha$, $f^{(\text{txt}, 7)}_{\text{left}} \to +\alpha$ leaves the composition unchanged in $97.5\%$ of generations, as the graph predicts, since the object is already on the left. A sweep of the pair over $\alpha \in \{3, 10, 30, 100\}$ bounds the usable range: at $\alpha = 3$ the composition is unchanged in $98.8\%$ of generations, while at $\alpha = 30$ the image already degrades beyond recognition in $59$ of $80$ and at $\alpha = 100$ in $74$ of $80$.

\subsection{Color leakage}
\label{sec:exp:color_leakage}

Color leakage -- the failure of text-to-image models to bind colors correctly to objects on prompts with multiple colored objects -- is a standard pathology of generative diffusion models. If the structure of the color circuit established in §\ref{app:exp:color_structure} is correct, leakage should appear as spurious attributions of the wrong color in the target's graph.

The attribution graph for $f^{(\text{img}, 10)}_{\text{blue}}$ on the prompt \textit{a red apple and a blue cup}, in addition to the expected blue sources, indeed contains a small number of $f^{(\text{txt}, 3)}_{\text{red}}$ features with attributions one to two orders of magnitude weaker than the dominant ones. If this spurious attribution is causal, positive steering of these red features in the blue graph should switch the cup to red.

\begin{figure}[h]
    \centering
    \includegraphics[width=\linewidth]{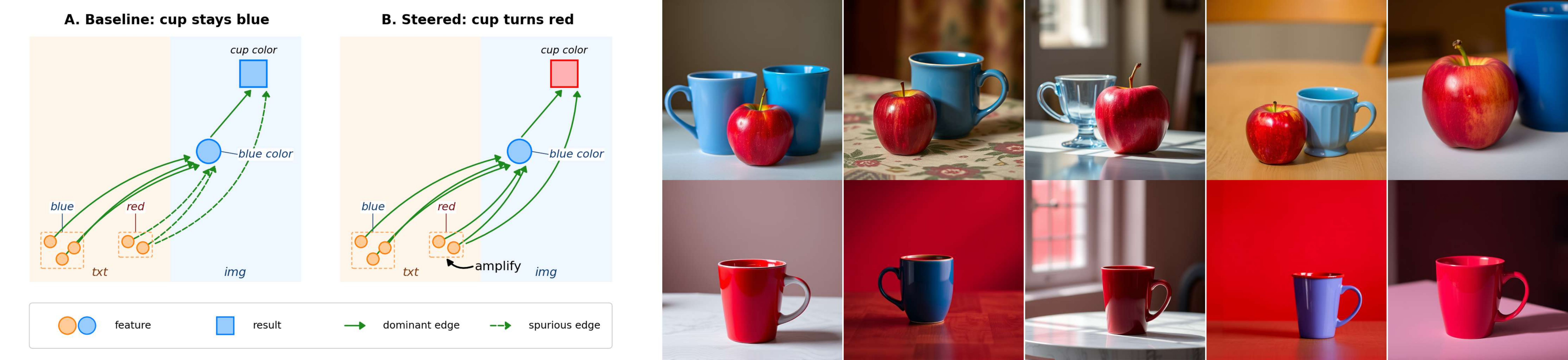}
    \caption{Prompt \textit{a red apple and a blue cup}. Rows: baseline; amplify red features from the blue graph, $\alpha = +100$. Columns: five seeds. On three seeds the cup turns red; on two seeds the spurious red features are absent from the graph, and steering instead colors the background while leaving the cup blue.}
    \label{fig:color_leakage}
\end{figure}

The experiment confirms the hypothesis (Fig.~\ref{fig:color_leakage}). On a grid of $16$ prompts with a red and a blue object $\times$ $5$ seeds, labeled blind (Appendix~\ref{app:blind}), the blue object turns red in $25\%$ of generations at $\alpha = +30$, while the more common outcome is that the surroundings turn red instead ($62.5\%$). The causal direction is clear, but the effect is seed-dependent, which is consistent with the nature of attribution graphs: the cause of leakage is localizable to specific features, and those features enter the graph only on some seeds -- on the seeds where they do not, the steering targets the wrong locations.

\subsection{Numerical concepts}
\label{sec:exp:counting}

Generating an exact number of objects is a known weakness of current text-to-image models. The model's internal representation of numerals separates into two conceptually distinct questions: whether the model has a visual representation of count, and whether the text stream carries a correct representation of specific numerals. Our analysis suggests that the difficulty is not where one might expect.

Using contrastive prompts we identify $f^{(\text{img}, 14)}_{\text{multi}}$, an image-stream feature whose activation grows with the number of objects. Its activation on \textit{five apples} is roughly equal to its activation on \textit{three apples}, and five times larger than activation on \textit{one apple}. Already this suggests that the feature does not encode an exact count, but rather a notion of multiplicity.

The attribution graph for $f^{(\text{img}, 14)}_{\text{multi}}$ on the prompts \textit{one/three/five red apples} has nearly identical image-stream parts; the differences are localized in the text stream. On \textit{one apple}, text-stream features for \textit{single} and \textit{one} are active; on \textit{three apples}, primarily \textit{three} is active, with side activations on \textit{two} and \textit{several}; on \textit{five apples}, a diffuse mixture is active, including features for \textit{two}, \textit{three}, \textit{four}, \textit{five}, \textit{six}, \textit{seven}, \textit{eight}, and \textit{several}. The presence of features for adjacent numerals in the \textit{five apples} graph indicates that the text-stream representation of five is not sharp: instead of a clean activation of \textit{five} features, the model activates a diffuse cluster of neighboring numerals. This diffuseness is a plausible candidate for the source of counting errors, and can be tested directly via steering.

\begin{figure}[h]
    \centering
    \includegraphics[width=0.75\linewidth]{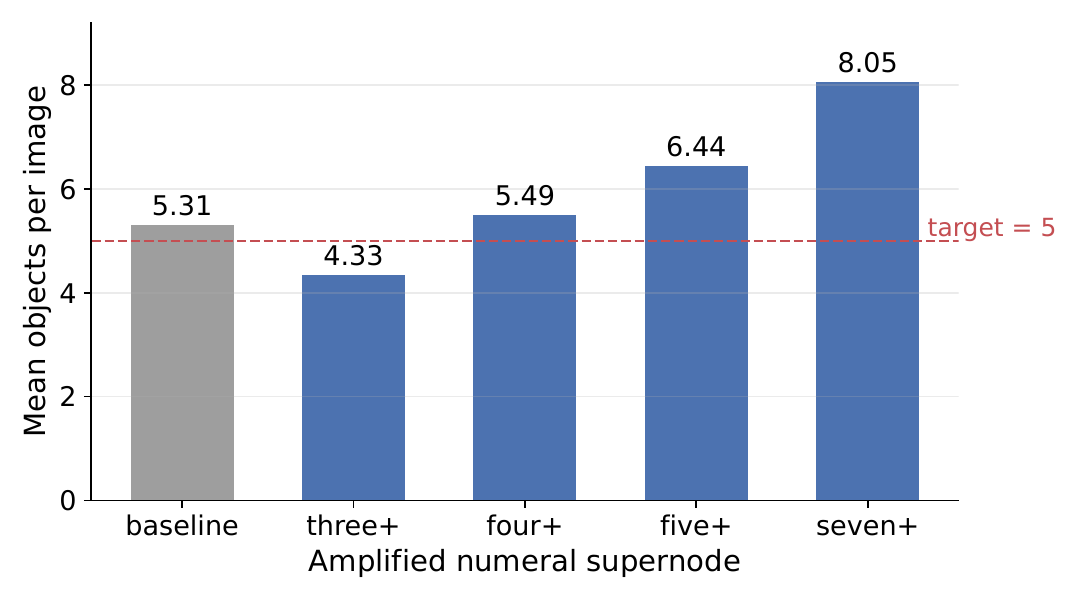}
    \caption{\textbf{Numerals order the count but do not set it.} Amplifying the text-stream supernode of a numeral ($\alpha = +30$) on prompts asking for five objects changes how many appear, monotonically in the numeral, but the mean never lands on the numeral itself: \textit{four}+ gives about five and a half and \textit{five}+ more than six. Bars: mean objects per image over $16$ prompts $\times$ $5$ seeds, counted blind; the $5$ images per condition labeled unclear are left out. Horizontal line: the requested count of five.}
    \label{fig:counting_chart}
\end{figure}

On a grid of $16$ prompts $\times$ $5$ seeds, counted blind (Appendix~\ref{app:blind}), amplifying \textit{three} shifts the mean to $4.33$ and amplifying \textit{seven} to $8.05$ (Figure~\ref{fig:counting_chart}). Yet amplifying \textit{five} yields $6.44$, and amplifying \textit{four} yields $5.49$, with the baseline at $5.31$ against a requested count of five. The pattern does not match the simple model in which amplifying the feature for numeral $N$ produces $N$ objects: amplifying \textit{five} shifts the mean upward, away from five, and amplifying \textit{four} barely moves it. On the other hand, monotonicity is preserved -- a larger numeral always yields more objects than a smaller one. The model thus carries a robust representation of \textit{ordering} (greater / lesser), but lacks a sharp representation of specific values.

\subsection{Failure modes from cross-stream disconnect}
\label{sec:exp:failures}

We close with two examples of systematic generation failures that our method diagnoses as failures of information transfer between streams. Both cases exhibit the same pattern: the text stream carries the information required by the prompt correctly, but that information does not drive the corresponding change in image-stream behavior. They simultaneously illustrate the diagnostic capabilities of the method and characterize its current limitations.

\paragraph{Negation: \textit{a room without a cat}.}
Contrasting \textit{a room with a cat} against \textit{a room without a cat}, we identify a text-stream feature $f^{(\text{txt}, 10)}_{\text{empty}}$ that activates on the token \textit{empty} and also on \textit{without} in the target prompt; its own graph contains other text-stream features for emptiness semantics (firing on tokens such as \textit{empty}, \textit{abandoned}, \textit{no}, and similar). The text stream therefore carries a correct representation of an empty room -- the model understands the negation at the linguistic level. Amplifying all of these features at $\alpha = +30$ removes the cat in only $12.5\%$ of the generations on a grid of $16$ negated prompts $\times$ $5$ seeds labeled blind (Appendix~\ref{app:blind}), against $8.75\%$ without the intervention; on the matched affirmative prompts (\textit{a room with a cat}), the same intervention removes it in $1.25\%$, so the effect is weak in both directions. The cat is absent from all $80$ generations only when an independently identified text-stream supernode for the cat itself is suppressed at $\alpha = -30$, with or without the emptiness features. Information about emptiness, correctly formed in the text stream, simply does not propagate into the image stream.

\begin{figure}[h]
    \centering
    \includegraphics[width=\linewidth]{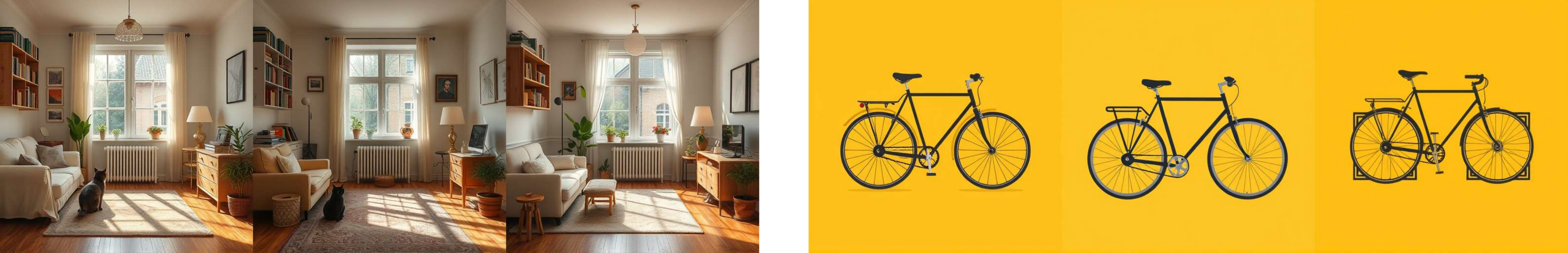}
    \caption{Failure modes from cross-stream disconnect. \textbf{Left:} \textit{a room without a cat}; baseline; amplification of $f^{(\text{txt}, 10)}_{\text{empty}}$ ($\alpha = +30$); suppression of the text-stream cat features ($\alpha = -30$). \textbf{Right:} \textit{a bicycle with square wheels}; baseline; suppression of $f^{(\text{txt}, 10)}_{\text{round}}$ ($\alpha = -80$); amplification of the image-stream angularity feature ($\alpha = +80$).}
    \label{fig:failures}
\end{figure}

\paragraph{Hard prior: \textit{bicycle with square wheels}.}
The model consistently draws round wheels despite the explicit qualifier in the prompt. A contrastively identified feature $f^{(\text{txt}, 10)}_{\text{round}}$ is active on \textit{a bicycle with round wheels}; on \textit{a bicycle with square wheels}, its preactivation magnitude drops below one -- that is, the roundness semantics in the text stream is substantially weakened in response to the qualifier \textit{square wheels}, but this does not produce square wheels in the generated image. Direct negative steering of $f^{(\text{txt}, 10)}_{\text{round}}$ has no effect. We then identify, again via contrastive prompts, angularity features in both streams. We ran the case on $16$ prompts $\times$ $5$ seeds under four conditions: the roundness features at $\alpha = -80$, the text-stream or the image-stream angularity features at $\alpha = +80$, and roundness at $-80$ together with image-stream angularity at $+80$. Only one of them yields something close to square wheels: amplifying the image-stream angularity features draws square outlines around wheels that stay round (Figure~\ref{fig:failures}).

In both cases we observe the same disconnect: the text stream represents the prompt requirement correctly, but this representation does not propagate to the image stream, and replacing it on the image-stream side succeeds only partially at best. The picture is consistent with the quantitative shift documented in §\ref{sec:exp:temporal}: the text-stream influence decays rapidly toward later denoising steps, and for strong image-side priors, the diminishing text-stream channel may be insufficient to overwrite the pretrained visual behavior, even when the text-stream semantics is set up correctly. The absence of a bridge between a correct semantic representation and its realization in visual behavior is potentially a primary source of systematic failures of FLUX on prompts with explicitly non-standard requirements.

\section{Blind evaluation of the case studies}
\label{app:blind}

\paragraph{Protocol.} Each item is one (baseline image, intervened image) pair. The annotator saw the prompt, the two images and one fixed question with a closed label set defined in advance; the condition that produced the intervened image was never displayed, and the presentation order was shuffled within each experiment. In total $4{,}099$ items were labeled across ten experiments (Table~\ref{tab:blind}; the intervention strengths are in Table~\ref{tab:strengths}). Five of the ten were labeled by three annotators ($1{,}524$ items); pairwise Cohen's $\kappa$ between these annotators is $0.71$--$0.80$ pooled over the five experiments and $0.60$--$0.86$ per experiment and annotator pair. For them we report the majority vote; the $53$ items without a majority take the first annotator's label. For the other five experiments, marked $^\dagger$, we report the labels of one annotator.

\begin{table}[H]
\centering
\caption{Blind evaluation. Columns: the labeling grid; the condition the graph predicts should work; the control or the alternative condition. Experiments marked $^\dagger$ report a single annotator's labels.}
\label{tab:blind}
\resizebox{\textwidth}{!}{%
\begin{tabular}{lllll}
\toprule
Experiment & \makecell[l]{Grid (prompts $\times$\\ seeds $\times$ modes)} & Measure & Graph-predicted condition & Control / comparison \\
\midrule
Temporal causality & $12 \times 3 \times 2$ & cat removed & steps $0$--$1$: $72.2\%$ & steps $2$--$3$: $0\%$ \\
\midrule
\multirow{3}{*}{Concept vs context} & \multirow{3}{*}{$16 \times 3 \times 6$} & identity stripped, shape kept & context: $87.5\%$ & concept: $27.1\%$ \\
 & & object replaced or removed & concept: $45.8\%$ & context: $6.3\%$ \\
 & & recognizable bat left & both: $2.1\%$ & -- \\
\midrule
\multirow{2}{*}{Suppressor} & \multirow{2}{*}{$12 \times 3 \times 4$} & cat appears (dog prompts) & joint: $36.1\%$ & \makecell[l]{dog features alone: $2.8\%$\\ suppressor alone: $0\%$} \\
 & & cat removed (cat prompts) & cat features: $88.9\%$ & -- \\
\midrule
\multirow{2}{*}{Color prior$^\dagger$} & \multirow{2}{*}{$8 \times 30 \times 6$} & requested color & graph: $82.1\%$ & co-activation: $76.7\%$ \\
 & & object destroyed & graph: $13.8\%$ & co-activation: $21.7\%$ \\
\midrule
Polysemy & $10 \times 3 \times 2$ & baseball reading & $\alpha = +100$: $70\%$ & $\alpha = +30$: $33\%$ \\
\midrule
\multirow{2}{*}{Reflections$^\dagger$} & \multirow{2}{*}{$17 \times 5 \times 3$} & surface intact (reflection removed) & layer $11$: $86.4\%$ & layer $12$: $37.5\%$ \\
 & & image unchanged & -- & control features: $100\%$ \\
\midrule
\multirow{3}{*}{Spatial composition} & \multirow{3}{*}{$16 \times 5 \times 12$} & red object moves right & pair, $\alpha = 10$: $85\%$ & left suppressed, $\alpha = 10$: $39\%$ \\
 & & composition unchanged & -- & \makecell[l]{reversed pair: $97.5\%$\\ pair, $\alpha = 3$: $98.8\%$} \\
 & & image degraded & -- & pair, $\alpha = 100$: $92.5\%$ \\
\midrule
Color leakage$^\dagger$ & $16 \times 5 \times 2$ & blue object turns red & $20$--$25\%$ & surroundings turn red: $62.5$--$65\%$ \\
\midrule
Counting$^\dagger$ & $16 \times 5 \times 5$ & mean objects (five requested) & \makecell[l]{three+ $4.33$, four+ $5.49$\\ five+ $6.44$, seven+ $8.05$} & baseline: $5.31$ \\
\midrule
\multirow{2}{*}{Negation$^\dagger$} & \multirow{2}{*}{$16 \times 5 \times 4$} & \multirow{2}{*}{cat absent} & emptiness amplified: $12.5\%$ & \makecell[l]{no intervention: $8.75\%$\\ affirmative prompts: $1.25\%$} \\
 & & & -- & cat feature suppressed: $100\%$ \\
\bottomrule
\end{tabular}%
}
\end{table}

\begin{table}[H]
\centering
\caption{Intervention strengths used in each experiment.}
\label{tab:strengths}
\begin{tabular}{ll}
\toprule
Experiment & Strength \\
\midrule
Temporal causality & $-30$ \\
\midrule
\multirow{2}{*}{Concept vs context} & $-15$ (the table above reports) \\
 & $-30$ \\
\midrule
\multirow{2}{*}{Suppressor} & $-60$ for each of the three single-channel conditions \\
 & $-20$ on the dog features with $-30$ on the suppressor \\
\midrule
Color prior & $-80$ \\
\midrule
\multirow{2}{*}{Polysemy} & $+30$ \\
 & $+100$ \\
\midrule
Red apple / blue apple & $\pm 15$ \\
\midrule
Watercolor & $\pm 15$ \\
\midrule
Reflections & $-30$ \\
\midrule
\multirow{3}{*}{Spatial composition} & $\pm 30$ single-feature \\
 & $\pm 10$ paired \\
 & sweep at $3$, $10$, $30$, $100$ \\
\midrule
\multirow{2}{*}{Color leakage} & $+30$ \\
 & $+100$ \\
\midrule
Counting & $+30$ \\
\midrule
\multirow{2}{*}{Negation} & $+30$ on the emptiness features \\
 & $-30$ on the cat feature \\
\bottomrule
\end{tabular}
\end{table}

\subsection{A vision-language model as a further rater}
\label{app:blind:vlm}

The color-prior comparison on FLUX.1[dev] and SD3.5-medium (Appendix~\ref{app:backbones:color}) is labeled by a vision-language model, not by annotators. To measure how far this instrument agrees with people, the same model labeled every FLUX.1[schnell] experiment of Table~\ref{tab:blind}; Table~\ref{tab:vlm_agreement} compares the two.

\paragraph{Setup.} The model is Qwen3.6-35B-A3B~\cite{qwen2026qwen36} in bfloat16, with greedy decoding and its reasoning mode off. It receives the annotators' instructions, translated into English (the question, the closed label set with its hints, and the tie-break rules), and sees what they saw: the prompt, and for an intervened image the baseline of the same seed next to it, never the condition. It answers with a one-sentence observation and a label. For the color questions the instructions first ask whether the object is still recognizable; all other experiments use the annotators' instructions unchanged.

\paragraph{Agreement.} On four of the five experiments with three annotators, the model agrees with an individual annotator about as often as two annotators agree with each other. The exception is concept vs context: the model calls a clear bat $53$ of the $114$ stick-like objects in which the annotators no longer recognized a baseball bat, so it finds the context set's ``identity stripped, shape kept'' outcome in $37.5\%$ of the generations, not $87.5\%$. On the experiments with one annotator, agreement is $0.81$--$0.997$ for color prior, color leakage and negation, $0.72$ for counting (the model's count is within one object of the annotator's on $96\%$ of the frames), and $0.63$ for reflections, where the model judges the reflection removed less often ($16$ against $44$ of the $85$ layer-$11$ generations).

Every comparison of Table~\ref{tab:blind} keeps its direction. The rates move by at most $7$ points and the mean counts by at most $0.4$ objects, with four exceptions: concept vs context, where the context set's success falls to $37.5\%$ from $87.5\%$, still above the concept set; the layer-$11$ reflection rate ($68.8\%$ against $86.4\%$, still above layer $12$); image degradation at $\alpha = 100$ in spatial composition ($71.3\%$ against $92.5\%$); and color leakage at $\alpha = +100$ ($32.5\%$ and $55\%$ against $20\%$ and $65\%$, with the surroundings still ahead). The color-prior destruction gap keeps its sign but shrinks from $7.9$ to $2.5$ points and is not significant under the model's labels. On the question asked on the other backbones, whether the object is intact and in which color, the model agrees with the annotator on $88\%$ of the frames, but on FLUX.1[schnell] it detects a smaller difference in object destruction than the annotator did.

\begin{table}[H]
\centering
\caption{The vision-language model against the annotators on the FLUX.1[schnell] experiments of Table~\ref{tab:blind}. \emph{Frames}: the items of each grid of Table~\ref{tab:blind}. \emph{Agree}, $\kappa$: against the label reported in Table~\ref{tab:blind}. \emph{Model / annotators}: for the experiments with three annotators, the model's mean agreement with each annotator next to the annotators' mean agreement with each other. Experiments marked $^\dagger$ report a single annotator's labels.}
\label{tab:vlm_agreement}
\begin{tabular}{lrrrl}
\toprule
Experiment & Frames & Agree & $\kappa$ & Model / annotators \\
\midrule
Temporal causality & $72$ & $0.94$ & $0.88$ & $0.93$ / $0.90$ \\
Concept vs context & $288$ & $0.49$ & $0.34$ & $0.49$ / $0.79$ \\
Suppressor & $144$ & $0.88$ & $0.82$ & $0.87$ / $0.85$ \\
Color prior$^\dagger$ & $960$ & $0.88$ & $0.76$ & -- \\
Polysemy & $60$ & $0.87$ & $0.78$ & $0.83$ / $0.87$ \\
Reflections$^\dagger$ & $255$ & $0.63$ & $0.33$ & -- \\
Spatial composition & $960$ & $0.78$ & $0.69$ & $0.72$ / $0.74$ \\
Color leakage$^\dagger$ & $160$ & $0.81$ & $0.65$ & -- \\
Counting$^\dagger$ & $400$ & $0.72$ & $0.67$ & -- \\
Negation$^\dagger$ & $320$ & $0.997$ & $0.99$ & -- \\
\bottomrule
\end{tabular}
\end{table}

\section{Transcoders}
\label{app:transcoders}

\subsection{Architecture details}
\label{app:tc:arch}

For each (layer, stream) pair $(\ell, s)$ with $\ell \in \{0, \ldots, 15\}$ and $s \in \{\mathrm{img}, \mathrm{txt}\}$ we train an independent temporal-aware transcoder $TC^s_\ell$. The architecture is the one summarized in §\ref{sec:method:architecture} and shown in Figure~\ref{fig:tc_architecture}. Here we give the full set of components together with the design choices that we found necessary in practice.

\paragraph{Timestep embedding.} The diffusion timestep $t \in [0, 1]$ is first mapped into a $d_t$-dimensional vector by a sinusoidal positional code $\mathrm{SinEmb}(t) \in \mathbb{R}^{d_t}$ with $d_t = 256$, of the same form and dimensionality as the base model's timestep embedding. The result is processed by a small MLP with two linear layers and SiLU activations, which adds capacity for the modulation parameters to depend nonlinearly on $t$ across the four denoising steps:

\begin{equation}
e_t = \mathrm{SiLU}\!\bigl(W_2\,\mathrm{SiLU}(W_1\,\mathrm{SinEmb}(t) + b_1) + b_2\bigr), \quad W_1, W_2 \in \mathbb{R}^{d_t \times d_t}.
\end{equation}

This time-conditioning subnetwork has its own weights for every transcoder. Sharing it across $(\ell, s)$ pairs would tie features across blocks in a way we explicitly want to avoid.

\paragraph{FiLM modulation of the encoder input.} A linear projection $W_{\mathrm{mod}} \in \mathbb{R}^{2 d_{\mathrm{model}} \times d_t}$ maps $e_t$ to a pair of scale and shift vectors,
\begin{equation}
[\,\mathrm{scale}^{\mathrm{tc}}(t)\,;\,\mathrm{shift}^{\mathrm{tc}}(t)\,] = W_{\mathrm{mod}}\, e_t + b_{\mathrm{mod}},
\end{equation}
which modulate the MLP input $x \in \mathbb{R}^{d_{\mathrm{model}}}$ elementwise:
\begin{equation}
x_{\mathrm{mod}} = x \odot \bigl(1 + \mathrm{scale}^{\mathrm{tc}}(t)\bigr) + \mathrm{shift}^{\mathrm{tc}}(t).
\end{equation}
Both $W_{\mathrm{mod}}$ and $b_{\mathrm{mod}}$ are initialized to zero so that $\mathrm{scale}^{\mathrm{tc}}(t) = \mathrm{shift}^{\mathrm{tc}}(t) = 0$ at the start of training and $x_{\mathrm{mod}} = x$. Without this zero initialization the modulation introduces a strong perturbation to the encoder input from step $0$ and disrupts early training.

\paragraph{Sparse encoder and linear decoder.} The modulated input is mapped to feature activations and back to $\mathbb{R}^{d_{\mathrm{model}}}$:
\begin{align}
z(x, t) &= \mathrm{ReLU}\!\bigl(W_{\mathrm{enc}}\, x_{\mathrm{mod}} + b_{\mathrm{enc}}\bigr), \\ TC^s_\ell(x, t) &= W_{\mathrm{dec}}\, z(x, t) + b_{\mathrm{dec}},
\end{align}
with $W_{\mathrm{enc}} \in \mathbb{R}^{d_{\mathrm{feat}} \times d_{\mathrm{model}}}$, $W_{\mathrm{dec}} \in \mathbb{R}^{d_{\mathrm{model}} \times d_{\mathrm{feat}}}$, and biases of matching shape. We use $d_{\mathrm{model}} = 3072$ and $d_{\mathrm{feat}} = 16\, d_{\mathrm{model}} = 49\,152$ throughout, giving each transcoder approximately $304$M trainable parameters.

\paragraph{Initialization.} The decoder weight $W_{\mathrm{dec}}$ is initialized with Kaiming uniform; the encoder weight is then tied to it, $W_{\mathrm{enc}} \leftarrow W_{\mathrm{dec}}^\top$. After this tying, the columns of $W_{\mathrm{dec}}$ are renormalized to unit norm. Both biases are initialized to zero, as are $W_{\mathrm{mod}}, b_{\mathrm{mod}}$. The two-layer time MLP uses Kaiming normal initialization.

\paragraph{Decoder column normalization.} After every optimizer step the columns of $W_{\mathrm{dec}}$ are projected back onto the unit sphere,
\begin{equation}
W_{\mathrm{dec}}[:, i] \;\leftarrow\; \frac{W_{\mathrm{dec}}[:, i]}{\|W_{\mathrm{dec}}[:, i]\|_2}, \qquad i = 1, \ldots, d_{\mathrm{feat}}.
\end{equation}
This is the standard SAE/transcoder practice and has a concrete purpose: without it, the optimizer can trivially evade the $L_1$ penalty on $z$ by inflating the columns of $W_{\mathrm{dec}}$ and shrinking $z$ proportionally, leaving $TC^s_\ell$ unchanged but reducing the sparsity term arbitrarily. Unit norm decoders fix the scale and make $\|z\|_1$ a meaningful proxy for the number of active features.

\subsection{Training data}
\label{app:tc:data}

\paragraph{Prompt corpus.} The activation buffers are populated by running the frozen FLUX.1[schnell] pipeline on prompts streamed from \texttt{yvdao/midjourney-v6}, a corpus of approximately $310\,000$ user prompts collected from Midjourney v6. Prompts of $16$ characters or fewer are skipped, longer prompts are truncated at $512$ characters.

\paragraph{Inference configuration.} All forward passes are run at $512 \times 512$ resolution with $4$ denoising steps and guidance scale $0$, which is the configuration FLUX.1[schnell] was distilled for. Each call to the FLUX.1[schnell] pipeline triggers $4$ transformer forward passes (one per denoising step), each of which fills the activation buffers with the corresponding records.

\paragraph{Activation harvesting.} For every target block $\ell$ and stream $s$ we register a forward hook on the corresponding feed-forward sublayer that captures the input $x \in \mathbb{R}^{B \times S \times d_{\mathrm{model}}}$ and the output $y = \mathrm{MLP}^s_\ell(x)$. A separate forward pre-hook on the transformer caches the current timestep $t$, which is broadcast to per-token records:
\begin{equation}
\bigl\{(x_b^{s, p}, y_b^{s, p}, t_b)\bigr\}_{b, p}, \qquad x_b^{s, p}, y_b^{s, p} \in \mathbb{R}^{d_{\mathrm{model}}}, \; t_b \in \mathbb{R}.
\end{equation}
These records are appended to a per-(layer, stream) buffer of size $10^6$ pairs; each transcoder has its own buffer.

\paragraph{Buffer asymmetry.} Within a single forward pass, the image stream produces $S_{\mathrm{img}} = 1024$ records per prompt and the text stream $S_{\mathrm{txt}} = 512$: the pipeline pads the T5 encoding of every prompt to $512$ tokens, so most text records come from padding positions. The data-collection loop terminates when \emph{any} buffer reaches capacity, which is always an image-stream buffer; at that point each text-stream buffer holds about half as many records. We deliberately do not equalize the streams by collecting more forward passes or oversampling text records: we found that simply sampling text batches with replacement from the partially-filled buffer during the optimization phase, with the same number of optimizer steps as for image transcoders, gives stable convergence. Image transcoders therefore see each example about once per cycle, while text transcoders see each example about twice.

\subsection{Loss and optimization}
\label{app:tc:loss}

\paragraph{Loss.} For each transcoder we minimize
\begin{equation}
\mathcal{L}^s_\ell = \underbrace{\frac{\mathbb{E}_{x, t}\,\|\mathrm{MLP}^s_\ell(x) - TC^s_\ell(x, t)\|_2^2}{\sum_{j=1}^{d_{\mathrm{model}}} \mathrm{Var}_{x, t}\bigl(\mathrm{MLP}^s_\ell(x)_j\bigr) + \varepsilon}}_{\text{normalized faithfulness loss}}
+ \underbrace{\lambda^s\, \mathbb{E}_{x, t}\,\|z(x, t)\|_1}_{\text{sparsity penalty}},
\label{eq:tc_loss}
\end{equation}
with $\varepsilon = 10^{-6}$. Both expectations are estimated by Monte Carlo over the current minibatch of $4096$ records drawn uniformly with replacement from the (layer, stream) buffer. The variance in the denominator is computed over the same minibatch, with $\mathrm{Var}$ unbiased$=$False.

\begin{figure}[H]
\centering
\includegraphics[width=\linewidth]{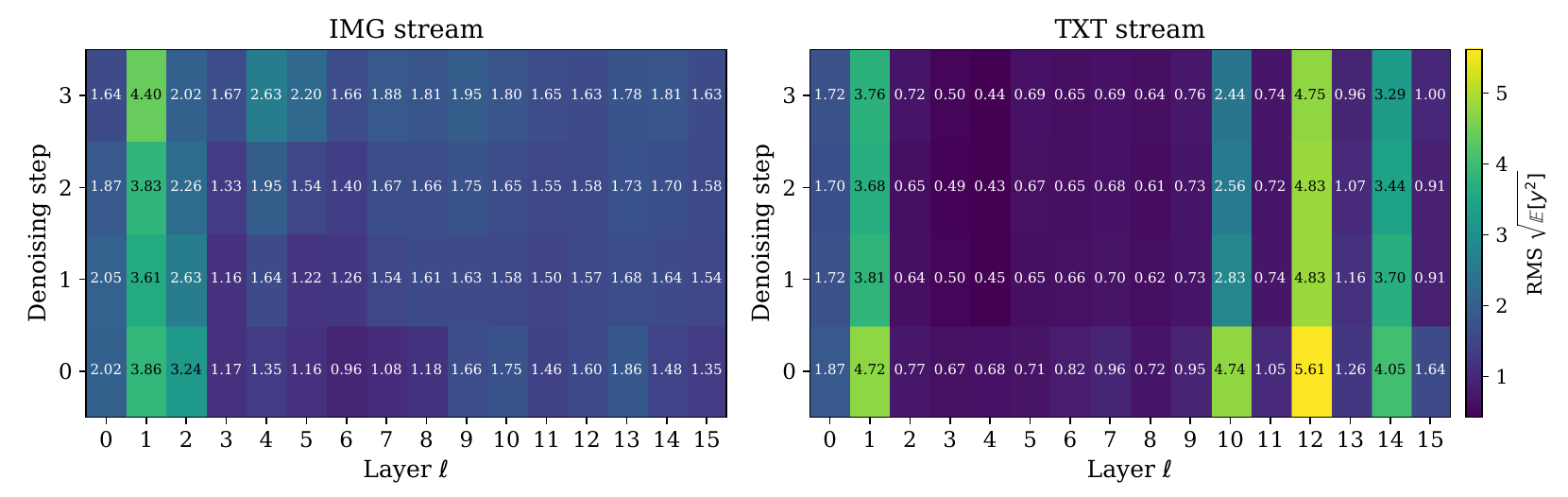}
\caption{MLP-output magnitude (RMS $\sqrt{\mathbb{E}[y^2]}$) per \emph{(layer, stream, step)} bucket, measured on $512$ prompts; linear color scale shared between the two panels. A $\sim\!13\times$ spread in RMS motivates the variance-normalized form of the faithfulness term.}
\label{fig:activation_heatmap}
\end{figure}

\paragraph{Why variance normalization matters.} Magnitudes of the MLP outputs $y$ in MM-DiT vary substantially across the $32$ transcoder targets. On $512$ prompts, all $4$ denoising steps, and all $16$ analyzed double-stream blocks ($128$ \emph{(layer, stream, step)} buckets in total), per-bucket RMS $\sqrt{\mathbb{E}[y^2]}$ spans $0.43$ to $5.61$ ($\sim\!13\times$), and tail magnitudes $\max |y|$ span $\sim\!19$ to $\sim\!1500$ -- close to two orders of magnitude (Figure~\ref{fig:activation_heatmap}). Under a plain squared-error loss, per-bucket expected loss scales as $\mathrm{RMS}^2$ and would differ by a factor of $\sim\!170$ across buckets at equal reconstruction quality; rare outlier tokens with $|y| \sim 10^3$ then contribute single-element errors several further orders of magnitude above the typical. The variance-normalized form of the faithfulness term in (\ref{eq:tc_loss}) absorbs each transcoder's output scale into the denominator, giving $\lambda$ the same meaning for all $32$ transcoders; the smaller spread across the four steps of one block (at most about $2\times$) is not normalized. This is what allowed us to reach a uniform sparsity-faithfulness operating point across all $32$ transcoders with two stream-level $\lambda$ values.

\paragraph{Per-stream sparsity coefficients.} The image and text streams differ qualitatively in the distribution of MLP activations. Empirically the same $\lambda$ for both streams either drives image transcoders to dense activations (if low) or collapses text transcoders to high reconstruction error (if high). We therefore use $\lambda^{\mathrm{img}} = 3 \times 10^{-4}$ and $\lambda^{\mathrm{txt}} = 5 \times 10^{-5}$.

\paragraph{Optimizer and schedule.} Each transcoder is optimized independently with AdamW (zero weight decay, default $\beta$). The learning rate is $2 \times 10^{-4}$ for both streams, decayed by a cosine annealing schedule over $256$ training cycles. We define one cycle as: clear all buffers, run inference until any buffer fills to $10^6$ records, then perform one optimizer epoch over each buffer with replacement-sampled batches. The total training budget is approximately $256$M activation records per transcoder.

\paragraph{Multi-run training.} Holding $32$ transcoders in GPU memory simultaneously together with the FLUX.1[schnell] base model exceeds the memory budget of a single NVIDIA H100. We therefore train transcoders in disjoint groups of at most six (three layers $\times$ two streams), with the same fixed random seed for the data sampler and the same training schedule. A group of six takes about ten hours on one NVIDIA H100; each transcoder has about $304$M parameters and is trained on about $256$M records. The base model and the data corpus are identical across runs; only the active set of transcoders differs.

\subsection{Timestep conditioning}
\label{app:tc:film}

\begin{figure}[H]
\centering
\includegraphics[width=\linewidth]{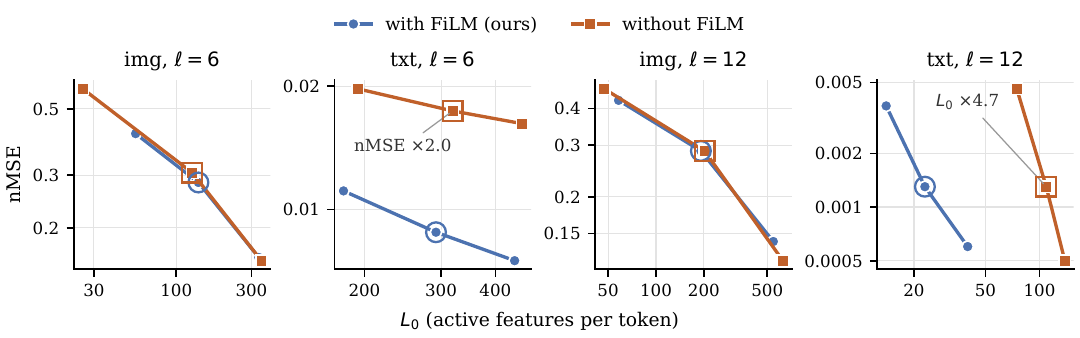}
\caption{Sparsity--faithfulness frontiers of transcoders with and without the FiLM timestep conditioning, layers $6$ and $12$ of FLUX.1[schnell]. Each curve connects three transcoders trained at twice, once and half the stream's default $\lambda$ (the default ringed); lower left is better. Without FiLM the image-stream frontier does not move; the text stream reaches twice the nMSE at layer $6$ and needs $3.5$ to $5.3$ times as many active features at layer $12$.}
\label{fig:film_ablation}
\end{figure}

To see whether the conditioning is needed, we retrained the transcoders of layers $6$ and $12$, both streams, without it -- the same encoder and decoder with the time MLP and the FiLM head removed, the same data, schedule and seed -- at the three sparsity coefficients of Figure~\ref{fig:sae_comparison}: the stream's default $\lambda$ (Table~\ref{tab:hyperparams}), twice and half of it. Figure~\ref{fig:film_ablation} compares the frontiers. In the image stream they coincide: at the default $\lambda$ the layer-$6$ transcoder reaches an nMSE of $0.283$ at $L_0 = 138$ with FiLM and $0.306$ at $126$ without, the layer-$12$ one $0.287$ at $192$ and $0.287$ at $202$. In the text stream the conditioning matters. Without it the layer-$6$ transcoder has $1.8$--$2.2$ times the nMSE at a slightly larger $L_0$ ($0.0174$ against $0.0088$ at the default $\lambda$), and the layer-$12$ one needs $3.5$--$5.3$ times as many active features at a comparable nMSE ($109$ against $23$ at the default $\lambda$, both at an nMSE of $0.0013$). Timestep-aware dictionaries were used before for SAEs on diffusion transformers~\cite{huang2026tide}; for transcoders the conditioning is a design choice that pays off in the text stream.

\subsection{Quantitative evaluation}
\label{app:tc:eval}

We evaluate the trained transcoders along two axes: their direct fit to the per-block MLPs they replace (sparsity and faithfulness curves over training), and their effect on the model's outputs when all $32$ transcoders are simultaneously substituted for the corresponding MLPs and a full image is generated (end-to-end faithfulness).

\begin{figure}[H]
\centering
\includegraphics[width=\linewidth]{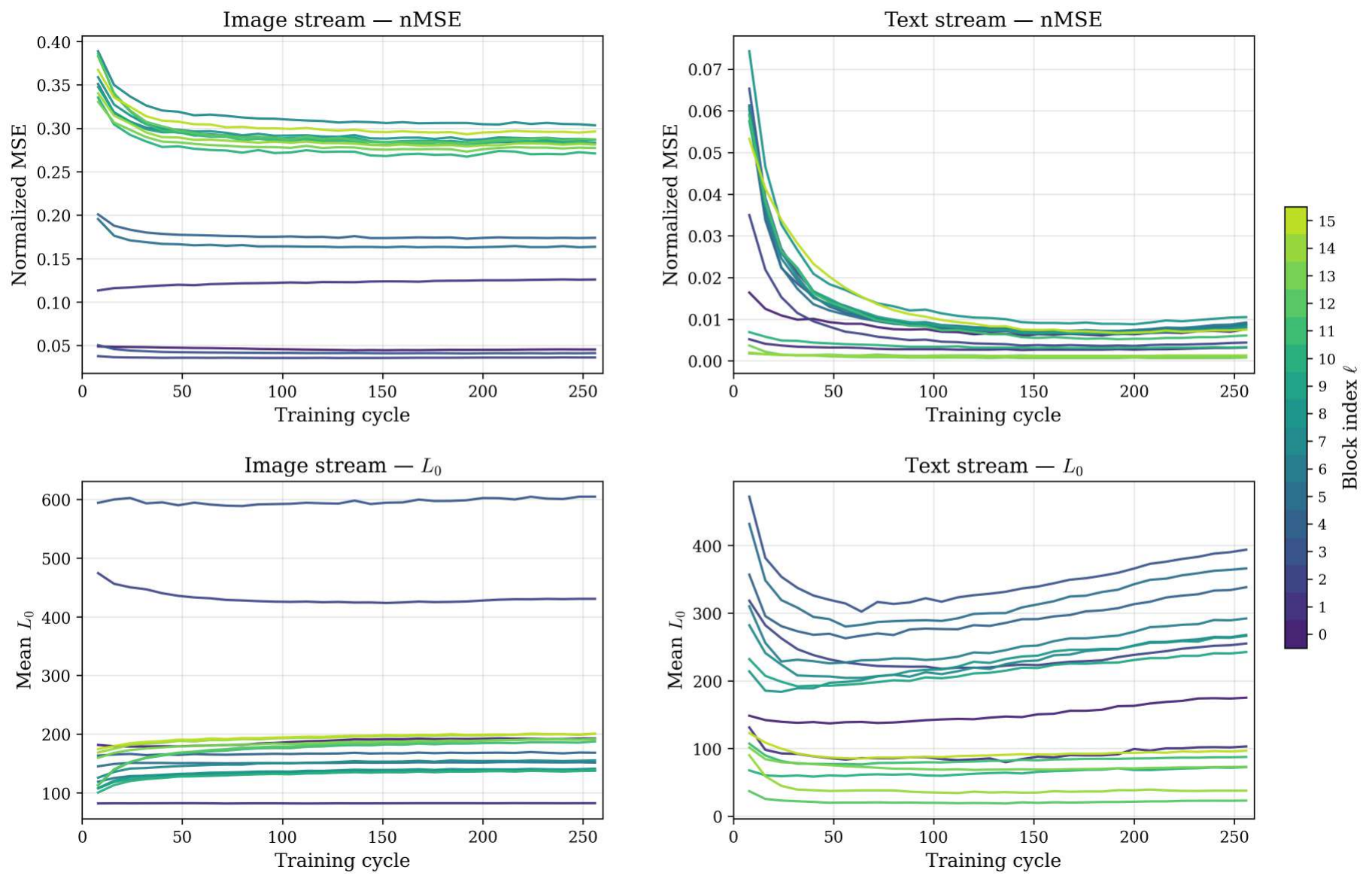}
\caption{Training curves for all $32$ transcoders, recorded every $8$ cycles over $256$ cycles. \textbf{Top row:} normalized MSE. \textbf{Bottom row:} mean $L_0$ activation. One curve per transcoder, colored by block index $\ell$.}
\label{fig:tc_training_curves}
\end{figure}

\paragraph{Per-transcoder training metrics.} Figure~\ref{fig:tc_training_curves} reports two metrics per transcoder, recorded every $8$ training cycles: the normalized MSE between $\mathrm{MLP}^s_\ell(x)$ and $TC^s_\ell(x, t)$ on the current training buffer, and the mean $L_0$ of $z(x, t)$ on the same batch (defined as the mean number of strictly positive feature activations per token). The four panels split metrics by stream and by axis (nMSE vs $L_0$); within each panel one curve is drawn per transcoder, colored by layer. By the end of training nMSE plateaus at $0.04$--$0.30$ for image transcoders and $0.001$--$0.011$ for text transcoders, while $L_0$ reaches $82$--$605$ active features per token (image) and $23$--$394$ (text), corresponding to $0.05\%$--$1.2\%$ of $d_{\mathrm{feat}} = 49\,152$ active per token. Text-stream nMSE values are systematically lower than image-stream values, while $L_0$ is not (at layer $6$ the text transcoder is the denser one, Appendix~\ref{app:tc:film}): text-stream MLPs in double-stream blocks perform less drastic transformations than image-stream MLPs, since the text branch primarily carries T5 prompt features through, while the image branch performs the bulk of cross-modal integration, so the reconstruction is easier.

\paragraph{End-to-end faithfulness.} A small per-block reconstruction error can compound over $16$ layers and $4$ denoising steps into a substantial drift in the generated image, so per-block metrics alone do not establish that the transcoders are useful as drop-in replacements. We therefore measure the end-to-end faithfulness of the full replacement model (all $32$ MLPs replaced by their transcoders, attention and normalization untouched, no error correction terms) against the original FLUX.1[schnell] on a fixed set of $512$ prompts. We compare in latent space, before VAE decoding, by computing two metrics per prompt: cosine similarity $\cos(l_{\mathrm{orig}}, l_{\mathrm{tc}})$ and the mean squared error $\tfrac{1}{n}\|l_{\mathrm{orig}} - l_{\mathrm{tc}}\|_2^2$ over the $n$ entries of the final flat latents. Aggregate values are reported in Table~\ref{tab:tc_e2e}.

\begin{table}[H]
\centering
\caption{End-to-end faithfulness of the full replacement model (all $32$ MLPs replaced) against FLUX.1[schnell], on $512$ prompts at $512 \times 512$ resolution and $4$ denoising steps.}
\label{tab:tc_e2e}
\begin{tabular}{lcccc}
\toprule
& \multicolumn{2}{c}{Latent Cosine Similarity $\uparrow$} & \multicolumn{2}{c}{Latent MSE $\downarrow$} \\
\cmidrule(lr){2-3} \cmidrule(lr){4-5}
& Mean & Median & Mean & Median \\
\midrule
Replacement vs.\ original & 0.7839 & 0.7960 & 0.4786 & 0.4313 \\
\bottomrule
\end{tabular}
\end{table}

\begin{figure}[H]
\centering
\includegraphics[width=\linewidth]{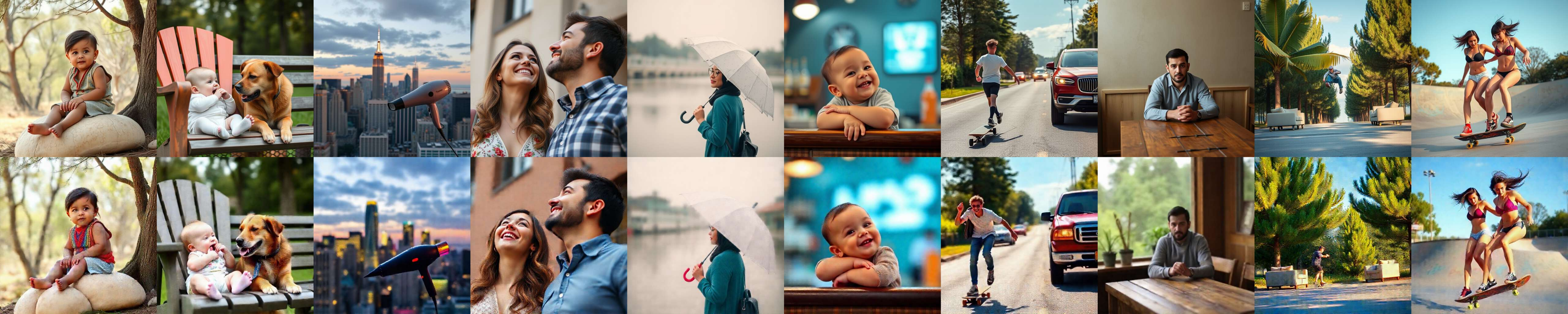}
\caption{Generated images for $10$ prompts at $512 \times 512$ and $4$ denoising steps. \textbf{Top row:} original FLUX.1[schnell]. \textbf{Bottom row:} full replacement model with all $32$ MLPs substituted by their transcoders.}
\label{fig:tc_e2e_grid}
\end{figure}

\paragraph{Visual comparison.} Figure~\ref{fig:tc_e2e_grid} shows generated images for $10$ prompts from the set, with the original model in the top row and the full replacement model in the bottom row. The replacement model recovers the global composition, object placements, broad shape outlines, and stylistic register of the original; deviations are concentrated in fine details (textures, small objects, sharp edges).

These results establish that the dictionaries learned by our transcoders are sufficiently faithful for circuit analysis: the transcoder-replaced model is not bit-exact with the original, but it generates qualitatively the same images on the same inputs.

\section{Local replacement model}
\label{app:lrm}

The local replacement model (LRM) takes the trained transcoders of §\ref{app:transcoders} and embeds them inside the base model in such a way that, on the cached prompt and timestep, the modified model's outputs exactly reproduce the originals up to floating-point error, while every interaction between transcoder features becomes linear under the assumption of a fixed active set.

\subsection{Cached quantities}
\label{app:lrm:cache}

The construction begins with a single forward pass of FLUX.1[schnell] on the chosen prompt at the chosen denoising step $t$. Forward hooks intercept and cache the following quantities, all per block $\ell \in \{0, \ldots, 15\}$ and stream $s \in \{\mathrm{img}, \mathrm{txt}\}$:

\begin{itemize}[leftmargin=*]
\item \emph{Boundary residual streams.} $r_0^{s} = x_{\mathrm{pre}}^{(0, s)}$, the residual stream entering block $0$ from each stream. For $s = \mathrm{img}$ this is the patch embedding of the noisy latent; for $s = \mathrm{txt}$ it is the projected T5 prompt embedding. These serve as the input layer of the LRM.

\item \emph{AdaLN modulation parameters.} The six per-(layer, stream) vectors $\mathrm{shift}^{\ell, s}_{\mathrm{msa}}$, $\mathrm{scale}^{\ell, s}_{\mathrm{msa}}$, $\mathrm{gate}^{\ell, s}_{\mathrm{msa}}$, $\mathrm{shift}^{\ell, s}_{\mathrm{mlp}}$, $\mathrm{scale}^{\ell, s}_{\mathrm{mlp}}$, $\mathrm{gate}^{\ell, s}_{\mathrm{mlp}} \in \mathbb{R}^{d_{\mathrm{model}}}$ produced by the AdaLayerNormZero modules in the block. These depend only on $t$ and the pooled CLIP embedding (on FLUX.1[dev] also on the guidance embedding, which is fixed for a given guidance scale), so they are constants of the LRM.

\item \emph{LayerNorm denominators.} For both the inner LayerNorm of \texttt{norm1}/\texttt{norm1\_context} (the parameter-free LayerNorm wrapped by AdaLayerNormZero, applied to the residual before joint attention) and \texttt{norm2}/\texttt{norm2\_context} (the parameter-free LayerNorm applied to the residual before the MLP), the cached inverse denominator $1 / \sqrt{\mathrm{Var}(x) + \varepsilon}$ at each token position. The mean is recomputed at runtime (a linear operation in $x$); only the denominator is frozen.

\item \emph{Joint attention probabilities and reconstruction error.} The attention probability tensor $P^\ell \in \mathbb{R}^{B \times H \times (S_{\mathrm{txt}} + S_{\mathrm{img}}) \times (S_{\mathrm{txt}} + S_{\mathrm{img}})}$, computed from the modulated $Q$ and $K$ projections of both streams concatenated along the token axis with text first, image second, and a per-stream attention reconstruction error
\begin{equation}
\varepsilon^{\ell, s}_{\mathrm{attn}} = \mathrm{attn}^\ell_{\mathrm{orig}, s} - W_O^{(\ell, s)}\!\bigl((P^\ell V^\ell)_s\bigr),
\end{equation}
where $V^\ell$ is the cached concatenation of the two streams' $V$-projections, $(P^\ell V^\ell)_s$ is the per-stream slice of the attention output along the token axis, and $W_O^{(\ell, s)}$ is the corresponding per-stream output projection. The error $\varepsilon^{\ell, s}_{\mathrm{attn}}$ accounts for the small numerical discrepancy between the original attention output and the same quantity recomputed from cached probabilities and $V$-projections.

\item \emph{Per-block transcoder caches.} For each (layer, stream) pair, the input $x^{\ell, s}$ to the feed-forward sublayer, the activation vector $z^{\ell, s} = z(x^{\ell, s}, t)$, the preactivation vector $h^{\ell, s}_{\mathrm{pre}}$, and the MLP reconstruction residual
\begin{equation}
\varepsilon^{\ell, s}_{\mathrm{mlp}} = \mathrm{MLP}^s_\ell(x^{\ell, s}) - TC^s_\ell(x^{\ell, s}, t),
\end{equation}
where $TC^s_\ell(x, t) = W_{\mathrm{dec}}^{(\ell, s)} z^{\ell, s} + b_{\mathrm{dec}}^{(\ell, s)}$ is the full transcoder output including the decoder bias. The role of the decoder bias is discussed in §\ref{app:attribution:beff}.
\end{itemize}

\subsection{Component substitutions}
\label{app:lrm:substitution}

The base model is then re-run with the following per-block substitutions, applied to all $\ell \in \{0, \ldots, 15\}$ and both streams. Outside this range the original blocks are kept intact, and the LRM is therefore identical to the original model on blocks $16$--$18$ (double-stream) and on the $38$ single-stream blocks that follow.

\paragraph{LayerNorm.} In each analyzed block we replace four LayerNorm modules: the inner LayerNorm of \texttt{norm1} and \texttt{norm1\_context}, and \texttt{norm2}/\texttt{norm2\_context}. Each is replaced by
\begin{equation}
\mathrm{FrozenNorm}_{\ell, s}(x) = (x - \bar{x}) \odot \nu^{\ell, s}_{\mathrm{cached}},
\end{equation}
where $\bar{x} = \tfrac{1}{d_{\mathrm{model}}} \sum_j x_j$ is recomputed at runtime and $\nu^{\ell, s}_{\mathrm{cached}} = 1 / \sqrt{\mathrm{Var}(x_{\mathrm{cached}}) + \varepsilon}$ is the inverse denominator from §\ref{app:lrm:cache}. Mean subtraction is linear in $x$, so the only nonlinear component of LayerNorm has been removed from the LRM. The AdaLayerNormZero wrapper around \texttt{norm1} continues to apply its scale-and-shift modulation around the frozen inner LayerNorm; only the LayerNorm denominator is frozen, not the modulation itself.

\paragraph{Joint attention.} The full joint attention block, including its $Q$ and $K$ projections, scaled dot product, softmax, and stream concatenation, is replaced by a per-stream affine function of the recomputed $V$-projections, with the cached probabilities as fixed weights:
\begin{equation}
\mathrm{FrozenAttn}_\ell(x_{\mathrm{img}}, x_{\mathrm{txt}})_s = W_O^{(\ell, s)}\!\bigl((P^\ell\, V_{\mathrm{cat}}^\ell(x_{\mathrm{img}}, x_{\mathrm{txt}}))_s\bigr) + \varepsilon^{\ell, s}_{\mathrm{attn}}, \qquad s \in \{\mathrm{img}, \mathrm{txt}\}.
\end{equation}
Here $V_{\mathrm{cat}}^\ell(x_{\mathrm{img}}, x_{\mathrm{txt}})$ concatenates the two streams' $V$-projections along the token axis ($V_{\mathrm{txt}}$ first, then $V_{\mathrm{img}}$, matching the original implementation), $P^\ell$ is the cached probability tensor, $(\cdot)_s$ extracts the per-stream slice along the token axis, and $W_O^{(\ell, s)}$ is the per-stream output projection that follows. The split between streams happens before the output projection, exactly as in the original implementation, and each stream uses its own $W_O$. Crucially, the $V$-projection still depends on the input residual streams (it is an affine operation, $V = W_V x + b_V$, applied to the AdaLN-modulated input $x$); only the $Q$-$K$ pathway through the softmax has been frozen. The reconstruction residual $\varepsilon^{\ell, s}_{\mathrm{attn}}$ ensures that on the cached input, $\mathrm{FrozenAttn}_\ell(x^{\mathrm{cached}}_{\mathrm{img}}, x^{\mathrm{cached}}_{\mathrm{txt}})_s$ matches the original attention output to floating-point precision.

\paragraph{MLP.} Each feed-forward sublayer is replaced by its transcoder plus the cached MLP reconstruction residual,
\begin{equation}
\mathrm{MLP}^{\mathrm{LRM}}_{\ell, s}(x) = TC^s_\ell(x, t) + \varepsilon^{\ell, s}_{\mathrm{mlp}}.
\end{equation}
On the cached input $x = x^{\ell, s}$ this is exact by definition of $\varepsilon^{\ell, s}_{\mathrm{mlp}}$.

\subsection{Linearization shortcut}
\label{app:lrm:shortcut}

The LRM is used in two regimes. In \emph{validation mode} (§\ref{app:lrm:validation}) we want the LRM's output as a function of its input, so the transcoders are run forward in the standard way. In \emph{tracing mode} (used to compute attribution edges, §\ref{app:attribution}) we run the LRM only on the cached prompt and only need it as an affine function of the source feature activations on that prompt; we therefore apply two simplifications.

First, in tracing mode the MLP substitution becomes
\begin{equation}
\mathrm{MLP}^{\mathrm{LRM}}_{\ell, s}(x) = y^{\ell, s}_{\mathrm{cached}},
\end{equation}
that is, we return the cached original MLP output directly without running the transcoder. On the cached input this is exact: by definition of $\varepsilon^{\ell, s}_{\mathrm{mlp}}$ we have $TC^s_\ell(x^{\ell, s}_{\mathrm{cached}}, t) + \varepsilon^{\ell, s}_{\mathrm{mlp}} = y^{\ell, s}_{\mathrm{cached}}$. The shortcut avoids a full transcoder forward pass per block and lets the transcoder weights be moved off-GPU during tracing; the per-target backward pass uses only the cached activations $z^{\ell, s}$ and decoder weights $W^{(\ell, s)}_{\mathrm{dec}}$.

Second, when computing the target preactivation $h^*$ as a function of source activations, the cached MLP outputs are returned as gradient-free constants. Gradients in the backward pass therefore flow only through residual connections and the affine OV pathway of frozen attention, which is precisely the linearization we want: each source feature contributes through its decoder vector being added to the residual stream and read out by the target's encoder vector.

\subsection{Validation of the LRM}
\label{app:lrm:validation}

The LRM is by construction exact on the cached prompt and timestep up to floating-point error. We validate that this is the case in practice and quantify the magnitude of the residual numerical drift.

\paragraph{Frozen attention numerical accuracy.} The attention reconstruction error $\varepsilon^{\ell, s}_{\mathrm{attn}}$ is defined as the difference between the original attention output and the same quantity recomputed from cached $P^\ell$ and $V$-projections. Although the recomputation is mathematically identical to the original, the two differ at the level of bfloat16 round-off, because the original attention runs through a fused CUDA kernel with a different reduction order than our explicit $W_O (P V)$ recomputation. These residuals are absorbed into the LRM as additive corrections (§\ref{app:lrm:substitution}), and into $b^*_{\mathrm{eff}}$ in the attribution graph (§\ref{app:attribution:beff}).

\paragraph{End-to-end LRM exactness.} On a set of $512$ prompts, for each of the $4$ denoising steps separately, we generate the final flat latent under the original model and under the LRM with all $16$ analyzed blocks substituted. Table~\ref{tab:lrm_exactness} reports the latent cosine similarity and the latent MSE between the two for each step. Mean cosine similarity is around $0.99$ across all four steps, ranging from $0.9854$ at $t = 0$ to effectively $1.0$ at $t = 3$. The trend across rows reflects how floating-point drift propagates through subsequent denoising steps: an LRM substitution at an earlier step is followed by additional original-model steps, each of which can amplify the residual numerical error, while a substitution at the final step ($t = 3$) is not propagated further and produces near-bit-exact agreement with the original model.

\begin{table}[H]
\centering
\caption{End-to-end LRM exactness against the original FLUX.1[schnell] on $512$ prompts. Each row corresponds to building the LRM at a single denoising step $t$ and replacing only that step's transformer call.}
\label{tab:lrm_exactness}
\begin{tabular}{lcccc}
\toprule
& \multicolumn{2}{c}{Latent Cosine Similarity $\uparrow$} & \multicolumn{2}{c}{Latent MSE $\downarrow$} \\
\cmidrule(lr){2-3} \cmidrule(lr){4-5}
Denoising step & Mean & Median & Mean & Median \\
\midrule
$t = 0$ & 0.9854 & 0.9889 & $2.5 \times 10^{-2}$ & $2.0 \times 10^{-2}$ \\
$t = 1$ & 0.9982 & 0.9986 & $3.1 \times 10^{-3}$ & $2.6 \times 10^{-3}$ \\
$t = 2$ & 0.9997 & 0.9997 & $5.4 \times 10^{-4}$ & $4.9 \times 10^{-4}$ \\
$t = 3$ & 1.0000 & 1.0000 & $7.9 \times 10^{-5}$ & $7.4 \times 10^{-5}$ \\
\bottomrule
\end{tabular}
\end{table}

\begin{figure}[H]
\centering
\includegraphics[width=0.75\linewidth]{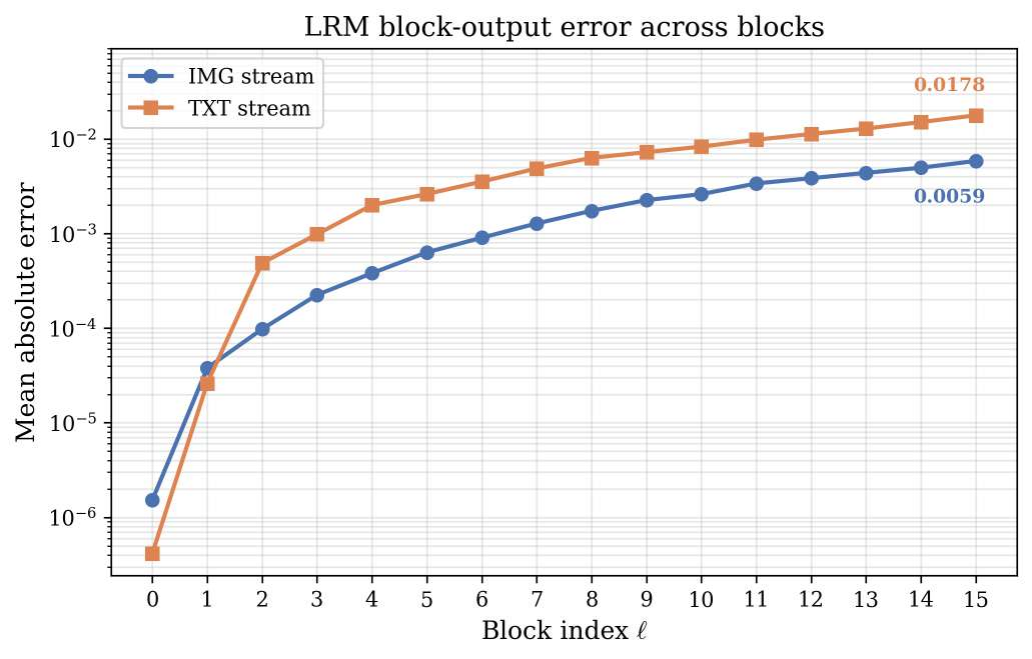}
\caption{Mean absolute error between the original model's block output and the LRM's block output at each of the analyzed blocks, for the image and text streams. Error grows monotonically with depth as floating-point discrepancies accumulate, but stays within numerical-precision range across all $16$ LRM blocks.}
\label{fig:lrm_drift}
\end{figure}

\paragraph{Compounding floating-point drift across blocks.} Although the LRM is exact at each individual block on the cached input, when the LRM is run forward the input to block $\ell + 1$ in the LRM is no longer exactly equal to the cached input to block $\ell + 1$ in the original model: it differs by the per-block floating-point error of all preceding blocks. This drift is small in absolute terms but grows monotonically with depth. Figure~\ref{fig:lrm_drift} plots, for a representative prompt and denoising step, the mean absolute error between the original block output and the LRM block output at each $\ell \in \{0, \ldots, 15\}$, separately for the two streams. Both curves are monotone in $\ell$, but the maximum mean absolute error at the deepest analyzed block is $5.86 \times 10^{-3}$ for the image stream and $1.78 \times 10^{-2}$ for the text stream. This confirms that drift remains bounded throughout depth and does not affect downstream behavior at the latent level (Table~\ref{tab:lrm_exactness}).

\begin{figure}[H]
\centering
\includegraphics[width=\linewidth]{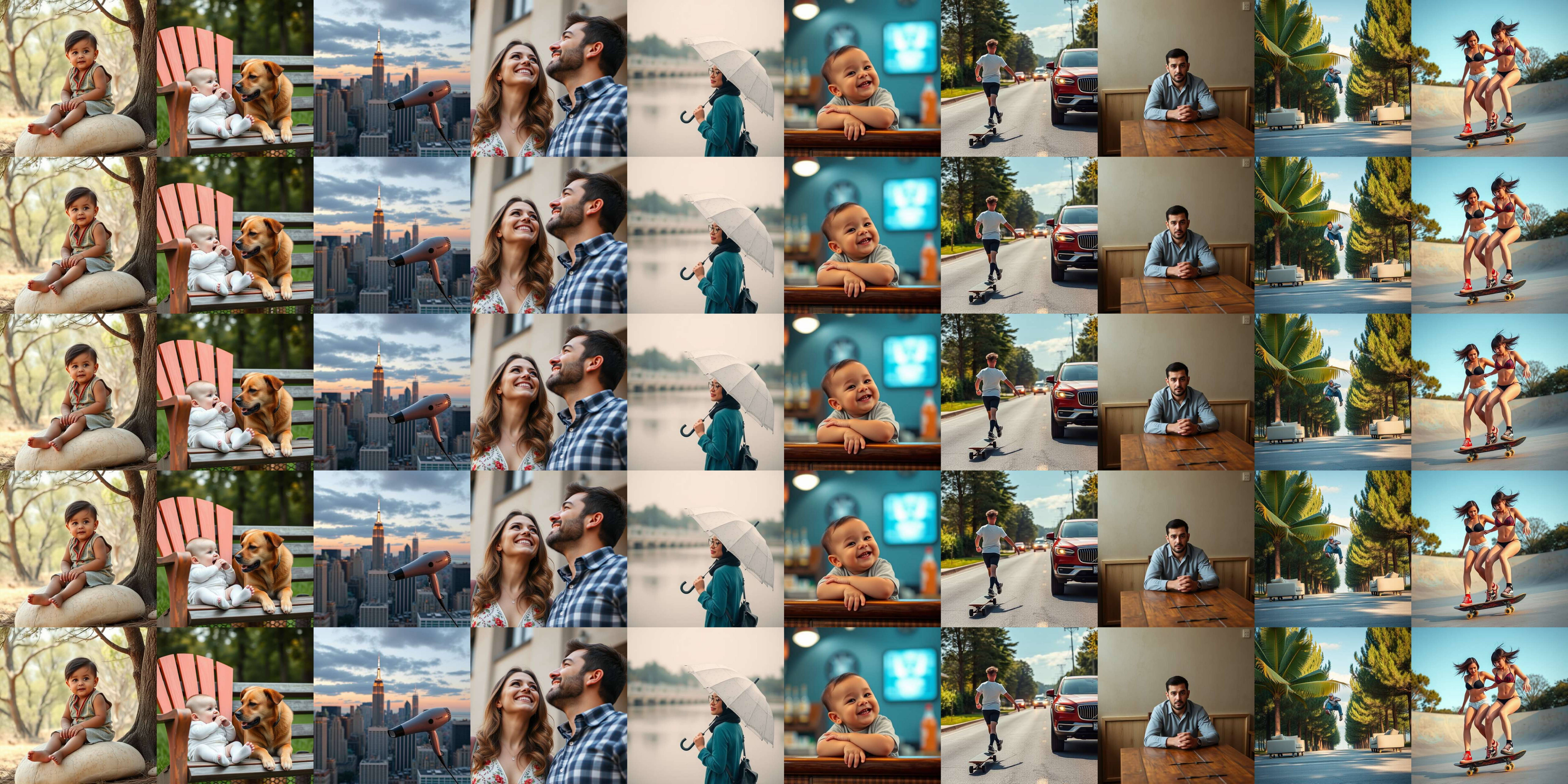}
\caption{Generated images for $10$ prompts. \textbf{Row 1:} original FLUX.1[schnell]. \textbf{Rows 2--5:} LRM applied only at step $t = 0, 1, 2, 3$ respectively, with the other steps run by the original model.}
\label{fig:lrm_e2e_grid}
\end{figure}

\paragraph{Visual comparison.} Figure~\ref{fig:lrm_e2e_grid} shows the qualitative effect of substituting the LRM at each of the four denoising steps separately; the generated images remain effectively indistinguishable from the originals.

\section{Attribution graph}
\label{app:attribution}

This section gives the complete derivation of the attribution graph from the LRM of §\ref{app:lrm}. We begin by writing the target preactivation $h^*$ as a fully expanded affine function of the cached residual stream and a collection of constants (§\ref{app:attribution:hstar}), separate the input-independent part into the effective bias $b^*_{\mathrm{eff}}$ (§\ref{app:attribution:beff}), and then derive the per-edge attribution formulas for feature, error, and input source vertices (§\ref{app:attribution:edges}). The conservation invariant $h^* - b^*_{\mathrm{eff}} = \sum_{\mathrm{src}} A_{\mathrm{src} \to f^*}$ follows by construction (§\ref{app:attribution:invariant}), and we close with a discussion of cross-stream edges and of what the graph does \emph{not} model (§§\ref{app:attribution:cross}--\ref{app:attribution:notmodeled}).

\subsection{Target preactivation as an affine function}
\label{app:attribution:hstar}

Fix a prompt, a denoising step $t$, and a target feature $f^*$ characterized by $(\ell^*, s^*, p^*, i^*)$. Write $r^{\ell, s}_p \in \mathbb{R}^{d_{\mathrm{model}}}$ for the residual stream of stream $s$ at position $p$ on entry to block $\ell$ in the LRM, so that $r^{0, s}_p = r^s_0(p)$ is the input embedding. The target preactivation is
\begin{equation}
h^* = \bigl(f^{(\ell^*, s^*, i^*)}_{\mathrm{enc}}\bigr)^{\!\top} x^{\ell^*, s^*}_{\mathrm{mod}}(p^*) + \bigl(b^{(\ell^*, s^*)}_{\mathrm{enc}}\bigr)_{i^*},
\label{eq:hstar_top}
\end{equation}
where $x^{\ell^*, s^*}_{\mathrm{mod}}(p^*)$ is the FiLM-modulated FF input to the target transcoder. We expand this quantity in two stages.

\paragraph{From mid-block residual to FF input.} The MLP sublayer of block $\ell^*$ reads the residual stream after the attention update of that same block, which we denote $x^{\ell^*, s^*}_{\mathrm{mid}}(p^*)$. Concretely,
\begin{equation}
x^{\ell^*, s^*}_{\mathrm{mid}}(p^*) = r^{\ell^*, s^*}_{p^*} + \mathrm{gate}^{\ell^*, s^*}_{\mathrm{msa}} \odot \mathrm{FrozenAttn}_{\ell^*}(\cdots)_{s^*}(p^*),
\end{equation}
which is itself affine in $r^{\ell^*, s^*}$ (and, via the cross-stream attention, in $r^{\ell^*, s'}$ for $s' \neq s^*$). The FF input is then obtained from $x^{\ell^*, s^*}_{\mathrm{mid}}(p^*)$ by frozen LayerNorm followed by AdaLN-Zero modulation:
\begin{equation}
x^{\ell^*, s^*}_{\mathrm{ff}}(p^*) = \mathrm{FrozenNorm}_{\ell^*, s^*}\!\bigl(x^{\ell^*, s^*}_{\mathrm{mid}}(p^*)\bigr) \odot \bigl(1 + \mathrm{scale}^{\ell^*, s^*}_{\mathrm{mlp}}\bigr) + \mathrm{shift}^{\ell^*, s^*}_{\mathrm{mlp}}.
\end{equation}
Both $\mathrm{scale}^{\ell^*, s^*}_{\mathrm{mlp}}$ and $\mathrm{shift}^{\ell^*, s^*}_{\mathrm{mlp}}$ are constants of the LRM. The same affineness extends to the residual streams entering all blocks $\ell < \ell^*$, since every component of the LRM up to that point is either affine or treats its nonlinearities as fixed (frozen norm denominators, frozen attention probabilities, fixed transcoder active sets).

\paragraph{From FF input to encoder input.} Inside the target transcoder, $x^{\ell^*, s^*}_{\mathrm{ff}}(p^*)$ is further modulated by FiLM:
\begin{equation}
x^{\ell^*, s^*}_{\mathrm{mod}}(p^*) = x^{\ell^*, s^*}_{\mathrm{ff}}(p^*) \odot \bigl(1 + \mathrm{scale}^{\mathrm{tc}}_{\ell^*, s^*}(t)\bigr) + \mathrm{shift}^{\mathrm{tc}}_{\ell^*, s^*}(t),
\end{equation}
where $\mathrm{scale}^{\mathrm{tc}}_{\ell^*, s^*}(t)$ and $\mathrm{shift}^{\mathrm{tc}}_{\ell^*, s^*}(t)$ are constants once $t$ is fixed.

Therefore,
\begin{equation}
x^{\ell^*, s^*}_{\mathrm{mod}}(p^*) = \mathrm{FrozenNorm}\!\bigl(x^{\ell^*, s^*}_{\mathrm{mid}}(p^*)\bigr) \odot c_1 + c_2,
\end{equation}
with
\begin{align}
c_1 &= \bigl(1 + \mathrm{scale}^{\ell^*, s^*}_{\mathrm{mlp}}\bigr) \odot \bigl(1 + \mathrm{scale}^{\mathrm{tc}}_{\ell^*, s^*}(t)\bigr), \\
c_2 &= \mathrm{shift}^{\ell^*, s^*}_{\mathrm{mlp}} \odot \bigl(1 + \mathrm{scale}^{\mathrm{tc}}_{\ell^*, s^*}(t)\bigr) + \mathrm{shift}^{\mathrm{tc}}_{\ell^*, s^*}(t),
\end{align}
both $c_1$ and $c_2$ constants of the LRM.

Plugging into~(\ref{eq:hstar_top}),
\begin{equation}
h^* = \bigl(f^{(\ell^*, s^*, i^*)}_{\mathrm{enc}}\bigr)^{\!\top} \bigl(\mathrm{FrozenNorm}\!\bigl(x^{\ell^*, s^*}_{\mathrm{mid}}(p^*)\bigr) \odot c_1\bigr) + \underbrace{\bigl(f^{(\ell^*, s^*, i^*)}_{\mathrm{enc}}\bigr)^{\!\top} c_2 + \bigl(b^{(\ell^*, s^*)}_{\mathrm{enc}}\bigr)_{i^*}}_{\text{input-independent}}.
\label{eq:hstar_split}
\end{equation}
The first term is affine in $x^{\ell^*, s^*}_{\mathrm{mid}}$, which is itself affine in all upstream sources. The second term is constant.

\subsection{Effective encoder bias}
\label{app:attribution:beff}

The constant part of $h^*$ has further contributions that we have not yet made explicit. The post-attention residual $x^{\ell^*, s^*}_{\mathrm{mid}}(p^*)$ read by the target is the sum of the input embedding, of the contributions of every sublayer of the blocks $\ell < \ell^*$ in both streams, and of the attention sublayer of block $\ell^*$ itself. Among these contributions are several that are constant in the LRM:

\begin{itemize}[leftmargin=*]
\item Each attention sublayer with $\ell \le \ell^*$ contributes $\mathrm{gate}^{\ell, s}_{\mathrm{msa}} \odot \mathrm{FrozenAttn}_\ell(\cdots)_s$ to the residual at every position. The frozen attention output decomposes as $W_O^{(\ell, s)}\!\bigl((P^\ell V^\ell)_s\bigr) + \varepsilon^{\ell, s}_{\mathrm{attn}}$. The first term is affine, not linear, in the input: the value and output projections carry biases ($V = W_V a + b_V$, and $W_O^{(\ell, s)}$ adds $b^{(\ell, s)}_O$), and the attention input $a = \mathrm{FrozenNorm}(r) \odot (1 + \mathrm{scale}_{\mathrm{msa}}) + \mathrm{shift}_{\mathrm{msa}}$ carries the AdaLN shift. It therefore splits into an input-dependent part, which is attributed to source vertices through the gradient (§\ref{app:attribution:edges}), and a constant part, obtained by setting the residual streams to zero, where $\mathrm{FrozenNorm}(0) = 0$:
\begin{equation}
\kappa^{\ell, s}(p) \;=\; W_O^{(\ell, s)}\!\Bigl(\bigl(P^\ell\, V^\ell_{\mathrm{cat}}\bigl(\mathrm{shift}^{\ell, \mathrm{img}}_{\mathrm{msa}}, \mathrm{shift}^{\ell, \mathrm{txt}}_{\mathrm{msa}}\bigr)\bigr)_s(p)\Bigr),
\label{eq:kappa_ov}
\end{equation}
where $V^\ell_{\mathrm{cat}}(\cdot, \cdot)$ concatenates the two streams' $V$-projections (biases included) of their AdaLN shifts, broadcast over positions, and $W_O^{(\ell, s)}$ includes its bias. Since $P^\ell$ is cached, $\kappa^{\ell, s}(p)$ is a constant of the LRM; it varies over positions only through the attention mass that $P^\ell$ assigns to each stream. The second term, $\varepsilon^{\ell, s}_{\mathrm{attn}}$, is the cached attention reconstruction error and is likewise constant.

\item Each MLP sublayer with $\ell < \ell^*$ contributes $\mathrm{gate}^{\ell, s}_{\mathrm{mlp}} \odot (TC^s_\ell(x, t) + \varepsilon^{\ell, s}_{\mathrm{mlp}})$. The transcoder output further decomposes as $\sum_i z_i^{(\ell, s)} f^{(\ell, s, i)}_{\mathrm{dec}} + b_{\mathrm{dec}}^{(\ell, s)}$. The first sum is affine in feature activations; the decoder bias $b_{\mathrm{dec}}^{(\ell, s)}$ and the cached residual $\varepsilon^{\ell, s}_{\mathrm{mlp}}$ are constants.
\end{itemize}

The constant terms $\kappa^{\ell, s}$, $\varepsilon^{\ell, s}_{\mathrm{attn}}$ and $b^{(\ell, s)}_{\mathrm{dec}}$ propagate forward through the LRM, are gated by the corresponding AdaLN gates, and accumulate into the constant part of (\ref{eq:hstar_split}). A sublayer of a block $\ell < \ell^*$ reaches $h^*$ through the residual stream, so its contribution is read off the backward pass of $h^*$ (§\ref{app:attribution:edges}) with the gradient $g^{\ell + 1, s}(p)$. The attention sublayer of the target block itself writes into $x^{\ell^*, s^*}_{\mathrm{mid}}(p^*)$ after the last residual-stream gradient; its multiplier is the sensitivity of $h^*$ to the post-attention residual at the target position, which follows from (\ref{eq:hstar_split}) because the frozen LayerNorm is linear:
\begin{equation}
m^* \;=\; \frac{\partial h^*}{\partial x^{\ell^*, s^*}_{\mathrm{mid}}(p^*)} \;=\; \nu^{\ell^*, s^*}_{\mathrm{cached}}(p^*)\,\Bigl(I - \tfrac{1}{d_{\mathrm{model}}}\mathbf{1}\mathbf{1}^{\!\top}\Bigr)\bigl(f^{(\ell^*, s^*, i^*)}_{\mathrm{enc}} \odot c_1\bigr),
\label{eq:mstar}
\end{equation}
where $\nu^{\ell^*, s^*}_{\mathrm{cached}}(p^*)$ is the cached inverse denominator of the LayerNorm before the MLP of block $\ell^*$ at position $p^*$ (§\ref{app:lrm:cache}). Only the target's own stream and position enter at $\ell = \ell^*$. Collecting terms gives the closed forms
\begin{align}
\beta_{\mathrm{attn}} &= \sum_{\ell < \ell^*}\, \sum_{s \in \{\mathrm{img}, \mathrm{txt}\}}\, \sum_p \bigl\langle \mathrm{gate}^{\ell, s}_{\mathrm{msa}} \odot \varepsilon^{\ell, s}_{\mathrm{attn}}(p),\ g^{\ell + 1, s}(p) \bigr\rangle \;+\; \bigl\langle \mathrm{gate}^{\ell^*, s^*}_{\mathrm{msa}} \odot \varepsilon^{\ell^*, s^*}_{\mathrm{attn}}(p^*),\ m^* \bigr\rangle, \\
\beta_{\mathrm{ov}} &= \sum_{\ell < \ell^*}\, \sum_{s \in \{\mathrm{img}, \mathrm{txt}\}}\, \sum_p \bigl\langle \mathrm{gate}^{\ell, s}_{\mathrm{msa}} \odot \kappa^{\ell, s}(p),\ g^{\ell + 1, s}(p) \bigr\rangle \;+\; \bigl\langle \mathrm{gate}^{\ell^*, s^*}_{\mathrm{msa}} \odot \kappa^{\ell^*, s^*}(p^*),\ m^* \bigr\rangle, \label{eq:beta_ov} \\
\beta_{\mathrm{dec}} &= \sum_{\ell < \ell^*}\, \sum_{s \in \{\mathrm{img}, \mathrm{txt}\}}\, \sum_p \bigl\langle \mathrm{gate}^{\ell, s}_{\mathrm{mlp}} \odot b^{(\ell, s)}_{\mathrm{dec}},\ g^{\ell + 1, s}(p) \bigr\rangle,
\end{align}
where $g^{\ell + 1, s}(p) \in \mathbb{R}^{d_{\mathrm{model}}}$ is the gradient of $h^*$ with respect to the residual stream of stream $s$ at position $p$ on entry to block $\ell + 1$, computed by the linearized backward pass described in §\ref{app:attribution:edges}. All three are constants of the LRM, since neither $\kappa^{\ell, s}$, $\varepsilon^{\ell, s}_{\mathrm{attn}}$, $b^{(\ell, s)}_{\mathrm{dec}}$, the AdaLN gates, $m^*$ nor the gradients $g^{\ell + 1, s}$ depend on any source feature activation under the fixed-active-set assumption.

The complete effective bias is then
\begin{equation}
\begin{aligned}
b^*_{\mathrm{eff}} \;=\; & \bigl(b^{(\ell^*, s^*)}_{\mathrm{enc}}\bigr)_{i^*} \\
& {} + \bigl(f^{(\ell^*, s^*, i^*)}_{\mathrm{enc}}\bigr)^{\!\top} \mathrm{shift}^{\mathrm{tc}}_{\ell^*, s^*}(t) \\
& {} + \bigl(f^{(\ell^*, s^*, i^*)}_{\mathrm{enc}}\bigr)^{\!\top}\, \bigl(\mathrm{shift}^{\ell^*, s^*}_{\mathrm{mlp}} \odot (1 + \mathrm{scale}^{\mathrm{tc}}_{\ell^*, s^*}(t))\bigr) \\
& {} + \beta_{\mathrm{attn}} + \beta_{\mathrm{ov}} + \beta_{\mathrm{dec}}.
\end{aligned}
\label{eq:beff_full}
\end{equation}
The first three terms come from (\ref{eq:hstar_split}): the encoder bias of the target feature, the FiLM shift propagated through the encoder, and the AdaLN MLP shift propagated first through FiLM and then through the encoder. The remaining terms are $\beta_{\mathrm{attn}}$, $\beta_{\mathrm{ov}}$ and $\beta_{\mathrm{dec}}$. The $\ell = \ell^*$ parts of $\beta_{\mathrm{attn}}$ and $\beta_{\mathrm{ov}}$ account for the target block's own attention, which enters $h^*$ without passing through any residual-stream gradient; for a target in block $0$ they are the only attention terms. By construction, $h^* - b^*_{\mathrm{eff}}$ is exactly the input-dependent part of (\ref{eq:hstar_split}) plus the input-dependent contributions of all upstream sources. On SD3.5-medium, whose blocks $0$--$12$ have a second, image-only attention sublayer, that sublayer contributes terms of the same form to $\beta_{\mathrm{attn}}$ and $\beta_{\mathrm{ov}}$ through its own gate, including at the target block for an image-stream target (Appendix~\ref{app:backbones:arch}).

\paragraph{Why the decoder bias is moved into $b^*_{\mathrm{eff}}$.} An alternative bookkeeping would treat $b^{(\ell, s)}_{\mathrm{dec}}$ as part of the MLP reconstruction residual by defining $\tilde{\varepsilon}^{\ell, s}_{\mathrm{mlp}} = \mathrm{MLP}^s_\ell(x) - W_{\mathrm{dec}}^{(\ell, s)} z^{\ell, s}$, so that error vertices carry $\tilde{\varepsilon}$ rather than $\varepsilon$. This is mathematically equivalent: it just folds $b^{(\ell, s)}_{\mathrm{dec}}$ from $\beta_{\mathrm{dec}}$ into the error edges. We prefer the present arrangement because $\varepsilon^{\ell, s}_{\mathrm{mlp}}$ then represents only the genuinely residual variance that the transcoder failed to capture, which is the quantity one wants to monitor as a measure of transcoder quality.

\subsection{Edge attributions}
\label{app:attribution:edges}

To compute the contribution of each source vertex we run a single backward pass of $h^*$ through the LRM in tracing mode (§\ref{app:lrm:shortcut}), in which all transcoder outputs are detached and gradients flow only through residual connections and the affine OV pathway of frozen attention. Denote by
\begin{equation}
g^{\ell, s}(p) \;=\; \frac{\partial h^*}{\partial r^{\ell, s}_p} \in \mathbb{R}^{d_{\mathrm{model}}}
\end{equation}
the gradient of $h^*$ with respect to the residual stream of stream $s$ at position $p$ on entry to block $\ell$, computed in the linearized LRM. Since gradients do not flow through MLP outputs, $g^{\ell, s}$ depends only on cached attention probabilities, frozen LayerNorm denominators, AdaLN-Zero modulation parameters, the target transcoder's FiLM scale $\mathrm{scale}^{\mathrm{tc}}_{\ell^*, s^*}(t)$, and the target's encoder vector $f^{(\ell^*, s^*, i^*)}_{\mathrm{enc}}$; it is fully determined by the cached forward pass and is therefore a constant of the LRM under the fixed-active-set assumption.

\paragraph{Feature edges.} A source feature at $(\ell, s, p, i)$ with activation $z^{(\ell, s, i)}(p)$ writes the vector $z^{(\ell, s, i)}(p)\, f^{(\ell, s, i)}_{\mathrm{dec}} \in \mathbb{R}^{d_{\mathrm{model}}}$ into the MLP output of stream $s$ at position $p$. This output is multiplied by the AdaLN gate $\mathrm{gate}^{\ell, s}_{\mathrm{mlp}}$ and added to the residual stream entering block $\ell + 1$. By the chain rule and the linearity of the LRM along this path, its contribution to $h^*$ is
\begin{equation}
A_{(\ell, s, p, i) \to f^*} \;=\; \underbrace{z^{(\ell, s, i)}(p)\vphantom{\bigl(g^{\ell + 1, s}(p)\bigr)}}_{\text{input-dependent}} \cdot \underbrace{\bigl(g^{\ell + 1, s}(p) \odot \mathrm{gate}^{\ell, s}_{\mathrm{mlp}}\bigr)^{\!\top} f^{(\ell, s, i)}_{\mathrm{dec}}}_{\text{virtual weight}}.
\label{eq:feat_edge}
\end{equation}
The input-dependent factor is the activation; the virtual weight depends on the cached forward pass through $g^{\ell + 1, s}(p)$ and on the input-invariant decoder vector. The factor $\mathrm{gate}^{\ell, s}_{\mathrm{mlp}} \in \mathbb{R}^{d_{\mathrm{model}}}$ reflects FLUX's AdaLN-Zero gating of the MLP output before the residual add and is constant across positions for fixed $(\ell, s)$.

\paragraph{Error edges.} The MLP reconstruction residual $\varepsilon^{\ell, s}_{\mathrm{mlp}}(p)$ enters the residual stream through the same gating as the transcoder output, so its contribution to $h^*$ is
\begin{equation}
A_{(\ell, s, p)_{\mathrm{err}} \to f^*} \;=\; \bigl(\varepsilon^{\ell, s}_{\mathrm{mlp}}(p) \odot \mathrm{gate}^{\ell, s}_{\mathrm{mlp}}\bigr)^{\!\top} g^{\ell + 1, s}(p).
\label{eq:err_edge}
\end{equation}
Unlike feature edges, error edges have no input-dependent factor: $\varepsilon^{\ell, s}_{\mathrm{mlp}}(p)$ is a cached constant. Each error vertex thus carries a single scalar attribution.

\paragraph{Input edges.} For each input position $(s, p)$, the embedding $r^s_0(p) \in \mathbb{R}^{d_{\mathrm{model}}}$ enters block $0$ directly, with no further gating. Its contribution is
\begin{equation}
A_{(s, p)_{\mathrm{in}} \to f^*} \;=\; r^s_0(p)^{\!\top}\, g^{0, s}(p).
\label{eq:in_edge}
\end{equation}

\paragraph{Implementation as a single backward pass.} In practice we compute all three edge types from a single VJP. The pipeline is summarized in Algorithm~\ref{alg:edge_extraction}.

\begin{algorithm}[H]
\caption{Per-target edge extraction in the LRM.}
\label{alg:edge_extraction}
\begin{algorithmic}[1]
\REQUIRE Cached forward state of the LRM; target $f^* = (\ell^*, s^*, p^*, i^*)$; threshold $\tau$.
\ENSURE Edge set $\mathcal{E}$ with attributions for all sources whose $|A| \ge \tau$ (for feature sources also $z > \tau$).
\STATE Compute the target preactivation $h^* = (f^{(\ell^*, s^*, i^*)}_{\mathrm{enc}})^{\!\top}\, x^{\ell^*, s^*}_{\mathrm{mod}}(p^*) + (b^{(\ell^*, s^*)}_{\mathrm{enc}})_{i^*}$ and effective bias $b^*_{\mathrm{eff}}$ via Eq.~(\ref{eq:beff_full}).

\STATE Run a backward pass of $h^*$ through the LRM in tracing mode (transcoder outputs detached). Cache the gradients $\{g^{\ell, s}(p)\}$ for $\ell \in \{0, \ldots, \ell^*\}$, $s \in \{\mathrm{img}, \mathrm{txt}\}$, all $p$, including the boundary gradient $g^{0, s}(p) = \partial h^* / \partial r^s_0(p)$ used for input edges.

\STATE Initialize $\mathcal{E} \gets \emptyset$.

\FOR{each $(\ell, s)$ with $\ell < \ell^*$}
    \STATE Read cached activations $z^{\ell, s} \in \mathbb{R}^{S_s \times d_{\mathrm{feat}}}$ and decoder $W^{(\ell, s)}_{\mathrm{dec}} \in \mathbb{R}^{d_{\mathrm{model}} \times d_{\mathrm{feat}}}$.
    \STATE Form $V^{\ell, s} \in \mathbb{R}^{S_s \times d_{\mathrm{model}}}$ by elementwise multiplying $g^{\ell + 1, s}$ by $\mathrm{gate}^{\ell, s}_{\mathrm{mlp}}$ across positions.
    \STATE Compute feature attributions $\mathbf{A}^{\ell, s}_{\mathrm{feat}} \gets z^{\ell, s} \odot (V^{\ell, s}\, W^{(\ell, s)}_{\mathrm{dec}})$ for all $i, p$.
    \STATE Insert into $\mathcal{E}$ all $(\ell, s, p, i)$ with $z^{\ell, s}(p, i) > \tau$ and $|A^{\ell, s}_{\mathrm{feat}}(p, i)| \ge \tau$.
    \STATE Compute error attributions $A^{\ell, s}_{\mathrm{err}}(p) \gets (\varepsilon^{\ell, s}_{\mathrm{mlp}}(p) \odot \mathrm{gate}^{\ell, s}_{\mathrm{mlp}})^{\!\top} g^{\ell + 1, s}(p)$ for all $p$.
    \STATE Insert into $\mathcal{E}$ all $(\ell, s, p)_{\mathrm{err}}$ with $|A^{\ell, s}_{\mathrm{err}}(p)| \ge \tau$.
\ENDFOR

\FOR{each $s \in \{\mathrm{img}, \mathrm{txt}\}$}
    \STATE Compute input attributions $A^s_{\mathrm{in}}(p) \gets r^s_0(p)^{\!\top}\, g^{0, s}(p)$ for all $p$.
    \STATE Insert into $\mathcal{E}$ all $(s, p)_{\mathrm{in}}$ with $|A^s_{\mathrm{in}}(p)| \ge \tau$.
\ENDFOR

\RETURN $\mathcal{E}$.
\end{algorithmic}
\end{algorithm}

A single VJP from $h^*$ thus suffices to extract all incoming edges to the target. The total cost is dominated by the matrix multiplications in the loop, which scale linearly in the number of layers and in $d_{\mathrm{feat}}$.

\subsection{Conservation invariant}
\label{app:attribution:invariant}

Combining (\ref{eq:hstar_split}), the propagation of $r^{\ell^*, s^*}_{p^*}$ through the LRM, and the closed forms for $\beta_{\mathrm{attn}}, \beta_{\mathrm{ov}}, \beta_{\mathrm{dec}}$, the input-dependent part of $h^*$ is exactly the sum of all source contributions:
\begin{equation}
h^* - b^*_{\mathrm{eff}} = \sum_{\mathrm{src}} A_{\mathrm{src} \to f^*},
\label{eq:invariant}
\end{equation}
where the sum runs over all feature, error, and input source vertices. This identity holds before any thresholding, aggregation, expansion or pruning. Position aggregation (§\ref{app:position}) preserves it exactly; compaction during iterative construction (§\ref{app:iterative}) preserves it except for the position buckets whose summed attribution falls below $\tau$, which it drops. Pruning, by contrast, deliberately drops low-influence sources and therefore does \emph{not} preserve (\ref{eq:invariant}); the magnitude of the resulting violation is itself a useful quality metric (§\ref{app:validation}).

We compute (\ref{eq:invariant}) at the raw stage (on the per-position graph after expansion and compaction) and at the pruned stage (after aggregation and pruning) as a numerical sanity check; the aggregated stage is omitted because aggregation preserves the invariant up to floating-point rounding. The raw measurement itself is not exactly zero: edge extraction applies a magnitude threshold $\tau$ (Algorithm~\ref{alg:edge_extraction}) that drops a long tail of small per-position contributions, compaction drops the truncation buckets whose total is below the same $\tau$ (Algorithm~\ref{alg:compaction}), and the base model runs in bfloat16. Empirical values, and the size of the bfloat16 effect, are reported in §\ref{app:validation:invariant}.

\subsection{Cross-stream edges}
\label{app:attribution:cross}

The frozen joint attention couples the streams. In $\mathrm{FrozenAttn}$, $V$ is the concatenation $V_{\mathrm{txt}} \mathbin\Vert V_{\mathrm{img}}$ along the token axis (§\ref{app:lrm:cache}), and the cached probability tensor $P^\ell$ mixes these into per-stream outputs:
\begin{equation}
\mathrm{FrozenAttn}_\ell(x_{\mathrm{img}}, x_{\mathrm{txt}})_s = W_O^{(\ell, s)}\!\bigl(\bigl(P^\ell\, [V_{\mathrm{txt}}(x_{\mathrm{txt}}) \mathbin\Vert V_{\mathrm{img}}(x_{\mathrm{img}})]\bigr)_s\bigr) + \varepsilon^{\ell, s}_{\mathrm{attn}}.
\end{equation}
Since $V_{\mathrm{txt}}$ is an affine function of $x_{\mathrm{txt}}$ and $V_{\mathrm{img}}$ an affine function of $x_{\mathrm{img}}$ (§\ref{app:lrm:substitution}), the gradient of $h^*$ with respect to the residual stream of stream $s$ at position $p$ has nonzero components both in the same-stream residual (via $V_s$) and, after a frozen-attention step, in the other-stream residual at \emph{every} position. Concretely, when the backward pass of $h^*$ traverses an attention block at layer $\ell$, the gradient on the post-attention residual flows back through $W_O^{(\ell, s)} P^\ell$ into both $V_{\mathrm{txt}}$ and $V_{\mathrm{img}}$, and from there into the pre-attention residuals of both streams.

The practical consequence is that the attribution graph naturally contains $\mathrm{txt} \to \mathrm{img}$ and $\mathrm{img} \to \mathrm{txt}$ feature edges. A text-stream feature at $(\ell, \mathrm{txt}, p, i)$ writes its decoder vector into the txt residual at position $p$; the gradient $g^{\ell + 1, \mathrm{txt}}(p)$ used in (\ref{eq:feat_edge}) carries contributions that originated, after one or more frozen-attention steps, in the image-stream residual feeding the target encoder. The corresponding edge weight is the inner product of that gradient with the source's decoder vector, and is computed by exactly the same formula as a same-stream edge: no special case is required.

This cross-stream propagation is the single most important property the LRM inherits from MM-DiT: it is what allows the attribution graph to expose, edge by edge, how textual features get instantiated into spatial regions of the image and conversely how visual features influence text-side computation. We exploit this property extensively in §\ref{sec:experiments}.

\subsection{What the graph does not model}
\label{app:attribution:notmodeled}

Several pieces of the original FLUX.1[schnell] computation are not represented in the attribution graph:

\begin{itemize}[leftmargin=*]
\item \emph{Attention $Q$-$K$ pathway.} Attention probabilities $P^\ell$ are cached and treated as constants. The graph thus explains where information flows through attention (via the OV pathway), but not why the model attends where it does. Decomposing $P^\ell$ itself into feature-level causes is a separate, harder problem and is left to future work.

\item \emph{Input embedding computation.} The input vertices carry the full residual stream entering block $0$ for each stream, but the production of these vectors -- the T5 encoding of the prompt and its linear projection for $s = \mathrm{txt}$, and the noisy latent of the current denoising step with its patch projection for $s = \mathrm{img}$ -- is upstream of the LRM and is not decomposed.

\item \emph{Downstream blocks.} The double-stream blocks $16$--$18$ and the single-stream blocks $19$--$56$ of FLUX.1[schnell], in which the streams are processed jointly with shared weights, lie downstream of every analyzed block and run unsubstituted in the LRM. An MLP feature whose effects manifest only after passing through these blocks will not have its downstream consequences represented in the graph.
\end{itemize}

These restrictions match those of prior circuit-tracing work in LLMs and are accepted for the same tractability reasons.

\section{Position aggregation}
\label{app:position}

The attribution graph constructed in §\ref{app:attribution} is per-position: each source feature, MLP error, truncation error (Appendix~\ref{app:iterative:compaction}) and input appears once for every position at which it contributes. For an image-stream target this typically means $\mathcal{O}(10^4)$ vertices in the raw graph (Table~\ref{tab:graph_size}): at most $N_{\max}$ expanded feature vertices, and MLP-error, truncation-error and input vertices at every position where their attribution reaches $\tau$. Such graphs are unwieldy for interpretation, and the typical question of interest is \emph{which} feature participates in a circuit, not at which position.

\paragraph{Aggregation rule.} We collapse all per-position vertices that share the same $(\ell, s, i)$ into a single \emph{aggregated feature vertex}, with edge weight equal to the algebraic sum of per-position attributions:
\begin{equation}
\bar{A}_{(\ell, s, i) \to f^*} \;=\; \sum_p A_{(\ell, s, p, i) \to f^*}.
\end{equation}
Error vertices are aggregated analogously, separately for MLP reconstruction errors and truncation errors (§\ref{app:iterative:compaction}): each is collapsed to one vertex per $(\ell, s)$ pair, with edge weight $\sum_p A_{(\ell, s, p)_{\mathrm{err}} \to f^*}$. Input vertices are aggregated per stream: $\bar{A}_{s_{\mathrm{in}} \to f^*} = \sum_p A_{(s, p)_{\mathrm{in}} \to f^*}$. Note that the target itself remains a single vertex; only sources are aggregated.

\paragraph{Activation maps.} We retain the per-position activation pattern of each aggregated feature as a sparse map
\begin{equation}
m_{(\ell, s, i)}: p \mapsto z^{(\ell, s, i)}(p),
\end{equation}
stored alongside the aggregated graph. These maps are the natural visualization of where in the image (or in the prompt) a feature fires; they are not used during pruning or analysis but are essential for human inspection.

\paragraph{Properties.} Aggregation strictly preserves the conservation invariant (\ref{eq:invariant}): it just regroups terms in the right-hand side. Aggregation can in principle hide structure when per-position attributions cancel, but on the targets analyzed in §\ref{sec:experiments} this does not appear to be the limiting factor. Qualitative inspection of the activation maps that are stored alongside aggregated vertices allows for easy interpretation of aggregated feature nodes. Aggregation reduces vertex count by approximately $12\times$ on image targets and $6\times$ on text targets in our experiments (averages, §\ref{app:validation:size}).

\begin{figure}[H]
\centering
\includegraphics[width=\linewidth]{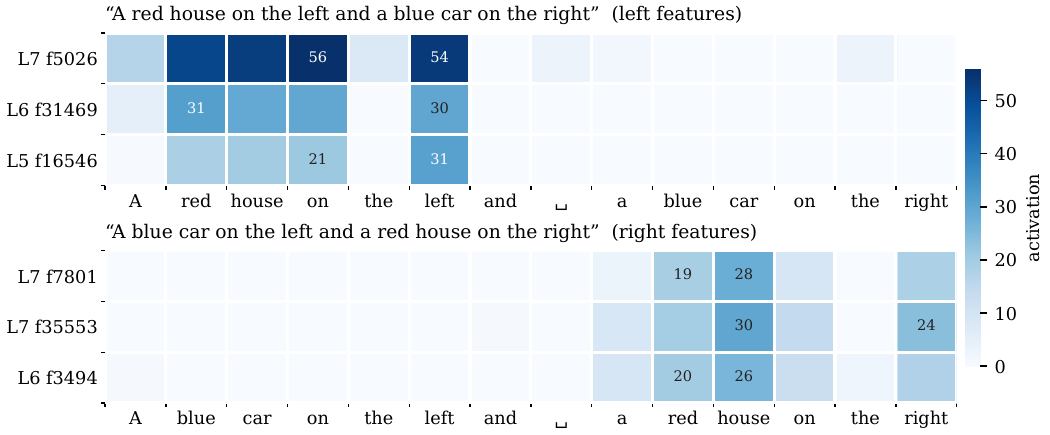}
\caption{Per-token activation maps kept with aggregated vertices. Two published graphs of $f^{(\mathrm{img},9)}_{\mathrm{red}}$ (step $0$, the same seed) on two prompts that swap the objects; rows are the location features present in each pruned graph, columns the first $14$ prompt tokens, and the two strongest activations per row are printed. The `left' features fire on \textit{red house on the left}, the `right' features on \textit{red house on the right}.}
\label{fig:position_maps}
\end{figure}

\paragraph{What the maps keep.} Summing attributions over positions can hide structure: a feature that helps the target at one position and hurts it at another gets a small aggregated edge, and the aggregated vertex no longer says where the feature acts. The activation maps kept with every aggregated vertex restore the second. Figure~\ref{fig:position_maps} shows them for the spatial case study (Appendix~\ref{sec:exp:spatial}). The pruned graph of $f^{(\mathrm{img},9)}_{\mathrm{red}}$ contains the three location features of the `left' supernode on \textit{A red house on the left and a blue car on the right} and the three of the `right' supernode on the swapped prompt, and none of the other set; the maps show that each set fires on the phrase of the red object together with its location word, which is why the location that enters the graph follows the red object.

\paragraph{Aggregation is post-hoc.} Iterative graph construction (§\ref{app:iterative}) operates on the per-position graph, so the budgeted expansion explores the full per-position structure before aggregation collapses it. Aggregating before expansion would change which sources are picked up, since a source whose per-position attributions happen to cancel would never enter the discovered set in the first place; doing it after expansion preserves coverage. The same applies to compaction: truncation-error vertices are introduced at full per-position resolution and only then aggregated.

\section{Iterative graph construction}
\label{app:iterative}

A naive construction would compute one VJP per feature vertex of interest, which is infeasible at our graph sizes: each source feature has its own incoming edges, those sources have their own incoming edges, and so on down to the first layer, so every per-position feature reached along the way needs its own VJP. The full graph for a typical layer-$15$ target on a $1024$-token image stream would require on the order of $10^4$ VJPs even before recursive expansion of those features' own sources.

We therefore use a budgeted greedy expansion algorithm: starting from the target, we iteratively expand the most influential frontier features and stop when a fixed number of VJPs has been spent. Unexpanded but discovered features are folded into truncation-error vertices to preserve the conservation invariant.

\subsection{Indirect-influence scoring}
\label{app:iterative:scoring}

Let $\mathcal{D}$ be the discovered set (vertices that appear as the source of at least one extracted edge) and $\mathcal{E} \subseteq \mathcal{D}$ the expanded set (vertices whose incoming edges have been computed via a VJP). At any point during expansion, we have a partial directed graph on $\mathcal{D}$ in which only the in-edges of $\mathcal{E}$ are filled in. We need a way to score the unexpanded discovered features by how much their eventual influence on the target is likely to be.

\paragraph{Reach over the expanded subgraph.} Define the column-normalized adjacency over expanded vertices:
\begin{equation}
A^{\mathrm{norm}}_{ij} \;=\; \frac{|A_{i \to j}|}{\sum_{i'} |A_{i' \to j}| + \varepsilon}, \qquad i, j \in \mathcal{E},
\end{equation}
which gives a nonnegative matrix over $\mathcal{E}$ whose columns sum to $\le 1$. The indirect-influence matrix is
\begin{equation}
B \;=\; (I - A^{\mathrm{norm}})^{-1} - I,
\end{equation}
whose entry $B_{u, f^*}$ sums the strengths of all paths from $u$ to $f^*$ through $\mathcal{E}$, where path strength is the product of per-edge column-normalized weights. We define the reach of $u$ from the target as
\begin{equation}
\mathrm{reach}(u, f^*) \;=\; \mathbf{1}[u = f^*] + B_{u, f^*}.
\end{equation}

\paragraph{Score for unexpanded features.} For a discovered but unexpanded feature $v \in \mathcal{D} \setminus \mathcal{E}$, its score is the sum of its outgoing edges into expanded vertices, weighted by the reach of those vertices to the target:
\begin{equation}
\sigma(v) \;=\; \sum_{u \in \mathcal{E},\ v \to u} |A_{v \to u}| \cdot \mathrm{reach}(u, f^*).
\end{equation}
The scoring is cheap: $A^{\mathrm{norm}}$ has size $|\mathcal{E}|^2$, the matrix inverse is computed once per scoring round, and the per-vertex update is a sparse dot product.

\subsection{Algorithm}
\label{app:iterative:algorithm}

\begin{algorithm}[H]
\caption{Budgeted iterative graph construction.}
\label{alg:iterative}
\begin{algorithmic}[1]
\REQUIRE Cached LRM forward state; target $f^*$; threshold $\tau$; batch size $k$; budget $N_{\max}$.
\ENSURE A directed graph $\mathcal{G}$ rooted at $f^*$ with $|\mathcal{E}| \le N_{\max}$ expanded vertices.

\STATE Initialize $\mathcal{E} \gets \{f^*\}$, $\mathcal{D} \gets \{f^*\}$, $\mathcal{G} \gets \emptyset$.
\STATE Run Algorithm~\ref{alg:edge_extraction} from $f^*$ with threshold $\tau$ to extract its incoming edges $E_0$.
\STATE $\mathcal{G} \gets \mathcal{G} \cup E_0$;\quad $\mathcal{D} \gets \mathcal{D} \cup \{\mathrm{src}(e) : e \in E_0\}$.

\WHILE{$|\mathcal{E}| < N_{\max}$}
    \STATE Compute $A^{\mathrm{norm}}, B$ over $\mathcal{E}$ and $\mathrm{reach}(u, f^*)$ for all $u \in \mathcal{E}$.
    \STATE Compute $\sigma(v)$ for all $v \in (\mathcal{D} \setminus \mathcal{E})$ that are feature vertices with $\ell(v) < \ell^*$.
    \STATE Let $V_{\mathrm{batch}}$ be the top $\min(k, N_{\max} - |\mathcal{E}|)$ such vertices by $\sigma$.
    \IF{$V_{\mathrm{batch}} = \emptyset$}
        \STATE \textbf{break} \hfill (no unexpanded feature vertex left)
    \ENDIF
    \FOR{each $v \in V_{\mathrm{batch}}$}
        \STATE Run Algorithm~\ref{alg:edge_extraction} from $v$ with threshold $\tau$ to extract its incoming edges $E_v$.
        \STATE $\mathcal{G} \gets \mathcal{G} \cup E_v$;\quad $\mathcal{D} \gets \mathcal{D} \cup \{\mathrm{src}(e) : e \in E_v\}$;\quad $\mathcal{E} \gets \mathcal{E} \cup \{v\}$.
    \ENDFOR
\ENDWHILE

\STATE Apply compaction with threshold $\tau$: for each unexpanded feature vertex $v \in \mathcal{D} \setminus \mathcal{E}$, redistribute its outgoing edges into expanded vertices into truncation-error vertices (Algorithm~\ref{alg:compaction}).

\RETURN $\mathcal{G}$ on vertices $\mathcal{E} \cup \{\text{MLP-error, truncation-error and input vertices}\}$.
\end{algorithmic}
\end{algorithm}

In the implementation, error and input vertices are never expanded (they have no incoming edges by construction) but are passed through to compaction and pruning; only feature vertices with $\ell < \ell^*$ are eligible for expansion. We use $\tau = 10^{-3}$ for the minimum-attribution threshold of edge extraction and compaction, $k = 50$ for the per-iteration expansion batch size, and $N_{\max} = 1000$ for the total VJP budget; with these values one graph, its perturbation validation included, took $30$--$60$ (median $40$) minutes on one NVIDIA H100. The choice of $N_{\max}$ trades graph size for quality: enlarging the budget retains more sources at the expense of larger graphs, but the marginal returns saturate quickly. To check this, we constructed graphs at $N_{\max} \in \{500, 1500\}$ on a fixed set of $30$ targets (feature, prompt, denoising step) and measured the Spearman correlation against pairwise ablation in the original model (§\ref{app:validation:perturbation}) and the resulting graph size (Table~\ref{tab:nmax_ablation}). Tripling the budget improves Spearman from $0.658$ to $0.691$, while doubling the number of pruned graph vertices from $317$ to $652$. The quality gain is modest relative to the size cost; we therefore set $N_{\max} = 1000$, between the two budgets.

\begin{table}[H]
\centering
\caption{Effect of the VJP budget $N_{\max}$ on graph quality and size, averaged over $30$ targets. All other hyperparameters fixed at their defaults.}
\label{tab:nmax_ablation}
\begin{tabular}{lcc}
\toprule
$N_{\max}$ & Spearman $\rho$ $\uparrow$ & Pruned vertices \\
\midrule
$500$  & 0.658 & 317 \\
$1500$ & 0.691 & 652 \\
\bottomrule
\end{tabular}
\end{table}

\subsection{Compaction}
\label{app:iterative:compaction}

When expansion terminates, $\mathcal{D} \setminus \mathcal{E}$ contains discovered but unexpanded feature vertices: their outgoing edges into expanded vertices have been computed and recorded in $\mathcal{G}$, but their own incoming edges are unknown. If we left these vertices in the graph as-is, the conservation invariant would still hold -- the vertices have no incoming edges but their outgoing contributions are already accounted for -- but every interpretation tool downstream would need to handle feature vertices whose contribution we know but whose computation we don't. We instead replace them with truncation-error vertices that aggregate per source position, turning the truncation into an explicit, auditable component of the graph.

\begin{algorithm}[H]
\caption{Compaction of unexpanded features.}
\label{alg:compaction}
\begin{algorithmic}[1]
\REQUIRE Graph $\mathcal{G}$; expanded set $\mathcal{E}$; threshold $\tau$.
\ENSURE Compacted graph $\mathcal{G}'$; the invariant holds except for the dropped buckets with $|a| < \tau$.

\STATE Initialize $\mathcal{G}' \gets \emptyset$.
\STATE Bucket $\mathcal{T} \gets \emptyset$ \hfill (truncation-error attributions, keyed by source position)
\FOR{each edge $(u \to v) \in \mathcal{G}$}
    \IF{$u \in \mathcal{E}$ \textbf{and} $v \in \mathcal{E}$}
        \STATE Add $(u \to v)$ to $\mathcal{G}'$.
    \ELSIF{$u$ is an error or input vertex \textbf{and} $v \in \mathcal{E}$}
        \STATE Add $(u \to v)$ to $\mathcal{G}'$.
    \ELSIF{$u$ is an unexpanded feature \textbf{and} $v \in \mathcal{E}$}
        \STATE Bucket: $\mathcal{T}[(\ell(u), s(u), p(u), v)] \mathrel{+}= A_{u \to v}$.
    \ENDIF
\ENDFOR

\FOR{each $((\ell, s, p, v), a) \in \mathcal{T}$ with $|a| \ge \tau$}
    \STATE Add a truncation-error vertex $\mathrm{trunc}_{\ell, s, p}$ if not present.
    \STATE Add edge $(\mathrm{trunc}_{\ell, s, p} \to v)$ to $\mathcal{G}'$ with attribution $a$.
\ENDFOR

\RETURN $\mathcal{G}'$.
\end{algorithmic}
\end{algorithm}

Truncation-error vertices are distinct from MLP reconstruction error vertices (§\ref{app:attribution}) in their semantics: an MLP error vertex carries the residual variance the transcoder failed to capture on that block, while a truncation-error vertex carries the contribution of source features that were too low-priority to expand. Both behave the same way during pruning (exempt) and validation (counted toward $\sum A$), and downstream tooling treats them under the unified ``error'' type, but reporting them separately during analysis is informative: a target whose attribution is dominated by truncation errors is one for which the budget was too tight, whereas a target dominated by MLP errors signals that the transcoders themselves are leaving variance on the table at that block.

\section{Pruning}
\label{app:pruning}

The graph produced by Algorithms~\ref{alg:iterative} and~\ref{alg:compaction} after position aggregation typically contains a thousand vertices and on the order of $10^5$ edges, of which only a small fraction carry significant influence on the target. Pruning reduces the graph to an interpretable size by removing the long tail, in two passes over the position-aggregated graph: first vertices, then edges.

\subsection{Indirect-influence preliminaries}
\label{app:pruning:influence}

Let $\mathcal{V}$ be the vertex set of the aggregated graph and $\mathcal{V}_{\mathrm{feat}} \subset \mathcal{V}$ its feature vertices. Define the column-normalized absolute adjacency on $\mathcal{V}$ exactly as in §\ref{app:iterative:scoring},
\begin{equation}
A^{\mathrm{norm}}_{ij} \;=\; \frac{|A_{i \to j}|}{\sum_{i'} |A_{i' \to j}| + \varepsilon},
\end{equation}
and the indirect-influence matrix
\begin{equation}
B = (I - A^{\mathrm{norm}})^{-1} - I.
\end{equation}

The influence of a vertex $v$ on the target is
\begin{equation}
\mathrm{infl}(v) \;=\; B_{v, f^*},
\end{equation}
which sums all path strengths from $v$ to $f^*$ in the aggregated graph. For the target itself we set $\mathrm{infl}(f^*) = 1$, its reach in Appendix~\ref{app:iterative:scoring}; this enters only the scores of the edges into the target.

\subsection{Vertex pruning}
\label{app:pruning:vertices}

We rank the feature vertices of a single stream together with that stream's input vertex, $\mathcal{V}^s = \mathcal{V}^s_{\mathrm{feat}} \cup \{\mathrm{in}_s\}$, by $\mathrm{infl}(v)$ and retain the smallest cumulative-influence prefix that covers $\theta^s_v = 80\%$ of the total influence:
\begin{equation}
\begin{aligned}
\mathcal{V}^{s, \mathrm{kept}} \;=\; \text{top-}K\bigl(\{(v, \mathrm{infl}(v))\}_{v \in \mathcal{V}^s}\bigr) \\
\text{with } K \text{ chosen so that } \tfrac{\sum_{v \in \mathcal{V}^{s, \mathrm{kept}}} \mathrm{infl}(v)}{\sum_{v \in \mathcal{V}^s} \mathrm{infl}(v)} \ge 0.8.
\end{aligned}
\end{equation}
The input vertex carries the total attribution of the boundary residual stream and typically ranks at or near the top of its stream; it was retained in every graph we report.

\paragraph{Per-stream pruning.} We apply this rule independently to image-stream and text-stream vertices, with separate $80\%$ thresholds. The layer-wise distribution of vertices is highly asymmetric: in our experiments, image-stream features dominate at deep blocks while text-stream features are concentrated in early blocks. A single $80\%$ threshold applied to the union of both streams ranks all vertices by global influence and cuts the long tail without regard to which stream they belong to, which can drop entire stream--layer regions that genuinely participate in the circuit but happen to fall below the global threshold. Per-stream pruning preserves a balanced view of both modalities at every depth.

\paragraph{Exempt vertices.} Error vertices (both MLP reconstruction and truncation) are exempt from pruning. They account for the variance not explained by the kept features, and silently dropping them would make the conservation invariant violation indistinguishable from genuine missing structure. Exempt vertices are always retained regardless of their influence score.

\subsection{Edge pruning}
\label{app:pruning:edges}

After vertex pruning we re-form the adjacency on the surviving vertices and assign each edge a contribution score
\begin{equation}
\mathrm{score}(u \to v) \;=\; A^{\mathrm{norm}}_{u \to v} \cdot \mathrm{infl}(v),
\end{equation}
which combines how much $u$ contributes to $v$'s preactivation with how much $v$ contributes to the target. Edges are ranked by score and the smallest cumulative-score prefix covering $98\%$ of total edge score is retained. As with vertex pruning, the threshold is applied separately to edges sourced from image-stream and text-stream vertices, for the same reason. Edges incident to error or input vertices participate in this ranking like any other.

\subsection{Algorithm}
\label{app:pruning:alg}

\begin{algorithm}[H]
\caption{Two-step pruning.}
\label{alg:pruning}
\begin{algorithmic}[1]
\REQUIRE Aggregated graph $\mathcal{G}$; vertex thresholds $\theta^{\mathrm{img}}_v, \theta^{\mathrm{txt}}_v$; edge thresholds $\theta^{\mathrm{img}}_e, \theta^{\mathrm{txt}}_e$.
\ENSURE Pruned graph $\mathcal{G}_{\mathrm{pruned}}$.

\STATE Compute $A^{\mathrm{norm}}, B$ on $\mathcal{V}(\mathcal{G})$; let $\mathrm{infl}(v) = B_{v, f^*}$ and $\mathrm{infl}(f^*) = 1$.
\STATE Split feature and input vertices by stream: $\mathcal{V}^{\mathrm{img}} = \mathcal{V}^{\mathrm{img}}_{\mathrm{feat}} \cup \{\mathrm{in}_{\mathrm{img}}\}$, $\mathcal{V}^{\mathrm{txt}} = \mathcal{V}^{\mathrm{txt}}_{\mathrm{feat}} \cup \{\mathrm{in}_{\mathrm{txt}}\}$.
\STATE For each stream $s$, retain the smallest top-influence prefix of $\mathcal{V}^s$ covering $\theta^s_v$ of stream $s$ influence.
\STATE Retain all error vertices (MLP reconstruction and truncation).
\STATE Form vertex-pruned graph $\mathcal{G}_v$.

\STATE Recompute $A^{\mathrm{norm}}, B, \mathrm{infl}$ on $\mathcal{G}_v$.
\STATE Score each edge by $A^{\mathrm{norm}}_{u \to v} \cdot \mathrm{infl}(v)$.
\STATE Split edges by source stream; for each stream $s$, retain the smallest top-score prefix covering $\theta^s_e$ of stream $s$ edge score.
\STATE Prune all other edges.

\RETURN $\mathcal{G}_{\mathrm{pruned}}$.
\end{algorithmic}
\end{algorithm}

We use $\theta^{\mathrm{img}}_v = \theta^{\mathrm{txt}}_v = 0.8$ and $\theta^{\mathrm{img}}_e = \theta^{\mathrm{txt}}_e = 0.98$ throughout. With these defaults, pruning reduces the number of vertices in the aggregated graph by approximately $2.4\times$ and the number of edges by approximately $10\times$, while increasing the mean conservation-invariant relative error by about $4$ percentage points (§\ref{app:validation:invariant}).

\paragraph{Loss of conservation.} Unlike position aggregation, and unlike compaction, which drops only buckets below $\tau$, pruning does \emph{not} preserve the conservation invariant: the dropped vertices and edges had nonzero attributions, and their removal changes $\sum A$. We track this loss explicitly as the pruned attribution relative error in §\ref{app:validation:invariant}.

\section{Empirical validation of attribution graphs}
\label{app:validation}

We validate the full pipeline of §§\ref{app:lrm}--\ref{app:pruning} on the subset of attribution graphs used in the experiments of §\ref{sec:experiments}: $86$ targets in total, of which $51$ have an image-stream target feature and $35$ have a text-stream target feature, drawn from a variety of prompts and denoising steps. For each target we run iterative graph construction with the parameters of §\ref{app:iterative:algorithm} ($\tau = 10^{-3}$, $k = 50$, $N_{\max} = 1000$), aggregate positions (§\ref{app:position}), apply two-step pruning with the defaults of §\ref{app:pruning:alg} ($80\%$ vertices, $98\%$ edges, per stream), and record three families of metrics: graph statistics (size after each step), conservation invariant residuals (raw and pruned), and pairwise mechanistic faithfulness against the original FLUX.1[schnell].

\subsection{Graph statistics}
\label{app:validation:size}

Table~\ref{tab:graph_size} reports the mean, median, minimum, and maximum number of vertices and edges at each stage of the pipeline, separately for image-stream and text-stream targets.

\begin{table}[H]
\centering
\caption{Graph size statistics across all evaluated targets, broken down by target stream and pipeline stage. Statistics are taken over targets within each group ($n=51$ for image targets, $n=35$ for text targets).}
\label{tab:graph_size}
\resizebox{\textwidth}{!}{%
\begin{tabular}{lcccccccc}
\toprule
& \multicolumn{4}{c}{Image targets ($n=51$)} & \multicolumn{4}{c}{Text targets ($n=35$)} \\
\cmidrule(lr){2-5} \cmidrule(lr){6-9}
Stage & Mean & Median & Min & Max & Mean & Median & Min & Max \\
\midrule
Vertices, raw         & $10{,}640$ & $10{,}356$ & $7{,}142$ & $14{,}601$ & $5{,}744$ & $4{,}092$ & $2{,}897$ & $14{,}599$ \\
Vertices, aggregated  & $888$ & $900$ & $704$ & $1{,}021$ & $1{,}007$ & $1{,}010$ & $916$ & $1{,}032$ \\
Vertices, pruned      & $337$ & $310$ & $233$ & $522$ & $473$ & $493$ & $304$ & $580$ \\
\midrule
Edges, raw            & $1{,}814{,}092$ & $1{,}442{,}714$ & $1{,}290{,}505$ & $4{,}311{,}539$ & $1{,}466{,}397$ & $1{,}302{,}581$ & $975{,}376$ & $2{,}896{,}847$ \\
Edges, aggregated     & $274{,}041$ & $272{,}805$ & $160{,}231$ & $396{,}517$ & $414{,}681$ & $421{,}750$ & $271{,}601$ & $447{,}408$ \\
Edges, pruned         & $20{,}713$ & $16{,}297$ & $8{,}436$ & $65{,}259$ & $55{,}381$ & $61{,}240$ & $15{,}875$ & $89{,}635$ \\
\bottomrule
\end{tabular}%
}
\end{table}

The reduction from raw to aggregated graphs is dominated by the position collapse of the error and input vertices: the per-position MLP-error, truncation-error and input vertices become one vertex per $(\ell, s)$ pair or per stream, while expanded features merge only where the same feature was expanded at several positions. The raw-to-aggregated reduction factor in vertex count is approximately $12\times$ for image targets and $4\times$ for text targets (medians). Edge counts reduce correspondingly by $5\times$ and $3\times$.

Pruning further reduces aggregated graphs to a small interpretable size, retaining a median of $310$ pruned vertices for image targets and $493$ for text targets. Notably, text-stream targets have larger pruned graphs than image-stream targets despite starting from smaller raw graphs. This reflects how the per-stream $80\%$ vertex threshold interacts with each stream's influence distribution: image-stream feature influence is more concentrated in a small subset of heavy hitters, so the $80\%$ cumulative-influence threshold is reached after retaining a smaller fraction of feature vertices, while text-stream feature influence is distributed more evenly, so reaching $80\%$ requires retaining a larger fraction.

\subsection{Conservation invariant}
\label{app:validation:invariant}

For each target we compute the relative error of the conservation invariant (\ref{eq:invariant}) at two stages of the pipeline. The raw relative error is computed on the per-position graph after expansion and compaction (Algorithms~\ref{alg:iterative} and~\ref{alg:compaction}), before aggregation and pruning; the pruned relative error is computed on the final graph after pruning. Position aggregation preserves the invariant exactly, so an aggregated-stage measurement coincides with the raw measurement and is omitted.

We define the relative error as
\begin{equation}
\delta \;=\; \frac{\bigl|\sum_{\mathrm{src}} A_{\mathrm{src} \to f^*} \;-\; (h^* - b^*_{\mathrm{eff}})\bigr|}{|h^* - b^*_{\mathrm{eff}}|}.
\end{equation}
Aggregate values are reported in Table~\ref{tab:invariant} for image and text targets separately.

\begin{table}[H]
\centering
\caption{Conservation invariant relative error $\delta$ (in percent), at the raw and pruned stages, broken down by target stream. Statistics are taken over $86$ targets in total ($51$ image, $35$ text).}
\label{tab:invariant}
\begin{tabular}{lcccccccc}
\toprule
& \multicolumn{4}{c}{Image targets ($n=51$)} & \multicolumn{4}{c}{Text targets ($n=35$)} \\
\cmidrule(lr){2-5} \cmidrule(lr){6-9}
Stage & Mean $\downarrow$ & Median $\downarrow$ & Min & Max & Mean $\downarrow$ & Median $\downarrow$ & Min & Max \\
\midrule
Raw $\delta$ (\%)    & 7.86 & 6.49 & 0.43 & 28.79 & 5.14 & 4.17 & $<0.1$ & 29.74 \\
Pruned $\delta$ (\%) & 12.38 & 10.87 & 1.15 & 30.55 & 9.13 & 8.29 & $<0.1$ & 29.10 \\
\bottomrule
\end{tabular}
\end{table}

\paragraph{Sources of raw error.} Under exact arithmetic and with $\tau = 0$, the raw $\delta$ would be zero by the derivation of §\ref{app:attribution}. The nonzero values in Table~\ref{tab:invariant} occur because of threshold truncation. Algorithm~\ref{alg:edge_extraction} retains only edges with $|A| \ge \tau = 10^{-3}$ from sources whose activation exceeds $\tau$ and drops a long tail of low-magnitude per-position contributions; compaction (Algorithm~\ref{alg:compaction}) also drops the truncation buckets whose total is below the same $\tau$. The bfloat16 arithmetic of the base model inside the LRM (Table~\ref{tab:hyperparams}) adds little: re-tracing four targets with the base model in float32 changes raw $\delta$ by at most $0.35$ percentage points. The wide range across targets (raw $\delta$ from $0.4\%$ to $28.8\%$ on image targets) reflects target-dependent variation in the denominator: targets with smaller $|h^* - b^*_{\mathrm{eff}}|$ produce larger relative errors for the same absolute mass dropped.

\paragraph{Stream comparison.} Image targets show a higher raw $\delta$ ($6.49\%$ median) than text targets ($4.17\%$ median). The gap is driven mainly by the conservation residual itself: image targets drop a $\sim 2.5\times$ larger absolute attribution mass than text targets ($\bigl|(h^* - b^*_{\mathrm{eff}}) - \sum A\bigr|$ medians of $3.10$ vs $1.25$), partially offset by image targets' $\sim 1.2\times$ larger denominator ($50.5$ vs $41.7$). The larger truncation mass in image graphs is consistent with each aggregated image edge unfolding into roughly $5$ per-position contributions versus $3$ for text targets, so under a fixed threshold $\tau = 10^{-3}$ image graphs accumulate truncation across more per-position contributions per aggregated edge.

\paragraph{Pruning loss.} The pruned $\delta$ is generally larger than the raw $\delta$, since pruning drops edges that contributed to the source-side sum. On the targets we evaluated, mean $\delta$ increases by approximately $4.5$ percentage points for image targets ($7.86\% \to 12.38\%$) and $4$ percentage points for text targets ($5.14\% \to 9.13\%$). Pruning typically increases $\delta$ but on some targets decreases it; both directions are explained by the relative-error metric being the absolute difference $|\sum A - (h^* - b^*_{\mathrm{eff}})|$. Pruning typically widens this gap by removing edges that contributed to $\sum A$, but it can also narrow the gap when the pruned edges happen to share sign with the residual already present from threshold truncation. Such reductions in $\delta$ are an artifact of the metric's symmetry around zero and not a sign of better explanatory coverage. Overall the pruning penalty is small relative to the order-of-magnitude graph-size reduction it provides (Table~\ref{tab:graph_size}).

\subsection{What the pruned graph accounts for}
\label{app:validation:shares}

The conservation invariant also says how the explained quantity $h^* - b^*_{\mathrm{eff}}$ is divided between the parts of a pruned graph. For every target we take the edges that enter it and express each source type as a share of $h^* - b^*_{\mathrm{eff}}$; the two remaining terms are the mass that the thresholds of edge extraction and compaction left out and the mass that pruning removed. The shares are signed and sum to $100\%$ per graph.

\begin{table}[H]
\centering
\caption{Shares of $h^* - b^*_{\mathrm{eff}}$ in the pruned graphs (percent, mean / median / pooled over targets), for the $86$ validation targets.}
\label{tab:shares}
\resizebox{\textwidth}{!}{%
\begin{tabular}{lccccc}
\toprule
Targets & Feature vertices & Error vertices & Input vertices & Threshold residual & Pruning loss \\
\midrule
All ($86$)   & 61.6 / 63.7 / 60.9 & 19.3 / 18.9 / 21.2 & 9.5 / 7.0 / 8.5 & 4.7 / 3.4 / 4.2 & 5.0 / 5.0 / 5.2 \\
Text ($35$)  & 84.1 / 84.2 / 83.3 & 3.5 / 3.5 / 4.8 & 5.9 / 3.8 / 4.6 & 0.7 / 0.9 / 0.5 & 5.8 / 5.4 / 6.8 \\
Image ($51$) & 46.1 / 44.6 / 48.6 & 30.1 / 26.8 / 30.3 & 11.9 / 10.3 / 10.7 & 7.5 / 6.5 / 6.2 & 4.4 / 4.7 / 4.3 \\
\bottomrule
\end{tabular}%
}
\end{table}

Feature vertices carry about three fifths of $h^* - b^*_{\mathrm{eff}}$ overall, and the honest picture is stream-dependent. Text-stream targets are nearly fully explained, at $84\%$. For image-stream targets about $30\%$ of the quantity sits in error vertices and about $7.5\%$ in the threshold residual, so \emph{an image-target graph should be read as accounting for roughly half of $h^* - b^*_{\mathrm{eff}}$, not all of it}.

\subsection{Mechanistic faithfulness via perturbation}
\label{app:validation:perturbation}

The conservation invariant verifies that the attribution graph is internally consistent on the LRM, but a graph that is internally consistent might still mis-predict what happens in the original model. To check this we perform a pairwise faithfulness evaluation. The $86$ targets are those of the case studies and are not an independent sample; what is not selected is the set of source--target pairs inside each graph.

\paragraph{Procedure.} Fix the per-position attribution graph of a target after compaction (Algorithm~\ref{alg:compaction}), before aggregation and pruning, and let $\mathcal{V}_{\mathrm{feat}}$ be its feature vertices $(\ell, s, p, i)$. We rank these by their total indirect influence on the other feature vertices, $\sum_{t \neq v} |B_{v, t}|$, and take the top $K = 30$ as the source set $\mathcal{S}$. For each source vertex $v \in \mathcal{S}$ we ablate the corresponding feature in the \emph{original} FLUX.1[schnell], not in the LRM, at the vertex's own position $p(v)$: a forward hook on the source's MLP block subtracts $z^{(\ell(v), s(v), i(v))}(p(v)) \cdot f^{(\ell(v), s(v), i(v))}_{\mathrm{dec}}$ from the MLP output at position $p(v)$, leaving all other positions untouched. We then measure the resulting change in $h_t$ for every other vertex $t \in \mathcal{S}$ by running the unmodified original model with this hook applied and re-extracting $h_t$ on the same prompt.

This gives, for each ordered pair $(v, t) \in \mathcal{S} \times \mathcal{S}$ with $v \neq t$ ($K(K-1) = 870$ pairs per graph), an actual ablation effect $|\Delta h_t|_{\mathrm{actual}} = |h_t^{\mathrm{ablated}} - h_t^{\mathrm{baseline}}|$. We compare it to the predicted effect from the graph: the absolute indirect-influence matrix entry $|B_{v, t}|$, which sums all paths from $v$ to $t$ in the column-normalized graph and is a dimensionless structural measure of how much $v$ should influence $t$; the actual effect $|\Delta h_t|$ is in the units of preactivations. We therefore evaluate the predicted-actual relationship through rank and linear correlations rather than absolute agreement. Stacking over these pairs and discarding those in which both quantities are below $10^{-12}$, we report the Spearman and Pearson correlations between predicted and actual effects.

\paragraph{Why ablate in the original model and not in the LRM.} A perturbation experiment in the LRM is by definition consistent with the graph (the LRM is what the graph was extracted from); the question is whether the graph faithfully describes the original model's mechanisms, not whether it is internally consistent. Running the ablation in the original model probes the gap.

\paragraph{Results.} Table~\ref{tab:perturbation} reports the Spearman and Pearson correlations across the validation set, broken down by target stream.

\begin{table}[H]
\centering
\caption{Pairwise mechanistic faithfulness via single-source ablation in the original FLUX.1[schnell], broken down by target stream. Top $K = 30$ feature vertices of the per-position graph per target; correlations over their ordered pairs. Statistics are taken over $86$ targets ($51$ image, $35$ text).}
\label{tab:perturbation}
\begin{tabular}{lcccccccc}
\toprule
& \multicolumn{4}{c}{Image targets ($n=51$)} & \multicolumn{4}{c}{Text targets ($n=35$)} \\
\cmidrule(lr){2-5} \cmidrule(lr){6-9}
Metric & Mean $\uparrow$ & Median $\uparrow$ & Min & Max & Mean $\uparrow$ & Median $\uparrow$ & Min & Max \\
\midrule
Spearman $\rho$ & 0.676 & 0.693 & 0.346 & 0.895 & 0.545 & 0.563 & 0.323 & 0.730 \\
Pearson $r$     & 0.769 & 0.778 & 0.610 & 0.910 & 0.744 & 0.764 & 0.368 & 0.931 \\
\bottomrule
\end{tabular}
\end{table}

The image-stream Spearman correlation (mean $0.68$, median $0.69$) is comparable to the mean of $0.72$ that \citet{ameisen2025circuit} report for cross-layer transcoders on a language model, in a related but not identical ablation test, indicating that per-layer transcoders on double-stream MM-DiT blocks capture the underlying mechanism with comparable fidelity to that prior work.

\begin{figure}[H]
\centering
\includegraphics[width=\linewidth]{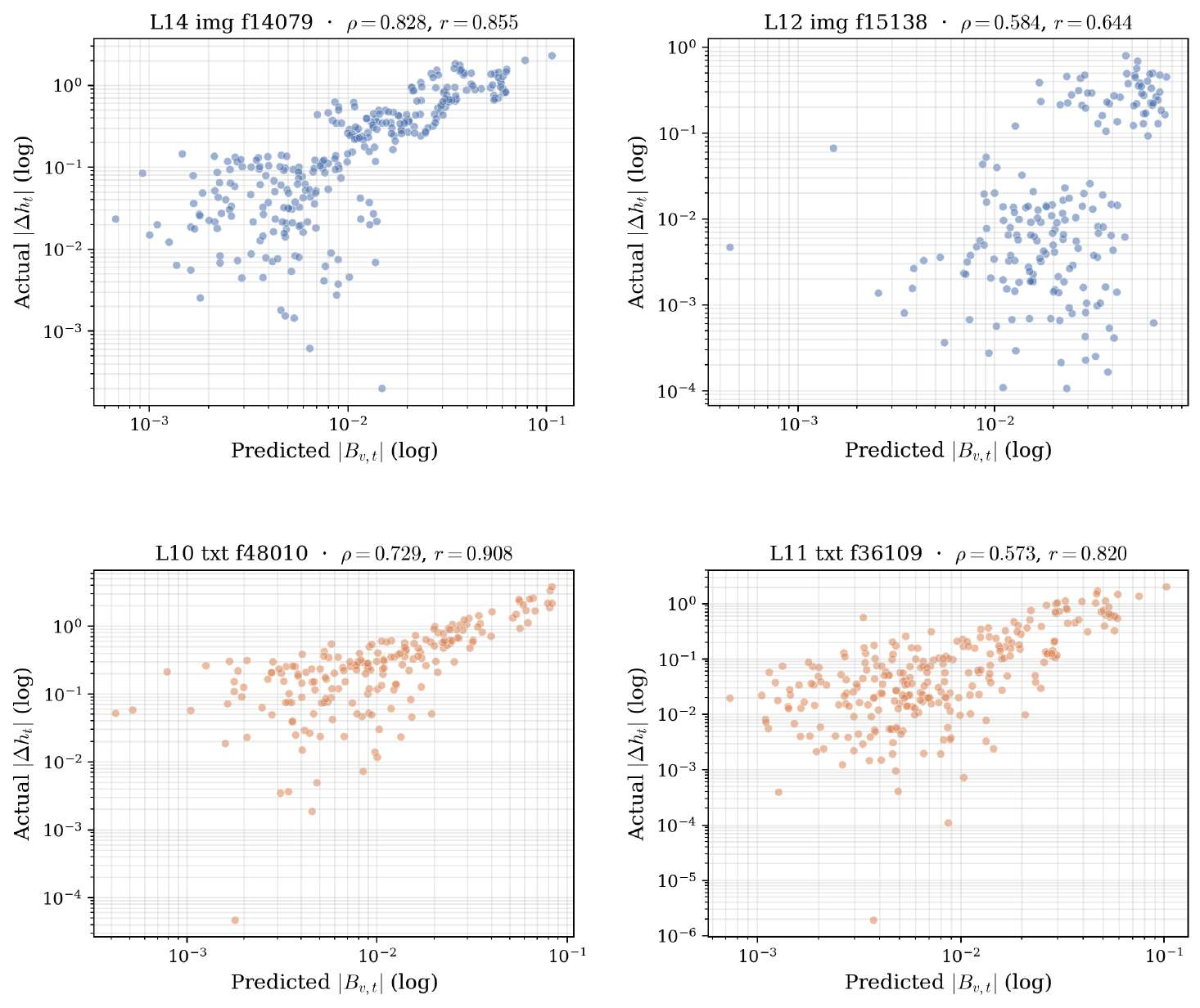}
\caption{Pairwise perturbation faithfulness scatter for four representative targets, in log-log coordinates: predicted indirect influence $|B_{v, t}|$ from the attribution graph (x-axis) versus actual ablation effect $|\Delta h_t|_{\mathrm{actual}}$ in the original FLUX.1[schnell] (y-axis). Each point is an ordered pair $(v, t)$ with $v, t \in \mathcal{S}$ and $v \neq t$. \textbf{Top:} two image-target examples. \textbf{Bottom:} two text-target examples. Per-graph Spearman $\rho$ and Pearson $r$ shown in titles. Note the diagonal-cloud geometry shared across all panels and the wider vertical spread among low/mid-influence pairs in text-target panels, which drives the larger Pearson-Spearman gap in the text stream.}
\label{fig:perturbation_scatter}
\end{figure}

\paragraph{Pearson-Spearman gap.} Pearson medians ($0.78$/$0.76$ image/text) systematically exceed Spearman medians ($0.69$/$0.56$). This gap reflects the structure of the predicted-actual scatter, illustrated in Figure~\ref{fig:perturbation_scatter}: in log-log coordinates, $|\Delta h_t|_{\mathrm{actual}}$ traces $|B_{v, t}|_{\mathrm{predicted}}$ as a diagonal cloud over roughly two decades of predicted influence and three or more decades of actual effect, with substantial vertical scatter at fixed $|B_{v, t}|$. Pearson, computed in linear space, is dominated by the small number of high-influence pairs whose contribution to the variance is large; the linear relationship there is well captured. Spearman ranks all pairs and is sensitive to the vertical scatter at low and intermediate predicted values, where pairs with similar $|B_{v, t}|$ can have actual effects differing by an order of magnitude or more.

\paragraph{Why text-stream Spearman is lower.} The text-stream Spearman is somewhat lower (median $0.56$ vs $0.69$), even though Pearson is comparable across streams ($0.76$ vs $0.78$). The Pearson-Spearman gap is therefore noticeably larger for text ($0.20$) than for image ($0.09$). Per-graph examples in Figure~\ref{fig:perturbation_scatter} (bottom row) show the mechanism directly: text-target scatters have a tightly aligned high-influence cluster (which Pearson captures cleanly) coexisting with a wider vertical spread among low- and mid-influence pairs (where actual $|\Delta h_t|$ varies over an order of magnitude at fixed predicted $|B_{v, t}|$). This vertical spread at fixed predicted value is what drives Spearman down without affecting Pearson, since rank order among pairs with similar $|B_{v, t}|$ is determined by noise. We additionally note that the per-edge attribution distribution among kept text-stream features is heavier-tailed than for image (Gini $0.52$ vs $0.49$, $10$th-percentile $|A| = 0.022$ vs $0.039$), although whether this distributional asymmetry causally drives the wider vertical spread or both reflect a common upstream cause is something we cannot disentangle from these data.

\subsection{Hyperparameters}
\label{app:validation:hyperparams}

Table~\ref{tab:hyperparams} consolidates all numerical parameters used throughout the pipeline.

\begin{table}[H]
\centering
\small
\caption{Pipeline hyperparameters (FLUX.1[schnell]). Table~\ref{tab:backbones} lists what differs on FLUX.1[dev] and SD3.5-medium.}
\label{tab:hyperparams}
\begin{tabular}{lll}
\toprule
\textbf{Section} & \textbf{Parameter} & \textbf{Value} \\
\midrule
Base model & FLUX.1[schnell], denoising steps & $4$ \\
& Resolution & $512 \times 512$ \\
& Guidance scale & $0.0$ \\
& Precision & bfloat16 \\
\midrule
Transcoders & $d_{\mathrm{model}}$ & $3072$ \\
& Expansion factor & $16$ \\
& $d_{\mathrm{feat}}$ & $49\,152$ \\
& Time embedding $d_t$ & $256$ \\
& Time MLP layers & $2$ (SiLU) \\
& Activation & ReLU \\
& Decoder column normalization & after every step \\
\midrule
Training & Optimizer & AdamW \\
& Weight decay & $0$ \\
& Learning rate & $2 \times 10^{-4}$ \\
& LR schedule & cosine annealing over $256$ cycles \\
& Batch size & $4096$ \\
& Buffer size & $10^6$ pairs \\
& Cycles & $256$ \\
& $\lambda^{\mathrm{img}}$ & $3 \times 10^{-4}$ \\
& $\lambda^{\mathrm{txt}}$ & $5 \times 10^{-5}$ \\
& Variance normalization $\varepsilon$ & $10^{-6}$ \\
& Prompt corpus & yvdao/midjourney-v6 ($\sim$310k prompts) \\
& Prompt length filter & $> 16$ chars, truncate at $512$ \\
& Activation buffers & bfloat16 \\
\midrule
LRM & Analyzed blocks & $\ell \in \{0, \ldots, 15\}$ \\
& Streams & img, txt \\
& Transcoders precision & float32 (TF32 disabled) \\
\midrule
Iterative construction & Min-attribution threshold $\tau$ & $10^{-3}$ \\
& Per-iteration batch size $k$ & $50$ \\
& VJP budget $N_{\max}$ & $1000$ \\
\midrule
Pruning & Vertex threshold $\theta^s_v$ (img, txt) & $0.80$, $0.80$ \\
& Edge threshold $\theta^s_e$ (img, txt) & $0.98$, $0.98$ \\
\midrule
Perturbation evaluation & Sources per graph & $K = 30$ \\
& Source position & the vertex's own position \\
\bottomrule
\end{tabular}
\end{table}

\section{Other backbones: FLUX.1[dev] and SD3.5-medium}
\label{app:backbones}

The case studies of §\ref{sec:experiments} use FLUX.1[schnell]. To see how much of the method and of its two quantitative results depends on that model, we trained $32$ transcoders with the same recipe for each of two further MM-DiTs, FLUX.1[dev]~\cite{bfl2024flux} and SD3.5-medium~\cite{stabilityai2024sd35medium}, and repeated on each the temporal analysis of §\ref{sec:exp:temporal} and the matched-budget color-prior comparison of §\ref{sec:exp:color_prior}. The transcoder recipe, the tracing and pruning parameters and the intervention strength are those of FLUX.1[schnell] (Table~\ref{tab:hyperparams}); Table~\ref{tab:backbones} lists what differs.

\begin{table}[H]
\centering
\caption{What differs between the three backbones; everything else is as in Table~\ref{tab:hyperparams}.}
\label{tab:backbones}
\resizebox{\textwidth}{!}{%
\begin{tabular}{llll}
\toprule
 & FLUX.1[schnell] & FLUX.1[dev] & SD3.5-medium \\
\midrule
Transformer blocks & $19$ double-stream, $38$ single-stream & $19$ double-stream, $38$ single-stream & $24$ joint, image-only second attention in $0$--$12$ \\
Transcoders & blocks $0$--$15$, both streams & blocks $0$--$15$, both streams & blocks $0$--$15$, both streams \\
$d_{\mathrm{model}}$ / $d_{\mathrm{feat}}$ & $3072$ / $49{,}152$ & $3072$ / $49{,}152$ & $1536$ / $24{,}576$ \\
Text positions $S_{\mathrm{txt}}$ & $512$ (T5) & $512$ (T5) & $333$ ($77$ CLIP, $256$ T5) \\
Denoising steps & $4$ & $50$ & $40$ \\
Guidance & none & embedded, scale $3.5$ & classifier-free, scale $4.5$ \\
Steps of the temporal analysis & $0$, $1$, $2$, $3$ & $0$, $19$, $33$, $42$ & $0$, $20$, $29$, $35$ \\
Buffer per cycle (records) & $10^6$ & $1.6 \times 10^6$ & $2.6 \times 10^6$ \\
Training prompts ($256$ cycles) & $65{,}536$ & $8{,}192$ & $8{,}192$ \\
\bottomrule
\end{tabular}%
}
\end{table}

\subsection{Architectures}
\label{app:backbones:arch}

The construction of §\ref{sec:method} needs stream-specific MLP sublayers that add to the residual stream through an AdaLN-Zero gate, a modulation that is constant within a graph, and joint attention over the concatenated streams. Both backbones have these properties; what differs is the following.

\paragraph{FLUX.1[dev]} has the architecture of FLUX.1[schnell] but is not timestep-distilled: it samples in $50$ steps on a shifted noise schedule, and it takes the guidance scale as an input embedding that enters the AdaLN modulation together with the timestep and the pooled CLIP embedding. Its modulation is therefore constant within a graph at a fixed timestep \emph{and} a fixed guidance scale. All FLUX.1[dev] results here are at the scale $3.5$ at which its transcoders were trained.

\paragraph{SD3.5-medium} consists of $24$ joint blocks, each with separate image and text weights, with $d_{\mathrm{model}} = 1536$. Its text stream concatenates $77$ CLIP positions with $256$ T5 positions, so $S_{\mathrm{txt}} = 333$. Blocks $0$--$12$ have a second attention sublayer over the image tokens only, with its own AdaLN shift, scale and gate. The LRM freezes it in the same way as the joint attention -- cached probabilities, an affine OV pathway and a cached reconstruction error -- and its constant part and its error enter $b^*_{\mathrm{eff}}$ through its own gate. SD3.5-medium samples in $40$ steps with classifier-free guidance at scale $4.5$. A graph decomposes a feature of the conditional copy of the batch. The transcoders were trained on the records of both conditional and unconditional copies, and an intervention scales a feature in both.

\subsection{Transcoders and the replacement model}
\label{app:backbones:tc}

The recipe of Appendix~\ref{app:transcoders} is used without change, apart from the buffer size and the number of prompts per cycle (Table~\ref{tab:backbones}). At the end of training nMSE of the FLUX.1[dev] transcoders is $0.04$--$0.23$ for the image stream and $0.001$--$0.029$ for the text stream, with $L_0$ of $137$--$584$ and $16$--$232$ active features per token. The SD3.5-medium transcoders reconstruct the image stream better and are far denser at the same $\lambda$: nMSE $0.013$--$0.062$ and $0.002$--$0.037$, $L_0$ of $867$--$2{,}986$ and $21$--$1{,}253$. Their feature activations are about an order of magnitude smaller than FLUX's: at $\sigma = 1$ the targets of Appendix~\ref{app:backbones:temporal} have a median preactivation of $2.0$, against $21.8$ on FLUX.1[dev]. The sparsity term, unlike the faithfulness term, is not normalized by the scale of the MLP output, so the same $\lambda$ penalizes these smaller activations less; we did not retune it. Tables~\ref{tab:backbones_repl} and~\ref{tab:backbones_lrm} repeat the end-to-end checks of Tables~\ref{tab:tc_e2e} and~\ref{tab:lrm_exactness} on the same $512$ prompts, with the same seeds, and with the LRM built at the four steps whose noise level matches FLUX.1[schnell]'s.

\begin{table}[H]
\centering
\caption{End-to-end faithfulness of the full replacement model (all $32$ MLPs replaced by their transcoders at every step), on $512$ prompts at $512 \times 512$ resolution. Latent MSE is in each backbone's own latent space. The FLUX.1[schnell] row repeats Table~\ref{tab:tc_e2e}.}
\label{tab:backbones_repl}
\begin{tabular}{lcccc}
\toprule
& \multicolumn{2}{c}{Latent Cosine Similarity $\uparrow$} & \multicolumn{2}{c}{Latent MSE $\downarrow$} \\
\cmidrule(lr){2-3} \cmidrule(lr){4-5}
Backbone & Mean & Median & Mean & Median \\
\midrule
FLUX.1[schnell] ($4$ steps) & 0.7839 & 0.7960 & 0.4786 & 0.4313 \\
FLUX.1[dev] ($50$ steps) & 0.8378 & 0.8611 & 0.4811 & 0.4294 \\
SD3.5-medium ($40$ steps) & 0.9217 & 0.9313 & 0.1909 & 0.1606 \\
\bottomrule
\end{tabular}
\end{table}

\begin{table}[H]
\centering
\small
\caption{End-to-end LRM exactness on $512$ prompts. Each row corresponds to building the LRM at a single noise level $\sigma$ and replacing only that step's transformer call. Latent cosine similarity, mean / median. The FLUX.1[schnell] column repeats Table~\ref{tab:lrm_exactness}.}
\label{tab:backbones_lrm}
\begin{tabular}{lcccc}
\toprule
& & & \multicolumn{2}{c}{SD3.5-medium} \\
\cmidrule(lr){4-5}
LRM step & FLUX.1[schnell] & FLUX.1[dev] & both halves & conditional half \\
\midrule
$\sigma = 1$  & 0.9854 / 0.9889 & 0.9943 / 0.9983 & 0.9602 / 0.9788 & 0.9649 / 0.9819 \\
$\sigma \approx 0.75$  & 0.9982 / 0.9986 & 0.9996 / 0.9997 & 0.9994 / 0.9997 & 0.9994 / 0.9997 \\
$\sigma \approx 0.5$  & 0.9997 / 0.9997 & 0.9999 / 0.9999 & 0.9998 / 0.9999 & 0.9998 / 0.9999 \\
$\sigma \approx 0.25$  & 1.0000 / 1.0000 & 1.0000 / 1.0000 & 0.9999 / 0.9999 & 0.9999 / 0.9999 \\
\bottomrule
\end{tabular}
\end{table}

\subsection{Attribution graphs}
\label{app:backbones:graphs}

The graphs are traced with the parameters of FLUX.1[schnell] ($\tau = 10^{-3}$, $k = 50$, $N_{\max} = 1000$, $80\%$ vertex and $98\%$ edge pruning per stream). One graph took $46$ minutes on average on FLUX.1[dev] ($35$--$65$) and $23$ minutes on SD3.5-medium ($17$--$33$), on one NVIDIA A100 (80\,GB).

\begin{table}[H]
\centering
\caption{Faithfulness of the graphs. Statistics are taken over the $80$ graphs of the temporal analysis, $20$ image-stream targets at four noise levels. Mean / median over graphs.}
\label{tab:backbones_graphs}
\resizebox{\textwidth}{!}{%
\begin{tabular}{lccccc}
\toprule
Graphs & Raw $\delta$ (\%) & Pruned $\delta$ (\%) & $|h^* - b^*_{\mathrm{eff}}|$ & Spearman $\rho$ & Pearson $r$ \\
\midrule
FLUX.1[dev] & 25.1 / 20.3 & 27.7 / 24.3 & 16.9 / 11.8 & 0.619 / 0.636 & 0.728 / 0.756 \\
SD3.5-medium & 60.6 / 42.6 & 62.3 / 41.1 & 2.0 / 1.8 & 0.584 / 0.566 & 0.700 / 0.697 \\
\bottomrule
\end{tabular}%
}
\end{table}

By the conservation invariant (Table~\ref{tab:backbones_graphs}) the new graphs look much less complete than the FLUX.1[schnell] validation graphs of Appendix~\ref{app:validation}, but most of the gap is in the denominator of $\delta$ rather than in the mass the thresholds leave out. On FLUX.1[dev] the absolute residual is close to FLUX.1[schnell]'s (median $2.55$ against $3.10$), while $|h^* - b^*_{\mathrm{eff}}|$ is about a quarter of it (median $11.8$ against $50.5$). SD3.5-medium's features are an order of magnitude weaker (Appendix~\ref{app:backbones:tc}), so $|h^* - b^*_{\mathrm{eff}}|$ is below $2$ in $45$ of its $80$ graphs, where a residual of under one unit is a large relative error; on the other $35$ the median raw $\delta$ is $26.5\%$. Under single-source ablation in the original model the graphs are somewhat less faithful than FLUX.1[schnell]'s image targets, with median Spearman correlations of $0.64$ on FLUX.1[dev] and $0.57$ on SD3.5-medium against $0.69$. Along the trajectory the correlation improves on FLUX.1[dev], from $0.56$ at $\sigma = 1$ to $0.66$ at $\sigma \approx 0.25$, and falls on SD3.5-medium, from $0.68$ to $0.50$.

\subsection{Temporal evolution}
\label{app:backbones:temporal}

\begin{figure}[H]
\centering
\includegraphics[width=\linewidth]{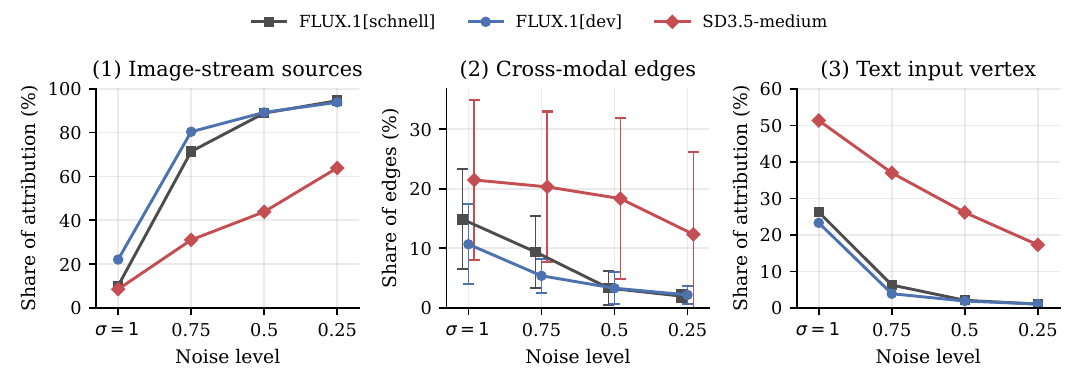}
\caption{Temporal evolution on the three backbones, $20$ targets each, at the four noise levels of FLUX.1[schnell] (FLUX.1[dev]: steps $0$, $19$, $33$, $42$ of $50$; SD3.5-medium: $0$, $20$, $29$, $35$ of $40$). \textbf{(1)} Share of absolute attribution from image-stream sources, over all source types, as in Figure~\ref{fig:temporal}. \textbf{(2)} Fraction of cross-modal edges; error bars show one standard deviation over targets. \textbf{(3)} Share of absolute attribution carried by the text input vertex, the direct path from the text embedding.}
\label{fig:temporal_backbones}
\end{figure}

We keep the $20$ targets of §\ref{sec:exp:temporal} as (prompt, seed, layer) triples, all in the image stream. Each backbone has its own dictionary, so the feature is selected again, at $\sigma = 1$. The graph is traced at the steps whose noise level matches FLUX.1[schnell]'s four (Table~\ref{tab:backbones}), each time at the position of the feature's largest preactivation at that step, as on FLUX.1[schnell]. The statistics are those of Figure~\ref{fig:temporal}, computed by the same code (Table~\ref{tab:backbones_temporal}, Figure~\ref{fig:temporal_backbones}).

\begin{table}[H]
\centering
\caption{Temporal evolution on the three backbones: share of absolute attribution from image-stream sources and fraction of cross-modal edges (percent, mean $\pm$ standard deviation over the $20$ targets).}
\label{tab:backbones_temporal}
\resizebox{\textwidth}{!}{%
\begin{tabular}{lcccccc}
\toprule
& \multicolumn{2}{c}{FLUX.1[schnell]} & \multicolumn{2}{c}{FLUX.1[dev]} & \multicolumn{2}{c}{SD3.5-medium} \\
\cmidrule(lr){2-3} \cmidrule(lr){4-5} \cmidrule(lr){6-7}
Noise level & Image & Cross-modal & Image & Cross-modal & Image & Cross-modal \\
\midrule
$\sigma = 1$ & $10.1 \pm 4.8$ & $14.9 \pm 8.4$ & $22.1 \pm 18.1$ & $10.7 \pm 6.7$ & $8.5 \pm 4.0$ & $21.5 \pm 13.5$ \\
$\sigma \approx 0.75$ & $71.4 \pm 23.3$ & $9.4 \pm 6.1$ & $80.4 \pm 28.6$ & $5.4 \pm 2.9$ & $31.0 \pm 19.6$ & $20.3 \pm 12.7$ \\
$\sigma \approx 0.5$ & $88.9 \pm 16.6$ & $3.3 \pm 2.9$ & $89.2 \pm 22.6$ & $3.3 \pm 2.7$ & $43.8 \pm 27.5$ & $18.3 \pm 13.6$ \\
$\sigma \approx 0.25$ & $94.6 \pm 8.1$ & $2.0 \pm 0.9$ & $93.9 \pm 16.9$ & $2.2 \pm 1.6$ & $63.9 \pm 33.4$ & $12.4 \pm 13.8$ \\
\bottomrule
\end{tabular}%
}
\end{table}

\paragraph{FLUX.1[dev] reproduces the result of FLUX.1[schnell].} At $\sigma \approx 0.5$ and $0.25$ the two FLUX models are indistinguishable: $89.2\%$ against $88.9\%$ and $93.9\%$ against $94.6\%$ image share, $3.3\%$ against $3.3\%$ and $2.2\%$ against $2.0\%$ cross-modal edges. All $20$ targets end with a higher image share than they start with, as on FLUX.1[schnell], and $18$ with a lower cross-modal fraction (FLUX.1[schnell]: $20$); the image share rises monotonically for $13$ of the $20$ on both models. The start differs, $22.1\%$ against $10.1\%$, and the difference comes from six targets, whose selected features are barely active at $\sigma = 1$ (preactivations $2.0$--$4.7$, against $10.9$--$51.4$ for the other fourteen) and whose small graphs are far more image-driven from the start. Without them FLUX.1[dev] reads $11.8$, $73.5$, $85.1$ and $91.5\%$, against $10.3$, $70.7$, $86.6$ and $93.2\%$ for the same fourteen targets on FLUX.1[schnell]. The per-layer profile of Figure~\ref{fig:temporal} (right) transfers with one difference. Summed over the five layer-$12$ targets at the last level, $88\%$ of the image-stream edges on FLUX.1[schnell] and $80\%$ on FLUX.1[dev] come from layers $0$--$1$ and $4$--$7$, and layer $3$ is nearly inactive on both; but layer $2$, nearly inactive on FLUX.1[schnell] ($177$ edges), contributes on FLUX.1[dev] ($14{,}123$ edges, $6\%$).

\paragraph{SD3.5-medium moves in the same direction, halfway.} On average the image share rises and the cross-modal fraction falls from level to level ($8.5\% \to 63.9\%$ and $21.5\% \to 12.4\%$), but at $\sigma \approx 0.25$ the image stream holds $64\%$ of the attribution where both FLUX models hold $94\%$; $19$ of the $20$ targets end with a higher image share than they start with and $6$ above $90\%$, but only $12$ with a lower cross-modal fraction. The text-side mass is mostly not carried by text-stream features. Half of all attribution at $\sigma = 1$ comes straight from the text input vertex, the direct path from the CLIP and T5 embeddings ($51.3\%$, against $26.2\%$ and $23.3\%$ on the two FLUX models; Figure~\ref{fig:temporal_backbones}, right), and at $\sigma \approx 0.25$ this path still carries $17.2\%$ ($1.0\%$ and $1.1\%$ on the two FLUX models). Image-stream growth is also placed differently: for the layer-$12$ targets, $79\%$ of the image-stream edges at the last level come from the middle layers $4$--$9$ and almost none from layers $0$ and $1$ ($130$ and $105$ edges, against $19{,}821$ and $61{,}891$ on FLUX.1[dev]). With the same layer indices the targets sit at blocks $9$--$14$ of SD3.5-medium's $24$ joint blocks rather than of FLUX's $19$ double-stream blocks, and SD3.5-medium runs classifier-free guidance.

\subsection{Color prior at a matched budget}
\label{app:backbones:color}

The grid of Table~\ref{tab:matched_budget} is reused: the same $8$ prompts, $30$ seeds, $6$ conditions and $\alpha = -80$ on every intervened feature at every step. The target and the companion sets are found again in each dictionary, by rules fixed before the generation. The three activation-based rules of §\ref{sec:exp:color_prior} are applied as on FLUX.1[schnell], to the text-stream features of layers $0$--$7$ and the same $48$-prompt corpus. The graph set was picked on a new dictionary using the target's attribution graph.

\paragraph{Labels.} Human labels exist only for FLUX.1[schnell], so every generation of the three backbones, the FLUX.1[schnell] grid included, is judged by one vision-language model, Qwen3.6-35B-A3B~\cite{qwen2026qwen36} in bfloat16 with greedy decoding; Appendix~\ref{app:blind:vlm} compares it with the annotators on every FLUX.1[schnell] experiment. It sees the prompt and the baseline and intervened images side by side and answers with the label set of the human study, after an explicit first decision on whether the named object is still recognizable. On the FLUX.1[schnell] matched-budget grid it agrees with the annotator on $87.6\%$ of the $960$ frames of the four conditions with companions (Cohen's $\kappa = 0.76$), marks as destroyed $236$ of the $268$ objects the annotator marked destroyed, reaches the annotator's decision against the baseline on object destruction in all $32$ prompt--condition cells and on the requested color in $23$ of them, and ranks the six conditions of Table~\ref{tab:matched_budget} in the annotator's order by the requested-color rate. It is more lenient about a partly recolored object: it judges $53.8\%$ of the baseline generations to show the requested color, the annotator $36.2\%$. We therefore report the pairs whose baseline the labeler itself judges red (Table~\ref{tab:backbones_color}); the requested color is the prior fixed and a destroyed object is the prior's object lost, whatever the strength of the prior on that backbone.

\begin{table}[H]
\centering
\caption{Color prior at a matched budget on the three backbones: the grid of Table~\ref{tab:matched_budget}, every frame labeled by the same vision-language model. Pairs whose baseline is judged red ($7$, $5$ and $6$ of the $8$ prompts contribute); percent of them in which the object takes the requested color or is destroyed.}
\label{tab:backbones_color}
\resizebox{\textwidth}{!}{%
\begin{tabular}{lcccccc}
\toprule
& \multicolumn{2}{c}{FLUX.1[schnell] ($n = 111$)} & \multicolumn{2}{c}{FLUX.1[dev] ($n = 83$)} & \multicolumn{2}{c}{SD3.5-medium ($n = 99$)} \\
\cmidrule(lr){2-3} \cmidrule(lr){4-5} \cmidrule(lr){6-7}
Selection of the $3$ companions & Requested color & Destroyed & Requested color & Destroyed & Requested color & Destroyed \\
\midrule
none, target feature only & 21.6 & 0.0 & 19.3 & 2.4 & 22.2 & 0.0 \\
by activation on prompt tokens & 52.3 & 0.0 & 19.3 & 42.2 & 36.4 & 1.0 \\
by contrast against a neutral prompt & 17.1 & 82.0 & 22.9 & 77.1 & 24.2 & 63.6 \\
by co-activation over a $48$-prompt corpus & 72.1 & 18.0 & 62.7 & 31.3 & 41.4 & 57.6 \\
\textbf{from the attribution graph (ours)} & \textbf{79.3} & \textbf{13.5} & \textbf{81.9} & \textbf{7.2} & \textbf{65.7} & \textbf{33.3} \\
\bottomrule
\end{tabular}%
}
\end{table}

\paragraph{Results.} On FLUX.1[dev] the graph set fixes the prior most often, $81.9\%$ against $62.7\%$ for co-activation, the next best, and destroys the object less often, $7.2\%$ against $31.3\%$. Activation-based selection, harmless on FLUX.1[schnell], destroys $42.2\%$ of the objects here. On SD3.5-medium the graph set again leads co-activation ($65.7\%$ against $41.4\%$ with the requested color, $33.3\%$ against $57.6\%$ destroyed), but it loses a third of the objects. On FLUX.1[schnell] the VLM agrees with the annotator that the graph and co-activation sets do not differ in success ($79.3\%$ against $72.1\%$ with the requested color), and its destruction difference, $13.5\%$ against $18.0\%$, is in the annotator's direction without being significant (over all pairs the annotator's labels give $13.8\%$ against $21.7\%$, Table~\ref{tab:matched_budget}).

\subsection{Summary}
\label{app:backbones:summary}

The transcoders, the LRM and the graph pipeline run on both backbones with the parameters of FLUX.1[schnell]. The transcoders reproduce the original model end to end at least as well as on FLUX.1[schnell], and the LRM is exact to the same degree (Tables~\ref{tab:lrm_exactness} and~\ref{tab:backbones_lrm}). FLUX.1[dev] reproduces both temporal evolution and color prior results. On SD3.5-medium both hold in direction -- attribution moves from the text stream to the image stream along the trajectory, and the graph companions fix the prior more often and destroy the object less often than co-activation -- but the temporal shift is half complete at $\sigma \approx 0.25$, and the graph set still destroys a third of the objects.

\section{Feature interpretation}
\label{app:interpretation}

The attribution graph treats transcoder features as the basic units of analysis, so its usefulness depends on these features corresponding to meaningful visual or textual concepts rather than arbitrary directions in activation space. In this section we describe a two-pass procedure for finding interpretable features in the transcoder dictionary by their top-activating examples and show qualitative results on representative blocks. For this analysis we examine three blocks: $\ell = 6$ (early), $\ell = 12$ (middle), and $\ell = 18$ (late); the layer-$18$ transcoders are those trained with the same recipe for the comparison with SAEs (§\ref{sec:exp:sae}). The evolution from $\ell = 6$ through $\ell = 12$ to $\ell = 18$ spans most of the double-stream segment and is informative for tracking how concepts develop with depth.

\subsection{Methodology}
\label{app:interp:method}

\paragraph{Activation statistics.} A corpus of $100\,000$ prompts from \texttt{yvdao/midjourney-v6} is run through the frozen FLUX.1[schnell] pipeline. For every prompt, every denoising step $t \in \{0, 1, 2, 3\}$, and every transcoder feature $f$ we record the maximum activation per-prompt.
\begin{equation}
v_t(f \mid \mathrm{prompt}) \;=\; \max_p \bigl(z^{(\ell, s, f)}(p)\bigr)
\end{equation}
where the maximum is taken over the prompt's image-stream patches ($s = \mathrm{img}$) or text-stream tokens ($s = \mathrm{txt}$). For each feature, we maintain three running quantities across the corpus: the top-$K$ ($K = 5$) prompts by $v_t(f \mid \cdot)$, sufficient statistics for the mean $\bar{a}_t(f)$ and standard deviation $\sigma_t(f)$ of activations, and the number of prompts on which $f$ ranks among the top-$M$ ($M = 128$) most active features.

\paragraph{Feature selection.} Out of $d_{\mathrm{feat}} = 49\,152$ features per transcoder we select $256$ for visualization. The search behind the case studies (§\ref{sec:experiments}) ranked features by the same score across layers, denoising steps and both streams and inspected roughly $2{,}000$ of them; this section shows three blocks.
For each feature $f$ and denoising step $t$, let $v^{\max}_t(f)$ denote the highest per-prompt maximum activation recorded at step $t$. We define the normalized activation strength and activation frequency as
\begin{equation}
Z_t(f) \;=\; \frac{v^{\max}_t(f) - \bar{a}_t(f)}{\sigma_t(f) + \varepsilon}, \qquad q_t(f) \;=\; \frac{|\{i : f \in \mathrm{TopM}_i^t\}|}{N},
\end{equation}
where $N$ is the size of the prompt corpus, $\bar{a}_t(f)$ and $\sigma_t(f)$ are the mean and standard deviation of maximum activations at step $t$, and $\mathrm{TopM}_i^t$ is the set of the top-$M$ most active features for prompt $i$ at timestep $t$.
The final selection score for a feature is computed as
\begin{equation}
\mathrm{score}(f) \;=\; \max_t \;Z_t(f) \cdot \sqrt{q_t(f)}.
\end{equation}
The first factor $Z_t(f)$ rewards features that produce sharp, high-confidence peak activations on certain prompts. The second factor $\sqrt{q_t(f)}$ penalizes features that activate strongly but too rarely -- i.e., those likely to be narrow artifacts triggered by only a few specific prompts.

\paragraph{Activation maps.} 
For each selected feature, we re-run the union of its top-$5$ activating prompts through the model while recording the full per-position activation map $\{z^{(\ell,s,f)}(p)\}_p$. These maps form the basis of all visualizations below.
In the image stream, the activation map (of length $S_{\mathrm{img}} = 1024$) is reshaped into a $32 \times 32$ patch grid corresponding to the $512 \times 512$ image and overlaid on the generated image. For text-stream features, the map assigns one activation value per prompt token and is visualized as a color overlay on the prompt text. Activations below 20\% of the per-example maximum are suppressed for clarity.
Additionally, we compute the mean activation of each feature across its top-$5$ prompts, broken down by denoising timestep, to reveal temporal specialization patterns.

\subsection{Early layer ($\ell = 6$) results}
\label{app:interp:layer6}

\begin{figure}[h]
\centering
\includegraphics[width=\linewidth]{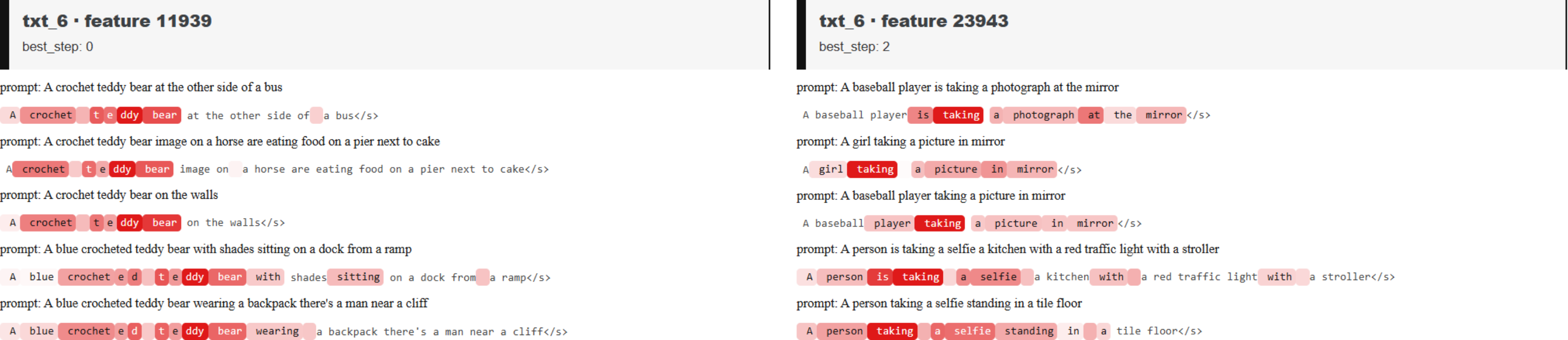}
\caption{Representative text-stream features at $\ell = 6$. \textbf{Left:} \texttt{txt-6-11939} (teddy bear). \textbf{Right:} \texttt{txt-6-23943} (taking a photo). Each row shows the top activating prompts for the feature, with per-token activation rendered as color intensity.}
\label{fig:feat_txt6}
\end{figure}

\paragraph{Text stream.} Text features at $\ell = 6$ are tightly bound to surface lexical content. Feature \texttt{txt-6-11939} fires on the phrase \emph{``crochet teddy bear''}, highlighting all three tokens whenever they appear together; \texttt{txt-6-36869} groups attributes of a franchise (\emph{``Mickey Mouse''}, \emph{``Disney World''}). Other prominent features in this group include action verbs (\texttt{txt-6-23943}: \emph{``taking a photograph''}/\emph{``selfie''}; \texttt{txt-6-26919}: \emph{``typing on keyboard''}), spatial-relation phrases (\texttt{txt-6-15466}: \emph{``stacked on each other''}; \texttt{txt-6-23336}: \emph{``on both sides''}), object-state descriptors (\texttt{txt-6-28486}: \emph{``empty store shelf''}), and what appears to be implicit color compositions: \texttt{txt-6-47738} fires on \emph{``Irish flags''}, \emph{``Mexico''}, and \emph{``Santa''} prompts, the common factor being a green/red/white palette. The interpretability rate at this depth is high: nearly every visualized feature corresponds to an identifiable lexical or semantic category.

\paragraph{Image stream.} Image features at $\ell = 6$ encode graphical primitives. A geometry-oriented group includes \texttt{img-6-5297} (vertical edges of monitors, bottles, doorframes), \texttt{img-6-31656} (diagonal lines on smartphone bezels, power lines, ski poles), \texttt{img-6-17202} (thin suspended cables and wires), and \texttt{img-6-48604} (regular grid and lattice patterns). A color-oriented group includes \texttt{img-6-15493} (red objects: life vests, jackets, plastic buckets) and \texttt{img-6-36726} (regions of pure white). Particularly notable is \texttt{img-6-2366}, which fires on the \emph{boundary} between blue/green and red regions independently of the underlying objects: active patches lie strictly along the seam of these two color regimes.

\begin{figure}[h]
\centering
\includegraphics[width=\linewidth]{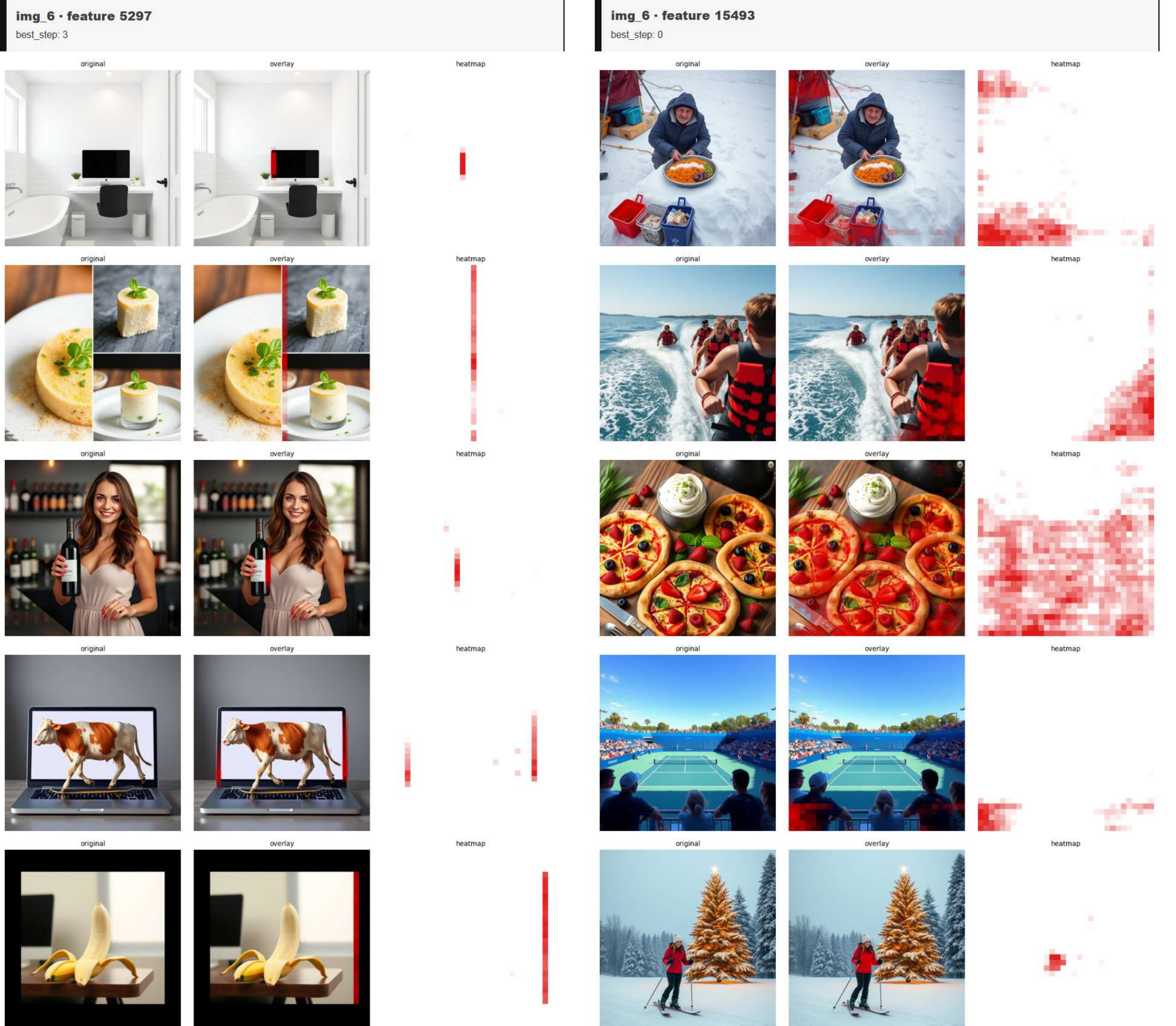}
\caption{Representative image-stream features at $\ell = 6$. \textbf{Left:} \texttt{img-6-5297} (vertical edges). \textbf{Right:} \texttt{img-6-15493} (red regions). For each feature we show the original generated image, the activation overlay, and the activation map alone.}
\label{fig:feat_img6}
\end{figure}

A temporal split is already visible at this depth. The geometry-oriented features (\texttt{5297}, \texttt{31656}, \texttt{2366}, \texttt{48604}) peak at denoising steps $2$--$3$, while the color-oriented features (\texttt{15493}, \texttt{36726}) peak at steps $0$--$1$. This is consistent with the iterative coarse-to-fine progression of diffusion sampling: bulk colors are placed first, fine geometric structure is sharpened later.

\subsection{Middle layer ($\ell = 12$) results}
\label{app:interp:layer12}

\paragraph{Text stream.} Text features at $\ell = 12$ assemble compositional concepts beyond the per-word level. \texttt{txt-12-40834} fires on personal names independently of context (\emph{``Matt Wieters''}, \emph{``Rachel Ray''}, \emph{``Jeff Bridges''}). Quantifier features appear: \texttt{txt-12-5888} on layout phrases (\emph{``Four photos''}, \emph{``Four square images''}) and \texttt{txt-12-33890} on plurality (\emph{``Several different kites''}, \emph{``Many white and yellow double decker bus''}). At the same time, some features have already lost their lexical anchor: \texttt{txt-12-43210} fires exclusively on the end-of-sequence token.

\paragraph{Image stream.} The middle layer shows the highest density of features with identifiable semantic referents. Object-level features include \texttt{img-12-8630} (bicycles), \texttt{img-12-22268} (wine-bottle necks, with activation strictly above the label), \texttt{img-12-244} (hanging vertical structures: chains, ropes, water streams), \texttt{img-12-25382} (hands gripping objects, with the active region tracking finger configuration around a phone, remote, or bottle), \texttt{img-12-44550} (cat eyes), \texttt{img-12-4113} (mustaches and beards), and \texttt{img-12-45841} (the nose region of human faces).

\begin{figure}[h]
\centering
\includegraphics[width=\linewidth]{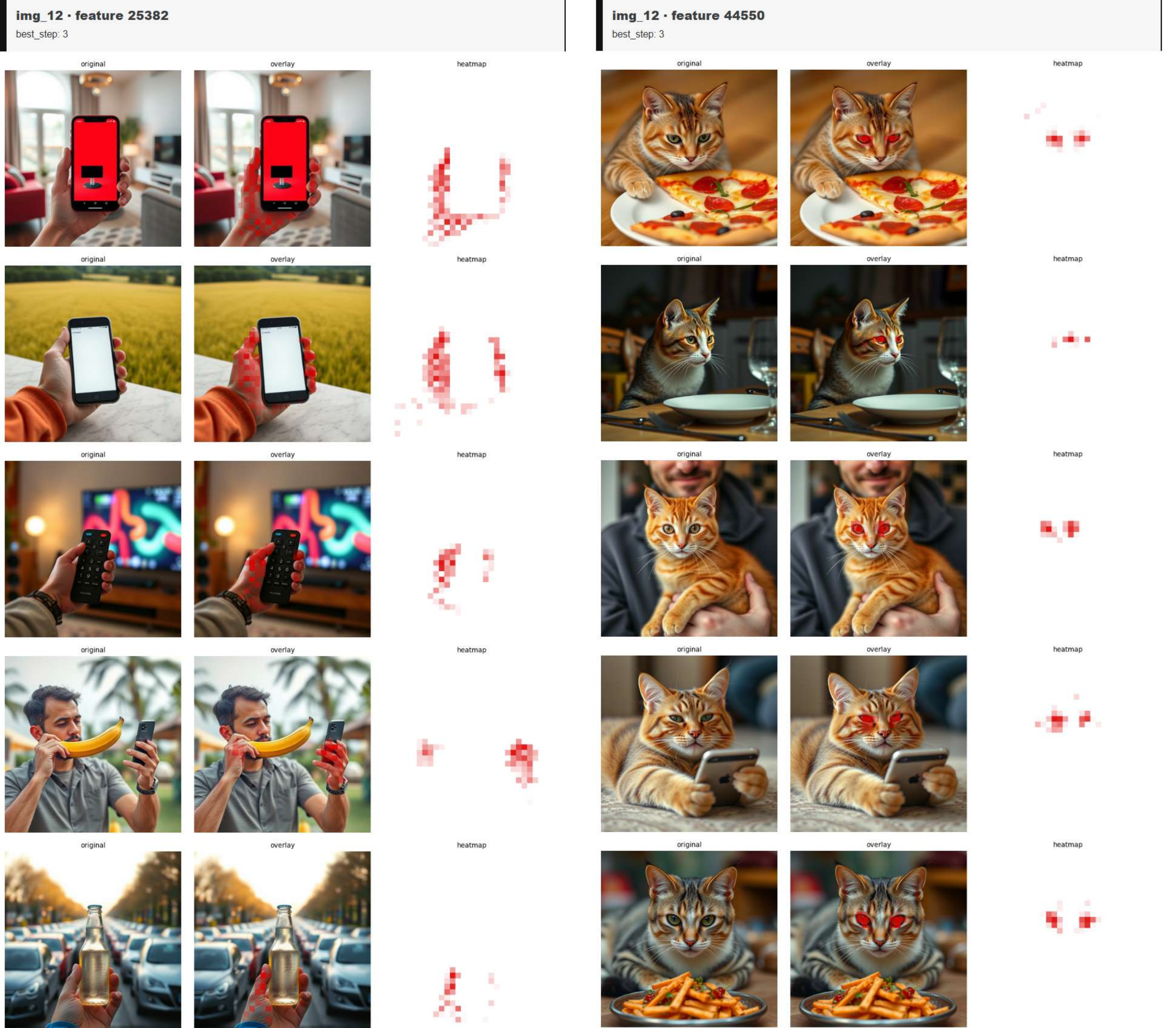}
\caption{Representative semantic features at $\ell = 12$. \textbf{Left:} \texttt{img-12-25382} localizes hands gripping objects across diverse instances. \textbf{Right:} \texttt{img-12-44550} fires on cat eyes.}
\label{fig:feat_img12_objects}
\end{figure}

The most striking finding at $\ell = 12$ is a small group of features encoding scene physics rather than object identity. \texttt{img-12-1023} fires on mirror-like reflections of objects in water, glass, and reflective surfaces, regardless of the object being reflected. \texttt{img-12-10694} activates on light-shadow boundaries (the edge of a tennis player's shadow on the court, the line where a window frame's shadow falls on a wall). \texttt{img-12-21708} highlights cast-shadow regions in their entirety (the shadow of a person's head on a wall, the shadow of a monitor on a desk). The presence of dedicated features for reflections and shadows -- properties of the rendering of a 3D scene rather than of any particular object -- suggests that the middle of the double-stream segment is where the model represents the scene geometrically and not just lexically.

\begin{figure}[h]
\centering
\includegraphics[width=\linewidth]{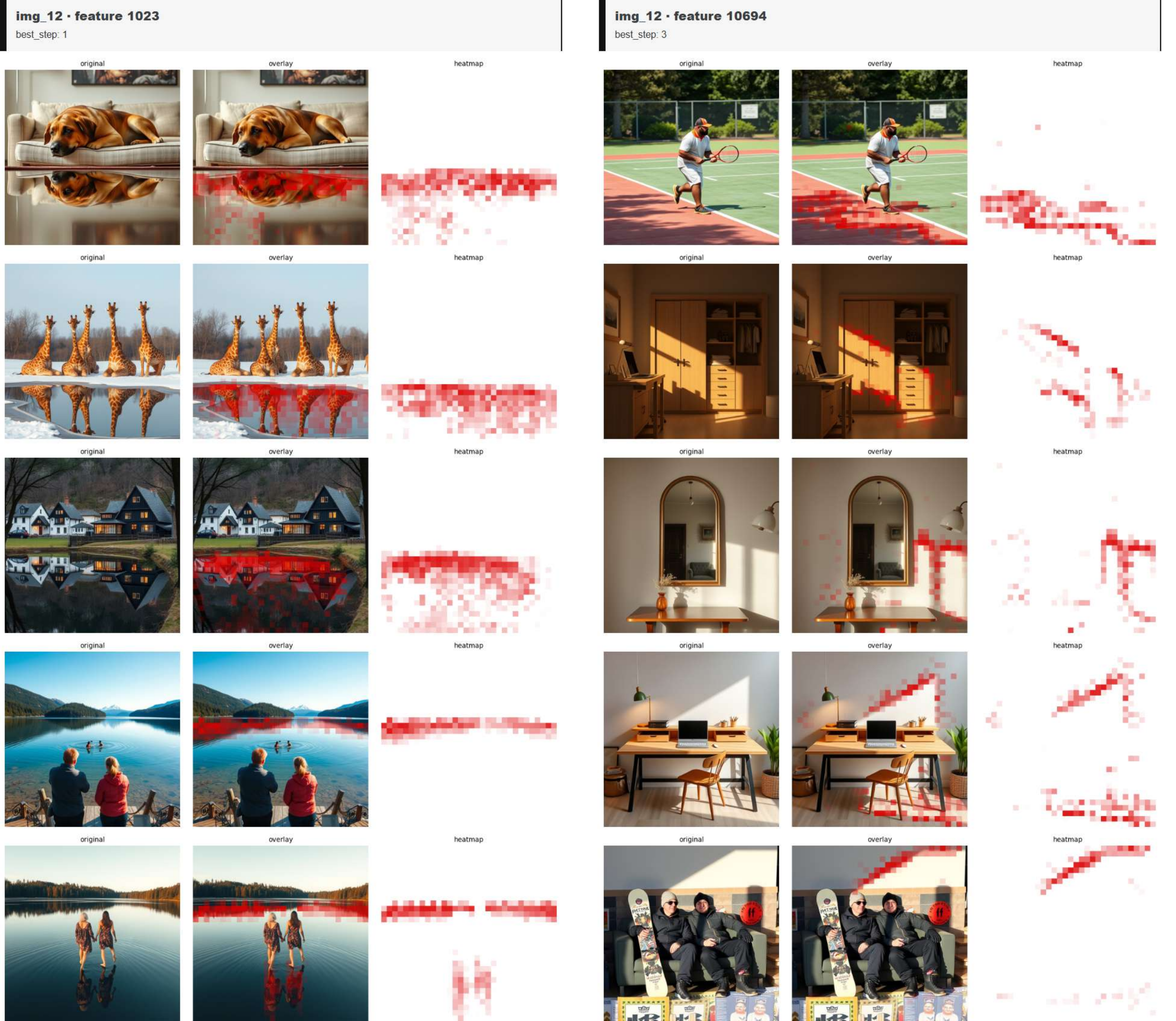}
\caption{Scene-physics features at $\ell = 12$. \textbf{Left:} \texttt{img-12-1023} on mirror reflections. \textbf{Right:} \texttt{img-12-10694} on light-shadow boundaries.}
\label{fig:feat_img12_physics}
\end{figure}

\subsection{Late layer ($\ell = 18$) results}
\label{app:interp:layer18}

\paragraph{Text stream.} The late text transcoder's visualized features overwhelmingly fail to carry lexical content. The dominant category fires on control tokens, primarily the end-of-sequence token \texttt{</s>} (e.g.\ \texttt{txt-18-29365} and many siblings). \texttt{txt-18-8395} fires preferentially on the first prompt token, typically the article \emph{``A''}, occasionally on other position-marking symbols (a leading period or whitespace). A plausible interpretation is that the late text stream, having largely handed its lexical content over to the image stream through preceding rounds of joint attention, repurposes its capacity for global aggregation through control-token positions. Substantive content features still exist but are rare; for instance, \texttt{txt-18-17681} responds to food contexts (\emph{``barbecue sandwich''}, \emph{``bunch of food''}).

\begin{figure}[h]
\centering
\includegraphics[width=\linewidth]{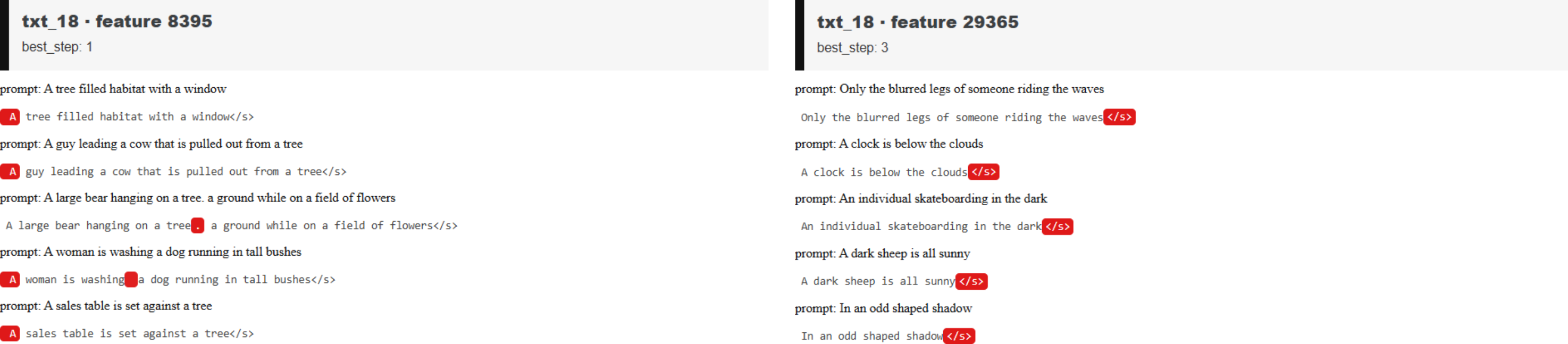}
\caption{Representative text-stream features at $\ell = 18$. \textbf{Left:} \texttt{txt-18-8395} (article, whitespace or dot). \textbf{Right:} \texttt{txt-18-29365} (end-of-sequence token).}
\label{fig:feat_txt18}
\end{figure}

\paragraph{Image stream.} Image features at $\ell = 18$ operate on composition and semantic context rather than primitives or individual objects. \texttt{img-18-47900} fires on the lower supporting plane of the scene (tables, floors), with peak activation at denoising step $0$ -- consistent with an interpretation as a scene-layout feature establishing the horizontal surface on which objects are subsequently placed. \texttt{img-18-18830} localizes the right outer boundary of central objects: active patches do not lie on the object itself but trace its right contour, a compositional feature about object placement rather than object identity. Several features encode high-level semantic context: \texttt{img-18-10948} on tiled walls and bathroom interiors and \texttt{img-18-46496} on urban landscapes.

\begin{figure}[h]
\centering
\includegraphics[width=0.9\linewidth]{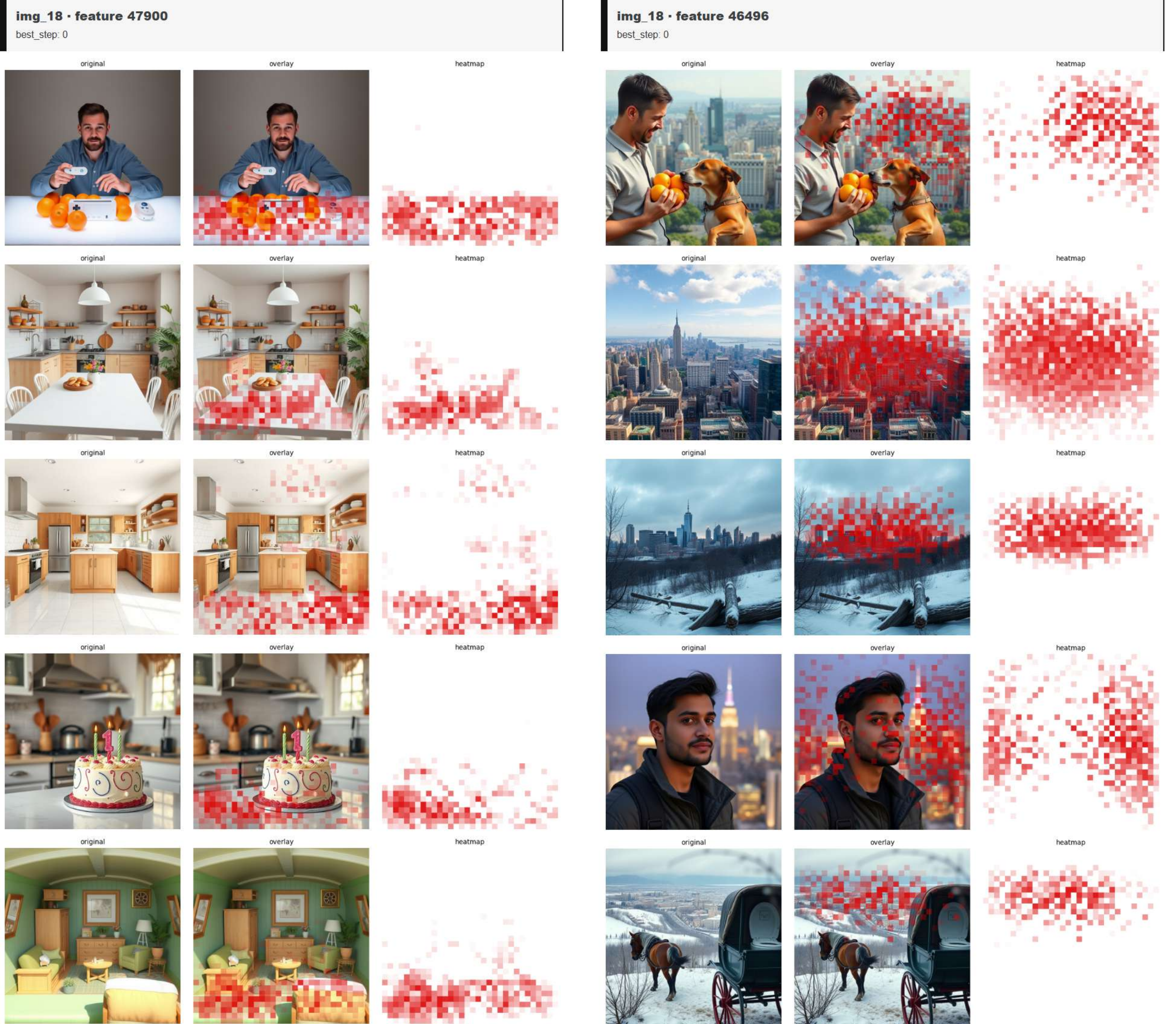}
\caption{Representative image-stream features at $\ell = 18$. \textbf{Left:} \texttt{img-18-47900} (lower supporting plane of the scene). \textbf{Right:} \texttt{img-18-46496} (urban landscape context).}
\label{fig:feat_img18}
\end{figure}

\subsection{Discussion}
\label{app:interp:discussion}

\paragraph{Hierarchy of abstractions.} The level of abstraction grows monotonically with depth in both streams. Text-stream features evolve from individual phrases ($\ell = 6$) through compositional name and quantity concepts ($\ell = 12$) toward control-token aggregators ($\ell = 18$). Image-stream features evolve from edges and color regions ($\ell = 6$) through object parts and scene physics ($\ell = 12$) toward compositional structure ($\ell = 18$). The trajectory parallels what has been reported for autoregressive language models with sparse dictionaries and supports the view that diffusion transformers form analogous hierarchies of representation.

\paragraph{Cross-modal information transfer.} The two streams show inverse interpretability profiles. The fraction of text features tied to substantive lexical content decreases monotonically with depth, while image features remain interpretable through the second half of the analyzed segment, with the highest density of semantic-object features at $\ell = 12$ and a shift toward compositional features by $\ell = 18$. Read together, the two trajectories suggest a one-directional transfer of content from text to image: by the late blocks, the text stream has shed most of its lexical specificity -- its content has already been read by the image stream through preceding rounds of joint attention -- while the image stream maintains a working representation of the scene.

\paragraph{Temporal specialization.} Image features show a consistent dependence on the denoising step that aligns with the diffusion coarse-to-fine progression. At $\ell = 6$, color-oriented features peak at the early steps ($0$--$1$) while geometry-oriented features peak at the late steps ($2$--$3$). At $\ell = 18$, the scene-layout feature \texttt{img-18-47900} peaks at step $0$, consistent with its role of establishing the supporting plane before object placement begins. The picture is consistent with prior reports of step-dependent specialization in diffusion models~\cite{tinaz2025emergence}, and our transcoder features capture it.

\end{document}